%% file: 0-Main.tex
\documentclass[acmlarge]{acmart}

\usepackage{booktabs} 
\usepackage{enumerate}
\usepackage{booktabs}
\usepackage{makecell}
\usepackage{multirow}
\usepackage{flushend}
\usepackage{latexsym}
\usepackage{multirow}
\usepackage{amsmath}
\usepackage{amsfonts}
\usepackage{subfigure}
\usepackage[noend]{algpseudocode}
\usepackage{graphicx}
\usepackage{color}
\usepackage{algpseudocode}  
\usepackage{amsmath}
\usepackage{rotating} 
\usepackage{bm}
\usepackage{lipsum}
\usepackage{verbatim}
\usepackage{adjustbox}
\usepackage[graphicx]{realboxes}
\usepackage{rotating}

\usepackage[linesnumbered,vlined,ruled,commentsnumbered]{algorithm2e}
\newlength\mylen

\makeatletter
\newcommand{\algorithmfootnote}[2][\footnotesize]{%
  \let\old@algocf@finish\@algocf@finish
  \def\@algocf@finish{\old@algocf@finish
    \leavevmode\rlap{\begin{minipage}{\linewidth}
    #1#2
    \end{minipage}}%
  }%
}

\theoremstyle{definition}

\newtheorem{defn}{Definition}[section]

\usepackage{xspace}
\newcommand\RioGNN{\textsc{RioGNN}\xspace}
\newcommand\RSRL{\textsc{RSRL}\xspace}

\usepackage{comment}

\acmJournal{TOIS}
\acmArticle{39}
\acmYear{2021}

\setcopyright{usgovmixed}

    \acmDOI{0000001.0000001}
    
    \received[Received]{April 2021}
    \received[Revised]{August 2021}
    \received[Accepted]{September 2021}

\usepackage{filecontents}

\begin{document}

\title{Reinforced Neighborhood Selection Guided Multi-Relational Graph Neural Networks}

\author{Hao Peng}
\authornote{This is the corresponding author.}
\affiliation{%
  \institution{Beihang University}
  \city{Beijing}
  \country{China}
  }
\email{penghao@buaa.edu.cn}
\author{Ruitong Zhang}
\affiliation{%
  \institution{Beihang University}
  \city{Beijing}
  \country{China}
  }
\email{rtzhang@buaa.edu.cn}
\author{Yingtong Dou}
\affiliation{%
  \institution{University of Illinois at Chicago}
  \city{Chicago}
  \country{USA}
 }
\email{ydou5@uic.edu}
\author{Renyu Yang}
\affiliation{%
  \institution{University of Leeds}
  \city{Leeds}
  \country{UK}
  }
\email{r.yang1@leeds.ac.uk}
\author{Jingyi Zhang}
\affiliation{%
  \institution{Beihang University}
  \city{Beijing}
  \country{China}
  }
\email{turuarua@163.com}
\author{Philip S. Yu}
\affiliation{%
  \institution{University of Illinois at Chicago}
  \city{Chicago}
  \country{USA}
 }
\email{psyu@uic.edu}

\begin{abstract}
Graph Neural Networks (GNNs) have been widely used for the representation learning of various structured graph data, typically through message passing among nodes by aggregating their neighborhood information via different operations.
While promising, most existing GNNs oversimplified the complexity and diversity of the edges in the graph, and thus inefficient to cope with ubiquitous heterogeneous graphs, which are typically in the form of multi-relational graph representations.
In this paper, we propose \RioGNN, a novel \textbf{R}einforced, recurs\textbf{i}ve and flexible neighb\textbf{o}rhood selection guided multi-relational \textbf{G}raph \textbf{N}eural \textbf{N}etwork architecture, to navigate complexity of neural network structures whilst maintaining relation-dependent representations.
We first construct a multi-relational graph, according to the practical task, to reflect the heterogeneity of nodes, edges, attributes and labels. 
To avoid the embedding over-assimilation among different types of nodes, we employ a label-aware neural similarity measure to ascertain the most similar neighbors based on node attributes. 
A reinforced relation-aware neighbor selection mechanism is developed to choose the most similar neighbors of a targeting node within a relation before aggregating all neighborhood information from different relations to obtain the eventual node embedding. 
Particularly, to improve the efficiency of neighbor selecting, we propose a new recursive and scalable reinforcement learning framework with estimable depth and width for different scales of multi-relational graphs.
\RioGNN can learn more discriminative node embedding with enhanced explainability due to the recognition of individual importance of each relation via the filtering threshold mechanism.
Comprehensive experiments on real-world graph data and practical tasks demonstrate the advancements of effectiveness, efficiency and the model explainability, as opposed to other comparative GNN models.
\end{abstract}


\authorsaddresses{
A preliminary version~\cite{dou2020enhancing} of this article appeared in the Proceedings of the 29th ACM International Conference on Information and Knowledge Management, Pages 315–324 (CIKM'20).
Authors' addresses: 
H. Peng, School of Cyber Science and Technology, Beihang University, No. 37 Xue Yuan Road, Haidian District, Beijing, 100191, China; email: \path{penghao@buaa.edu.cn}; 
R. Zhang, School of Cyber Science and Technology, Beihang University, No. 37 Xue Yuan Road, Haidian District, Beijing, 100191, China; email: \path{rtzhang@buaa.edu.cn};
Y. Dou, Department of Computer Science, University of Illinois at Chicago, Chicago, IL; email: \path{ydou5@uic.edu};
R. Yang, School of Computing, University of Leeds, Leeds, LS2 9JT, UK; email: \path{r.yang1@leeds.ac.uk};
J. Zhang, School of Cyber Science and Technology, Beihang University, No. 37 Xue Yuan Road, Haidian District, Beijing, 100191, China; email: \path{turuarua@163.com};
P. S. Yu, Department of Computer Science, University of Illinois at Chicago, Chicago, IL; email: \path{psyu@uic.edu}.
}


\maketitle

\renewcommand{\shortauthors}{H. Peng et al.}

\input{1-Introduction}
\input{2-Definition}
\input{3-Methodology}
\input{4-Experiment}
\input{5-Result}

\input{6-Relatedwork}

\input{7-Conclusion}

\section*{Acknowledgment}
The authors of this paper are supported by NSFC through grants 62002007 and U20B2053, S\&T Program of Hebei through grant 20310101D, Fundamental Research Funds for the Central Universities. 
Renyu Yang is partially supported by UK EPSRC Grant (EP/T01461X/1) and UK White Rose University Consortium. 
Philip S. Yu is partially supported by NSF under grants III-1763325, III-1909323,  III-2106758, and SaTC-1930941.

\bibliography{sample-base} 

\end{document}

%% file: 1-Introduction.tex
\section{Introduction}
\label{sec:Intro}
The advancement of Graph Neural Networks (GNNs) enables the effective representation learning for a variety of areas~\cite{wu2020comprehensive} including bioinformatics, chemoinformatics, social networks, natural language processing~\cite{peng2019hierarchical,peng2018large}, social events~\cite{peng2019fine,peng2021streaming,KPGNN2021}, recommender system~\cite{qiu2020exploiting}, spatial-temporal traffic~\cite{guo2019attention,peng2020spatial}, computer vision and physics~\cite{bapst2020unveiling} where graphs are primarily the denotation. 
GNN models are proved to reach the performance target over massive datasets -- citation networks~\cite{sun2020pairwise,peng2021lime}, biochemical networks~\cite{zitnik2017predicting,peng2020motif}, social networks~\cite{fan2019graph,KPGNN2021}, knowledge graph~\cite{schlichtkrull2018modeling,liu2020kg,yang2021kgsynnet}, commodity networks~\cite{peng2021lifelong}, API call networks~\cite{hei2021hawk}, etc. -- on different tasks, such as node classification~\cite{hamilton2017inductive,velivckovic2018graph,kipf2017semi,chiang2019cluster,peng2021federated}, node clustering~\cite{pan2018adversarially,jin2019graph}, link prediction~\cite{zhang2018link,kazemi2018simple,lei2019gcn}, graph classification~\cite{duvenaud2015convolutional,ying2018hierarchical,peng2020motif,sun2021sugar}, etc.
At the core of GNNs is to operate various aggregate functions~\cite{hamilton2017inductive,velivckovic2018graph,kipf2017semi,xu2018powerful} on the graph structure by passing node features to the neighbors; each node aggregates the feature vectors of its neighbors for computing and updating its new feature vector. 
Empirically, iterations of aggregation or message passing come into a node embedding vector -- a numerical capture of both structural information within the node's multi-hop neighborhood and the attribute information -- empowered by the label propagation mechanisms~\cite{zhu2002learning,liu2010advanced,gregory2010finding}.

Heterogeneous graphs are ubiquitous in real-world systems; a graph typically consists of nodes with multiple types and multi-relational edges between nodes. 
For example, in Yelp spam review data~\cite{rayana2015collective}, there are heterogeneous nodes (e.g., businesses, reviews, users, etc.) and relations (e.g., posted by the same user, under the same product with the same star rating, and under the same product posted in the same month between two reviews). 
However, existing iterative aggregation mechanisms of GNNs have yet to elaborately consider the diversity of semantic relations and the usability of the proposed models. 
Homogeneous GNNs such as GraphSAGE~\cite{hamilton2017inductive}, GCN~\cite{kipf2017semi}, GAT~\cite{velivckovic2018graph}, GIN~\cite{xu2018powerful} ignore or simplify the diversity and complexity of the nodes and edges in practical networks, which is inadequate to represent the heterogeneity of data. 
To solve the above problem, relational GNNs~\cite{schlichtkrull2018modeling,nathani2019learning,zhu2019relation} are proposed but fail to capture multiple hop or complex relations. 
Sampling-based heterogeneous GNNs guided by hand-crafted meta-paths~\cite{wang2019heterogeneous}, meta-graphs~\cite{yang2018meta} and meta-schema~\cite{he2019hetespaceywalk} are solely based on data types and their structured connections. 
This drawback substantially impedes the generalization of such heterogeneous GNNs in practical fine-grained tasks -- e.g., fraud detection~\cite{abdallah2016fraud,dou2019review}, disease diagnosis~\cite{fatima2017survey}, etc. -- where it is infeasible or inefficient to externalize the inherent entity relations through meta-structures such as meta-path, meta-graph and meta-schema. 
Take the detection and diagnosis of diabetes and its suspected diseases, based on the MIMIC-III dataset~\cite{johnson2016mimic}, as an example. 
Observably, a portion of patients with diabetes tends to have symptoms that cause glaucoma, while glaucoma patients do not often experience issues in blood sugar, insulin and other test indicators. Accordingly, one can easily define explicit relations such as \emph{having hyperglycemia score in the blood test}, \emph{having high proteinuria scores in urine test}, \emph{having symptoms of glaucoma in vision test}, \emph{having high intraocular in pressure test}, etc., between any two patients.
It is far more useful to specify relations based on common attributes shared by entities, and less dependent upon strict entity connections as opposed to the meta-structure based approaches which have to leverage complicated automated generation technology~\cite{meng2015discovering} or manual experience~\cite{cao2020multi,hosseini2018heteromed,liu2020health}. 
Hence, it is more effective to explore and exploit the explicit relations, stemming from task-specific characteristics, for carrying out downstream applications.

In an attempt to extend GNNs for supporting heterogeneous graph embedding, many approaches rely on a combination of sophisticated neural networks~\cite{wang2020survey}. 
For instance, HetGNN~\cite{zhang2019heterogeneous} aggregates multi-modal features from heterogeneous neighbors by combining bi-LSTM, self-attention, and types combination. 
RSHN~\cite{zhu2019relation} utilizes coarsened line graph neural network (CL-GNN) along with the message passing neural network (MPNN) to learn node and edge type embedding simultaneously.
HGT~\cite{hu2020heterogeneous} leverages type dependent parameters based mutual attention, message passing, residual connection, target-specific aggregation function, etc.
MAGNN~\cite{fu2020magnn} makes use of meta-path sampling, intra-meta-path and inter-meta-path aggregation technologies to embed a node with the targeted type.
Nevertheless, they lack the analysis of more practical or fine-grained application tasks and require strong domain knowledge to build the complex neural network structures. 
The usability of the proposed GNN model should also be designed in a more convenient way.

To this end, we propose \RioGNN, a novel \underline{R}einforced, recurs\underline{I}ve and scalable neighb\underline{O}rhood selection guided multi-relational \underline{G}raph \underline{N}eural \underline{N}etwork, to navigate complexity of customized neural network structures whilst maintaining relation-dependent representations. 
For domain task driven graph representation learning, we introduce \emph{multi-relational graph} to reflect the heterogeneity of nodes, edges, attributes and labels.
Herein, a relation is referred to as a specific type of edge between two nodes, connected with each other through explicit common attributes or implicit semantics, e.g., two products released in the same month, two movies directed by the same director, etc. 
Departing from heterogeneous information network (HIN)~\cite{shi2016survey}, the multi-relational graph is able to flexibly characterize and explicitly differentiate the edge types without the need for specifying semantic connectivity between any two nodes strictly following entity-associated meta-structures. 
For a given relation, we can conduct the sampling procedure upon the original graph for extracting neighbors for each node in the graph.

To diminish the complexity of neural network units, \RioGNN optimizes the process of neighbor selection when aggregating neighbor information for a center node embedding.  
To avoid the embedding over-assimilation among different types of nodes, we first employ a label-aware neural similarity measure to ascertain the most similar neighbors based on node attributes.
Particularly, this is achieved by a neural classifier that transforms the supervised signals (e.g., high-fidelity annotated labels) and original node features to calculate the node similarity. 
To follow up, we carry out a relation-aware neighbor selection to choose the most similar neighbors of a targeting node within a reinforced relation before aggregating all neighborhood information from different relations to obtain the eventual node embedding.  
To improve the neural classifier training efficiency, we optimize the filtering threshold within each relation through \RSRL, a novel \underline{R}ecursive and \underline{S}calable \underline{R}einforcement \underline{L}earning framework with estimatable depth and width for different scales of a heterogeneous graph in a recursive manner.
Specifically, we exploit two general relation-aware RSRL methods -- using both discrete and continuous strategies -- for pinpointing the optimal number of neighbors of different relations.
The discrete and continuous approaches can generally provide more choices in the face of different datasets and application scenarios. 
\RSRL not only facilitates to learn discriminative node embeddings, but also makes the model more explainable as we can recognize the individual importance of each relation via the filtering thresholds.
\RSRL-based relation-aware neighbor selector can be integrated with any mainstream reinforcement learning models~\cite{hasselt2016deep,schulman2017proximal, haarnoja2018soft,konda2000actor,watkins1992q} and neighbor aggregation functions~\cite{hamilton2017inductive,velivckovic2018graph} used for specific scenarios.

We integrate the aforementioned techniques with the vanilla GNN as a layer of \RioGNN and devise multi-layered \RioGNN to learn high-order node representations according to the specific requirements of downstream tasks. 
This paper mainly targets those tasks with node-level embedding and learns the multi-relation node representation in a semi-supervised manner.
We evaluate the effectiveness, efficiency and explainability of \RioGNN by applying it to two tasks of fraud detection and diabetes detection, using Yelp, Amazon and MIMIC-III datasets.
Experiments assess how \RioGNN underpins downstream tasks including transductive node-classification, inductive node-classification and node clustering. 
Results show that \RioGNN significantly improves various downstream tasks over state-of-the-art GNNs as well as dedicated heterogeneous models by 0.70\%–32.78\%.
We show that our RSRL framework not only boosts the learning time by up to 4.52x, but also achieves 4.90\% improvement in node classification.
We also evaluate the sensitivity of \RioGNN to hyper-parameters in the above tasks.
Finally, we carry out a series of case studies to showcase how \RSRL automatically learns the importance and engagement of implicit relations in different tasks. 
The source code and datasets are publicly available at \url{https://github.com/safe-graph/RioGNN}.

The contributions of this work are summarized as follows:
\begin{itemize}
\item The first task-driven GNN framework based on multi-relational graphs, making the best use of relational sampling, message passing, metric learning and reinforcement learning to guide neighbor selection within and across different relations. 

\item A flexible neighborhood selection framework that employs a reinforced relation-aware neighbor selector with label-aware neural similarity neighbor measures.

\item A recursive and scalable reinforcement learning framework that learns the optimized filtering thresholds via estimable depth and width for different scales of graphs or tasks. 

\item The first study on the explainability of multi-relational GNNs from the perspective of importances of different relations.
\end{itemize}

We expand upon our preliminary work~\cite{dou2020enhancing},  by extending CAmouflage-REsistant GNN (CARE-GNN) model exclusively for fraud detectors against camouflaged fraudsters to a more general architecture underpinning a wide range of practical tasks.
Specifically, the improvements encompass: 
1) giving a full version of definition, motivation and aim of multiple relation graph neural networks under different practical tasks; 
expanding the label-aware similarity neighbor measure from one layer to multiple layers to select the similar neighbors; 
2) proposing a novel recursive and scalable reinforcement learning framework to optimize the filtering threshold for each relation along with the GNN training process in a general and efficient manner, instead of the previous Bernoulli Multi-armed Bandit method; 
3) leveraging both discrete and continuous strategies to find the optimal neighbors of different relations to be selected under the reinforcement learning framework;
4) carrying out extensive experiments on three representative and general-purpose datasets, not limited to the fraud detection scenario. 
Furthermore, more in-depth experimental results are discussed to demonstrate the effectiveness and efficiency of the proposed architecture.
We supply the variances of the results of multi-relational graph representation learning.
We also showcase the explanation of importances of different relations from a new perspective based on the filtering threshold of the proposed \RSRL framework.

The paper is structured as follows: 
Section~\ref{sec:background} outlines the preliminaries and the problem formulation, and Section~\ref{sec:method} describes the technical details involved in \RioGNN. 
Experimental setup and results are discussed in Section~\ref{sec:experimental-setup} and Section~\ref{sec:result}, respectively. 
Section~\ref{sec:relatedwork} presents the related work before we conclude the paper in Section~\ref{sec:conclusion}. 

%% file: 2-Definition.tex
\section{Background and Overview}
\label{sec:background}

\begin{table}[b]
\caption{Notations.} 
\small 
\centering
\begin{tabular}{r|l}  
\hline\hline
\textbf{Symbol} & \textbf{Definition} \\
\hline
$\mathcal{G}; \mathcal{V}; \mathcal{E}; \mathcal{X}$ & Graph; Node set; Edge set; Node feature set \\
\hline
$y_{v}; Y$ & Label for node $v$; Node label set \\
\hline
$r; R$ & Relation; Total number of relations\\
\hline
$l; L$ & GNN layer number; Total number of layers \\
\hline
$b; B$ & Training batch number; Total number of batches \\
\hline
$e; E$ & Training epoch number; Total number of epochs \\
\hline
$\mathcal{V}_{train}; \mathcal{V}_{b}$ & Nodes in the training set; Node set at batch $b$ \\
\hline
$\mathcal{E}^{(l)}_{r}$ & Edge set under relation $r$ at the $l$-th layer \\
\hline
$\mathbf{h}_{v}^{(l)}$ & The embedding of node $v$ at the $l$-th layer  \\
\hline
$\mathbf{h}_{v, r}^{(l)}$ &  The embedding of node $v$ under relation $r$ at the $l$-th layer     \\
\hline
$\mathcal{D}^{(l)}(v, v^{\prime})$ & The distance between node $v$ and $v^{\prime}$ at the $l$-th layer \\
\hline
$S^{(l)}(v, v^{\prime})$ &  The similarity between node $v$ and $v^{\prime}$ at the $l$-th layer\\ 
\hline
$p_{r}^{(l)}\in P$   &  The filtering threshold for relation $r$ at the $l$-th layer  \\
\hline
$RLF^{(l)}$   &  The Reinforcement Learning Forest at the $l$-th layer  \\
\hline
$RLT_r^{(l)}$   &  The Reinforcement Learning Tree for relation $r$ at the $l$-th layer  \\
\hline
$RL_{r}^{(l)(d)}$   &  The Reinforcement Learning Module for relation $r$ at the $l$-th layer in $d$-th depth \\
\hline
$W_r^{(l)(h)}$   &  The width of Reinforcement Learning Tree for relation $r$ at the $l$-th layer in $d$-th depth \\
\hline
$D_r^{(l)}$   &  The depth of  Reinforcement Learning Tree for relation $r$ at the $l$-th layer  \\
\hline
$a_{r}^{(l)} \in A$ & RL action space;\\
\hline
$s(\mathcal{D}_{r}^{(l)(d)})^{(e)}$ & RL state for relation $r$ at the $l$-th layer in $d$-th depth when the epoch is $e$  \\
\hline
$g_{r}^{(l)(d)(e)}$ & RL reward for relation $r$ at the $l$-th layer in $d$-th depth when the epoch is $e$  \\
\hline
$f_{r}^{(l)(d)(e)} $ &  RL iterative function for relation $r$ at the $l$-th layer in $d$-th depth when the epoch is $e$ \\
\hline
$\mathcal{N}_{r}^{l}(v)$ & Nodeset after RL filtering for relation $r$ at the $l$-th layer\\
\hline
$\textnormal{AGG}_{r}^{(l)}$ &  Intra-relation aggregator for relation $r$ at the $l$-th layer           \\
\hline
$\textnormal{AGG}_{all}^{(l)}$ & Inter-relation aggregator at the $l$-th layer \\
\hline
$\mathbf{z}_{v}$ & Final embedding for node $v$         \\
\hline
\hline
\end{tabular}
\label{tab:notation}
\end{table}

\subsection{Problem Definition}

In this section, we firstly define the multi-relational graph and multi-relational GNN.
All important notations in this paper are summarized in Table~\ref{tab:notation}.

\begin{defn}
\label{def:multi-relation-graph}
\textbf{Multi-Relational Graph (MR-Graph).}
MR-Graph is defined as $\mathcal{G}=\left\{\mathcal{V}, \mathcal{X}, \{\mathcal{E}_{r}\}|_{r=1}^{R}, Y\right\}$, where $\mathcal{V}$ is the set of nodes $\{v_{1}, \dots, v_{n}\}$, and n is the number of nodes in the graph.
Each node $v_{i}$ has a $d$-dimensional feature vector $\mathbf{x}_{i}\in\mathbb{R}^{d}$ and $\mathcal{X} = \{\mathbf{x}_{1}, \dots, \mathbf{x}_{n}\}$ represents a set of all node features.
$e_{i, j}^{r} = (v_{i}, v_{j})\in \mathcal{E}_{r}$ is an edge between $v_{i}$ and $v_{j}$ with a relation $r \in \{1, \cdots, R\}$, where R is the number of relations.
Note that an edge can be associated with multiple relations, and there are $R$ different types of relations.
$Y$ is the set of labels for each node in $\mathcal{V}$.  
\end{defn}

\begin{figure}[t]
\centering
\scalebox{0.88}{
\subfigure[An MR-Graph example for fraud review detection.]{
\includegraphics[width=0.48\textwidth]{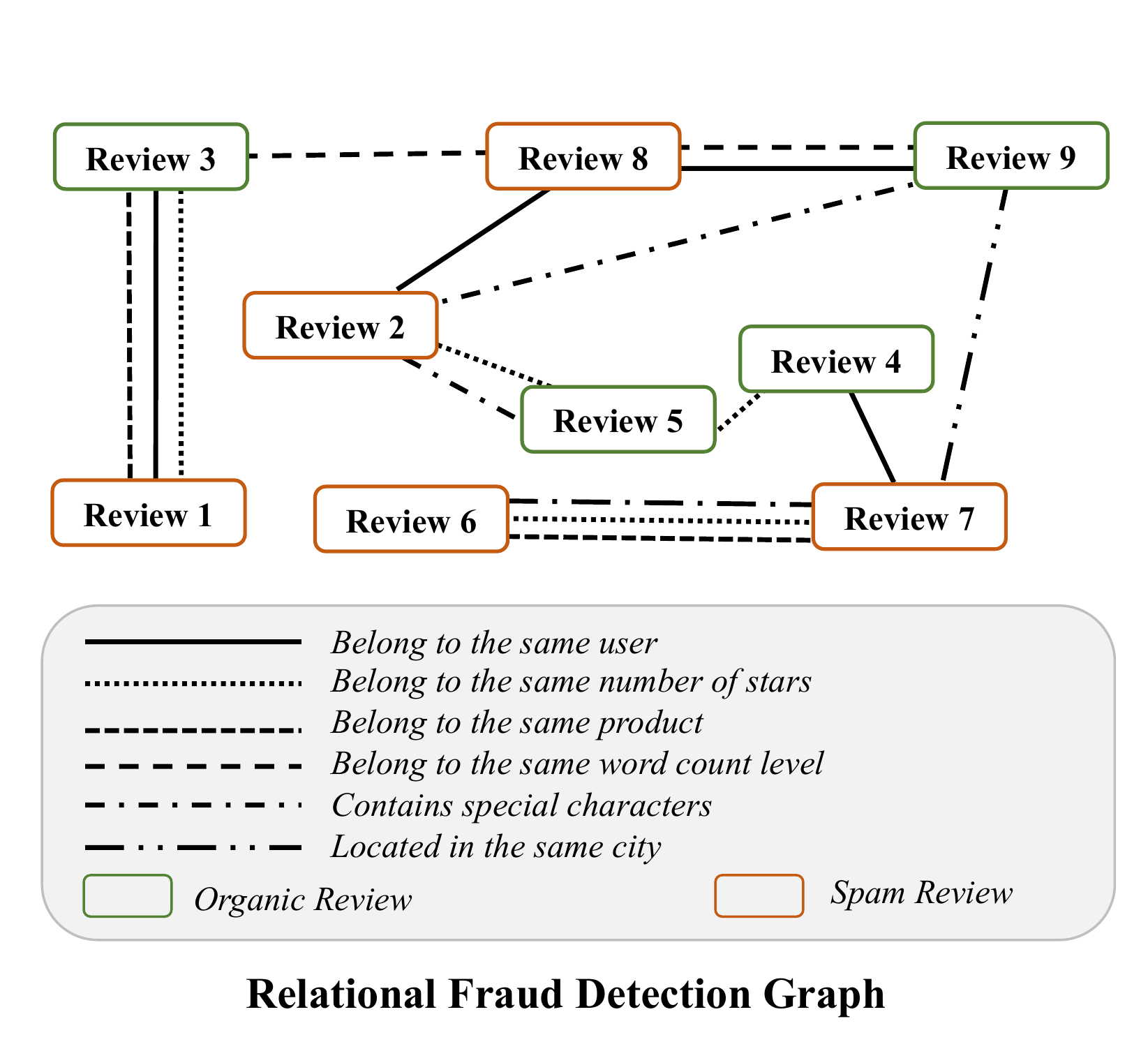}\label{fig:relation-fraud}}
\subfigure[A HIN example from review data~\cite{sun2019opinion,cao2017hitfraud}.]{
\includegraphics[width=0.48\textwidth]{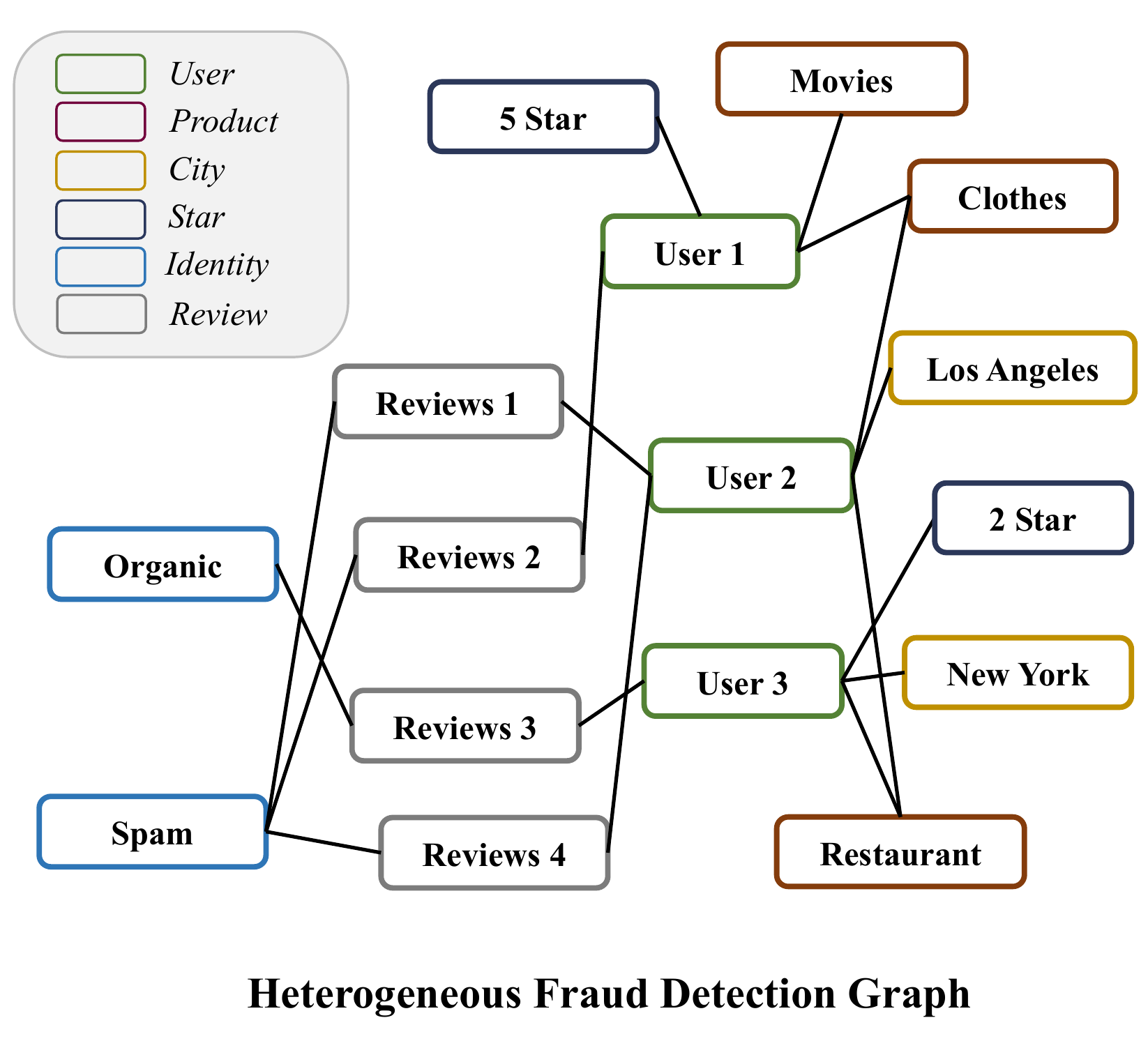}\label{fig:raw-hin-yelp}
}
}
\vspace{-1mm}
\caption{Graph Modeling in Fraud Review Detection.}\label{fig:multi-relation-fraud}
\end{figure}

The multi-relational graph directly uses the elements to be classified as nodes, and the key relations of elements with different labels are used as multiple connections, which can be widely used in challenging classification tasks.
It is worth noting that, departing from HIN, the multi-relational graph is able to flexibly characterize and explicitly differentiate the edge types without the need for specifying semantic connectivity between any two nodes strictly following entity-associated meta-structures. 
We exemplify its applicability by two real-world applications and compare the difference between the MR-Graph based modeling and the HIN-based approach:

\textbf{Spam Review Detection.} 
Spam reviews are referred to as those fabricated reviews posted to products or merchants with the intent of promoting their targets.
Fraud detection has to identify spam reviews from organic ones.
As spam comments add some special characters or simulate benign email behaviors (such as one user who posts spam emails while maintaining a certain frequency of organic comments) to avoid being found out, this brings challenges to distinguishing spam comments.
We consider the comments with different labels as nodes, and different representative interactions as different types of connections to build a multi-relational graph, thereby transforming this problem into a two-classification problem.
As shown in Figure~\ref{fig:relation-fraud}, an MR-Graph example depicts the organic reviews, spam reviews and their interactions extracted from the e-commerce review data~\cite{mukherjee2013yelp,rayana2015collective}.  
We extracted representative interactions between two reviews that are closely associated with the fraudulent behavior, and represented them as different types of edges -- \emph{Belonging to the same user}, \emph{Having the same star rating}, \emph{Targeting the same product posted in the same month}, \emph{Belonging to the same word count level}, \emph{Containing special characters} and \emph{Targeting products located in the same city}. 
As an alternative, traditional HIN-based modeling (Figure~\ref{fig:raw-hin-yelp}) pays more attention to the relations outlined by structured connections.

\begin{figure}[t]
\centering
\scalebox{0.88}{
\subfigure[An MR-Graph for diabetes and disease detection.]{
\includegraphics[width=0.48\textwidth]{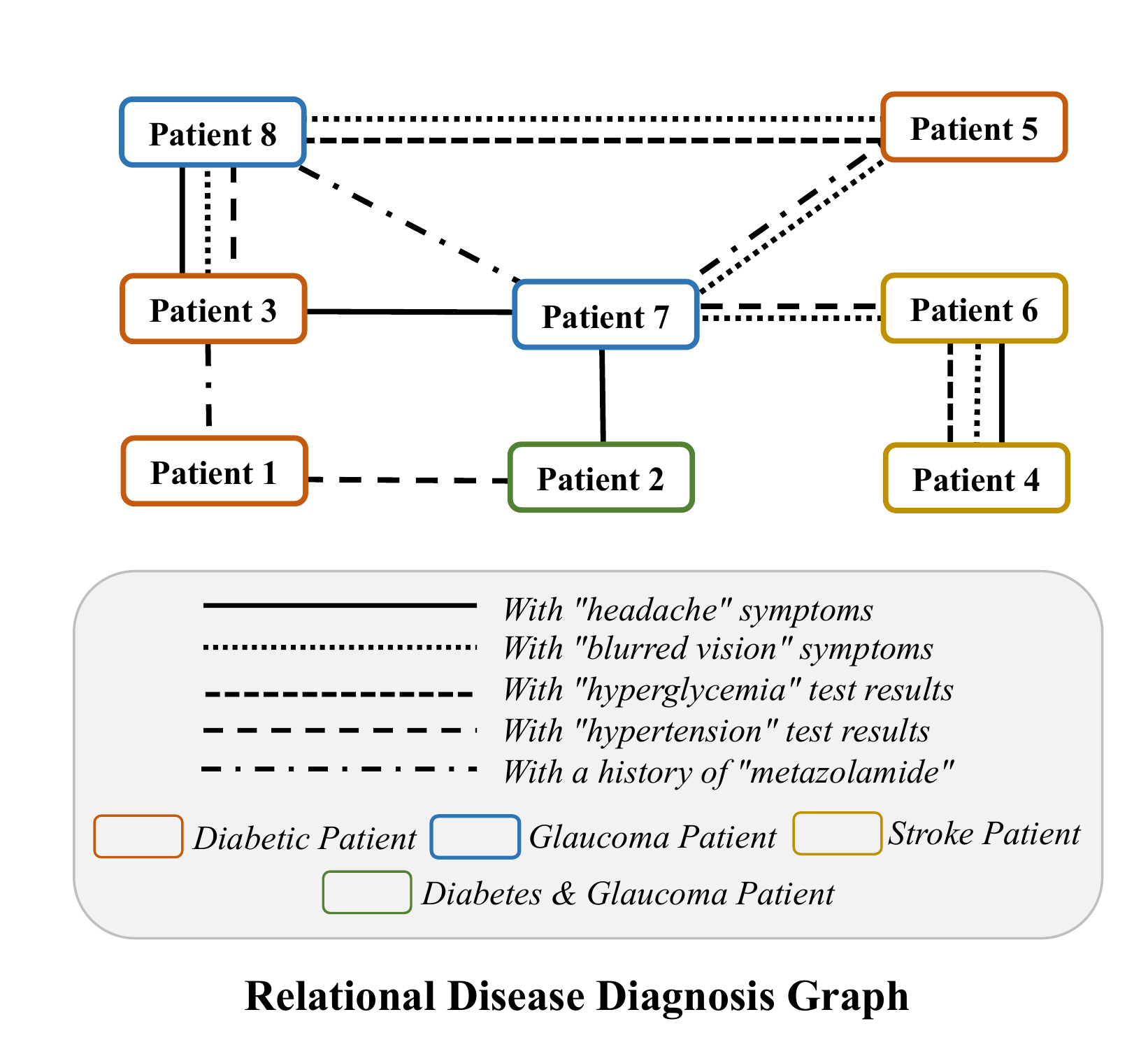}\label{fig:relation-patient}}
\subfigure[An example of heterogeneous Electronic Health Records (EHR) graph, reproduced from~\cite{liu2020health}.]{
\includegraphics[width=0.48\textwidth]{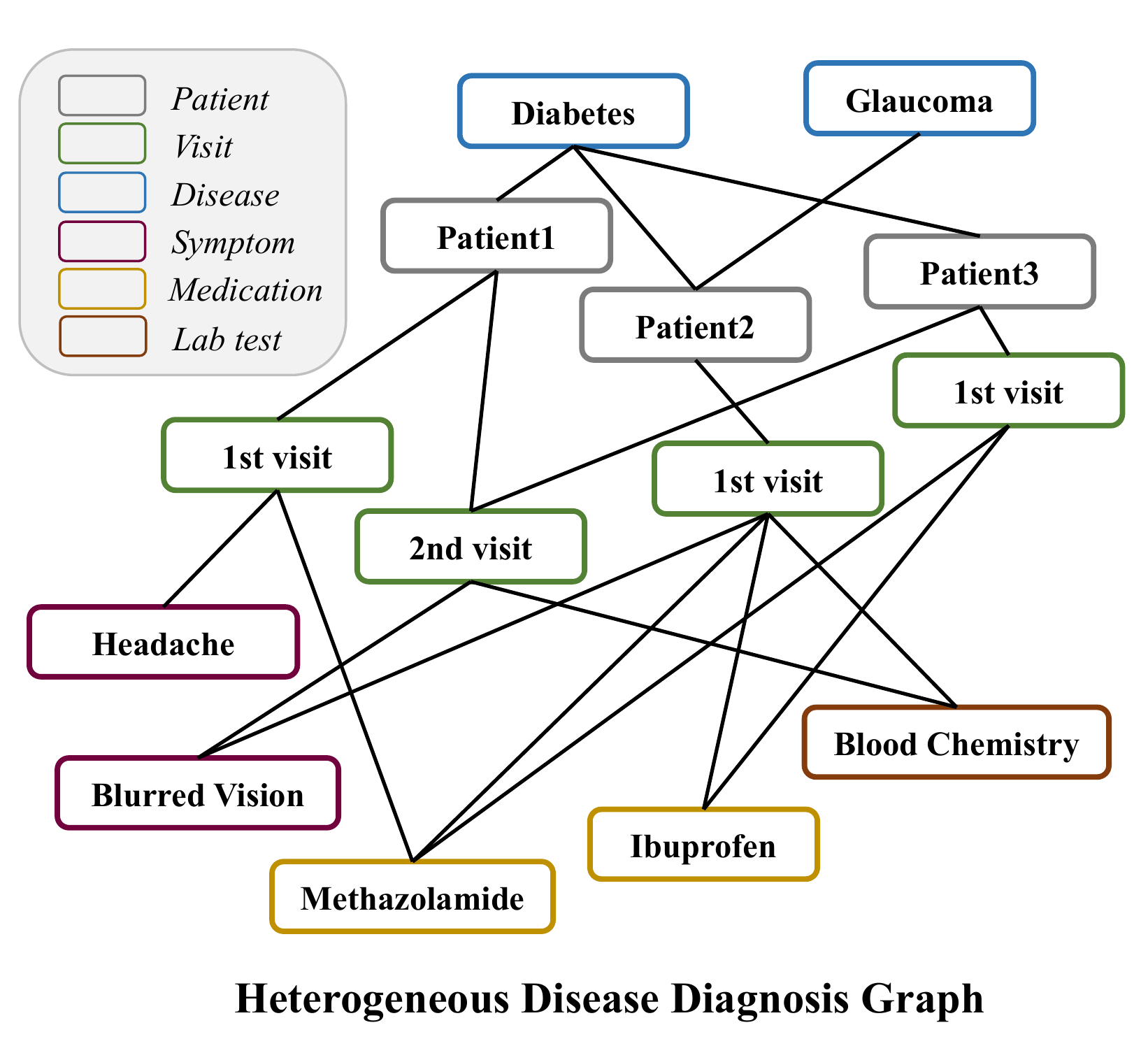}\label{fig:raw-hin-ehr}}
}
\vspace{-1mm}
\caption{Graph Modeling in Disease Diagnosis.}
\label{fig:multi-relation-disease}
\end{figure}

\textbf{Disease Diagnosis.}  
In the task of disease diagnosis, diabetes, stroke, and glaucoma are common diseases in middle-aged and elderly people, and their early symptom recognition is very important. 
However, because these three diseases have similar symptoms, it is difficult to distinguish patients in clinical practice.
For example, the symptoms of stroke include loss of vision, sudden weakness and tingling sensations, which are similar to symptoms in patients with type II diabetes.
In addition, one of the early symptoms of diabetes is blurred vision caused by changing fluid levels. Therefore, the eyes may change shape, disturbing the focusing ability of the eyes. Although this visual impairment may indicate diabetes, it is also true in glaucoma patients.
Here, taking the patient as the node of the multi-relational graph and connecting the patients with different similar symptoms into different types of edges can convert the task into a multi-classification task.
Figure~\ref{fig:relation-patient} illustrates an MR-graph of patients for disease diagnosis.
We believe to model and represent the relationship between the following types of patients: \emph {with "headache" symptoms}, \emph{with "blurred vision" symptoms}, \emph{with "hyperglycemia" test results}, \emph{with "hypertension" test results}, \emph{with a history of "metazolamide"}, etc. are more helpful for the diagnosis of diabetes and its suspected diseases of patients.
Even, there may be multiple relationships between two patients.
For example, \emph{Patient 5} and \emph{Patient 8} have two relationships of \emph{with "blurred vision" symptoms} and \emph{with "hypertension" test results} in common.
Meanwhile, previous HIN-based Electronic Health Records (EHR) modelings~\cite{hosseini2018heteromed,cao2020multi,liu2020health} focused on the correlation and fusion of different attributes or types of data, and the corresponding methods are suitable for the diagnosis of all kinds of diseases.
Moreover, there is a lack of fine-grained analysis of certain definite diseases.
We also give an illustration of the heterogeneous Electronic Health Records graph in Figure~\ref{fig:raw-hin-ehr}.

\begin{defn}\label{def:MR-GNN}
\textbf{Multi-Relational GNN.}
Graph neural network (GNN) is a deep learning framework to embed graph-structured data via aggregating the information from its neighboring nodes~\cite{hamilton2017inductive,kipf2017semi,gilmer2017neural,velivckovic2018graph}.
Based on Definition~\ref{def:multi-relation-graph}, we can then outline a unified formulation of the Multi-Relational GNN from the perspective of multi-layer neighbor aggregation according to different relations.
For a central node $v$, the hidden or aggregated embedding of node $v$ at $l$-$th$ layer is referred to as $\mathbf{h}_{v}^{(l)}$:
\begin{equation}\label{equ:MR-GNN}
\mathbf{h}_{v}^{(l)}=\sigma(\mathbf{h}_{v}^{(l-1)}\oplus{}AGG^{(l)}(\{\mathbf{h}_{v^{'}, r}^{(l-1)}:(v, v^{'})\in\mathcal{E}^{(l)}_{r}\}|_{r=1}^{R})),
\end{equation}
where $\mathcal{E}^{(l)}_{r}$ denotes the edges under the relation $r$ at the $l$-th layer, and $\mathbf{h}_{v', r}^{(l-1)}$ indicates to the aggregated embedding of neighboring node $v'$ under relation $r$.
$AGG$ denotes the aggregation function that maps the neighborhood information from different relations into a vector, e.g., mean aggregation~\cite{hamilton2017inductive} and attention aggregation~\cite{velivckovic2018graph}.
$\oplus{}$ is the operator that combines the information of node $v$ and its neighboring information through either concatenation or summation~\cite{hamilton2017inductive}.
We initialize the node embedding $\mathbf{h}_{v}^{(0)}$ with the input d-dimensional feature vector $x$.
The GNN is trained with partially labeled nodes with binary classification loss functions. 
Instead of directly aggregating the neighbors for all relations, we separate the aggregation part as \emph{intra-relation} aggregation and \emph{inter-relation} aggregation process. 
During the intra-relation aggregation process, the embedding of neighbors under each relation is aggregated simultaneously. 
Then, the embeddings for each relation are combined during the inter-relation aggregation process. 
Finally, the node embeddings at the last layer are used for predictions.
\end{defn}

\subsection{Problem Scope and Challenges}\label{sec:challenges}

In practical applications, we model multi-relation graphs, and take actual problems as node classification tasks in a semi-supervised learning manner.
After constructing a multi-relational graph according to domain knowledge, e.g., Spam Review Detection in Figure~\ref{fig:relation-fraud}, Disease Diagnosis in Figure~\ref{fig:relation-patient}, etc., a multi-relational GNN can be trained.
However, when we train more discriminative, effective and explainable node embedding, there are three main challenges facing the Multi-Relational GNNs:

\textbf{How to cope with misbehaved nodes during neighbor aggregation in GNNs (Challenge 1).}
The input node features $\mathcal{X}$, often extracted based on heuristic methods such as TF-IDF, Bag-of-Words, Doc2Vec, etc., are susceptible to such misbehavior as adversarial attacks,  camouflages~\cite{sun2018adversarial, dou2020enhancing}, or simply imprecise feature selection. 
Consequently, the numerical embedding of a central node tends to be assimilated by misbehaved neighboring nodes.
For instance, in the spam review detection task, adversarial or camouflaged behaviors are non-negligible noises that drastically reduce the accuracy of feature representation learning by GNNs. 
Either feature~\cite{dou2020robust,li2019spam} or relational~\cite{kaghazgaran2018combating,zheng2017smoke} camouflages could similarize the features of misbehaved and benign entities, and further mislead GNNs to generate uninformative node embeddings. In the medical disease diagnosis task, textual attribute based feature selection may not be able to extract high-level or fine-grained semantics, and thus easily lead to imprecise node characteristics. 
Hence, these issues necessitate an effective similarity measure to filter the neighbors before applying into any GNNs.

\textbf{How to adaptively select the most suitable neighbor nodes based on the similarity measure (Challenge 2).}
Data annotation is expensive for most practical problems, and we cannot select all similar neighbors under each relationship through data labeling.
The method of directly regarding the filtering threshold as a hyper-parameter ~\cite {Chen2019iterative, liu2020alleviating} is no longer valid for multiple relationship graphs with numerous noisy or misbehaved nodes.
First, different relationships have different feature similarity and label similarity.
Secondly, different relationships have different precision requirements for the filtering threshold.
Therefore, an adaptive sampling mechanism must be designed so that the optimal number of similar neighbors can be selected for specific relationship requirements in a dynamic environment.

\textbf{How to efficiently learn and optimize the filtering threshold in a continuous manner (Challenge 3).}
Our preliminary work~\cite{dou2020enhancing} adopts the Bernoulli multi-armed bandit framework~\cite{sutton2018reinforcement} with a fixed strategy to strengthen the learning of the filtering threshold. 
However, it is substantially limited by the observation range of the state and manually-specified strategies, and hence the final convergence result of the filtering threshold tends to be locally optimal.
In addition, for maintaining the prediction accuracy, in the face of large-scale datasets, it is imperative to reduce the adjustment step size of the filtering threshold or use continuous action space. 
This procedure will undoubtedly expand the action space, leading to an increased number of convergence periods and a huge growth of calculation, possibly with a loss of accuracy. 
This issue therefore necessitates an automatic and efficient reinforcement learning framework that can quickly obtain sufficient and high-quality solutions.

%% file: 3-Methodology.tex
\section{Methodology}\label{sec:method}

\begin{figure}[h]
  \centering
  \includegraphics[width=0.93\textwidth]{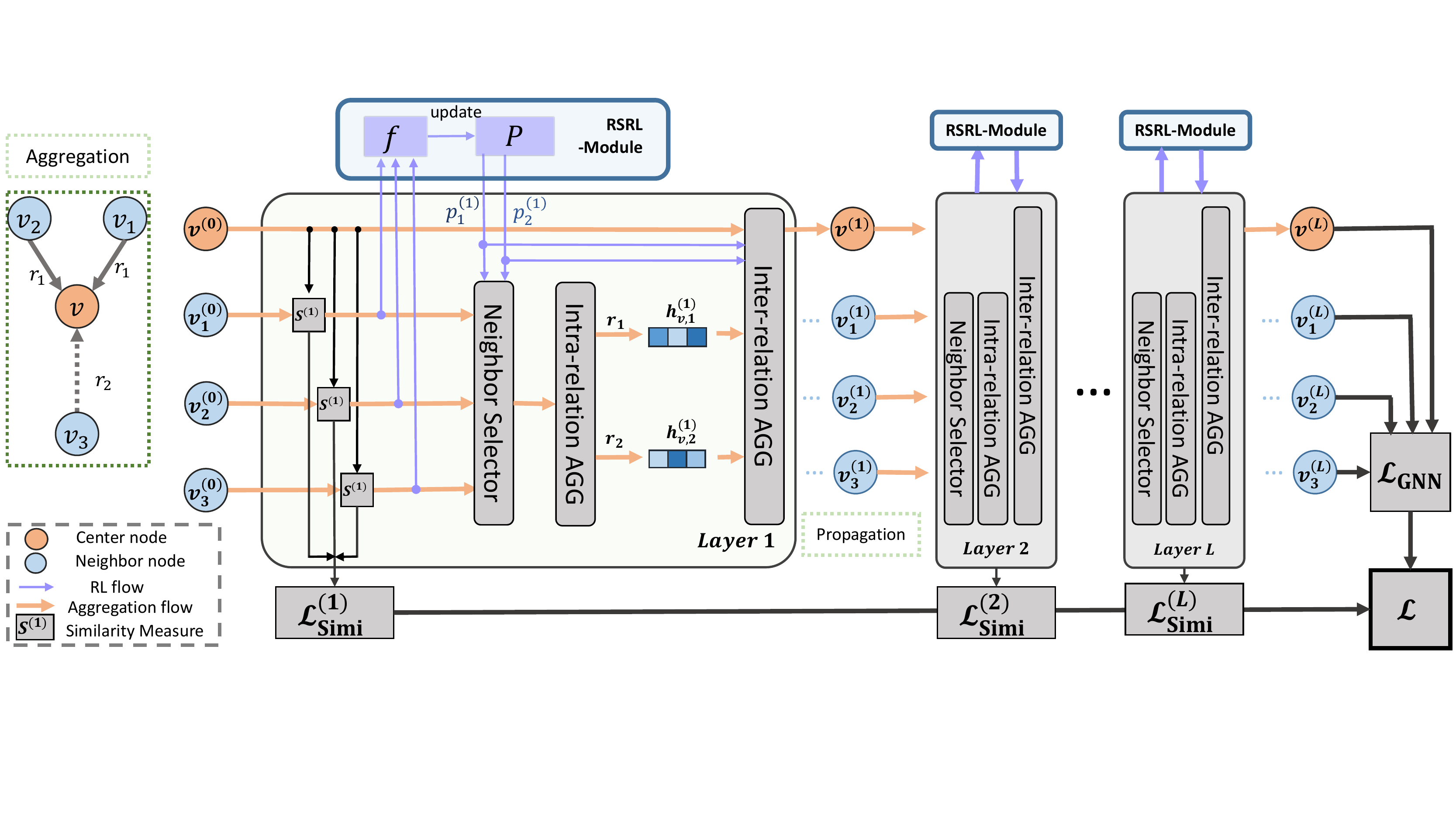}
  \vspace{-1.2mm}
  \caption{\RioGNN architecture.}    
  \vspace{-1.2mm}
  \label{fig:Framework-RioGNN}
\end{figure}

Figure~\ref{fig:Framework-RioGNN} depicts the \RioGNN's overall architecture consisting of three key modules -- label-aware similarity measurement  (Section~\ref{sec:label-aware-similarity}), similarity-aware neighbor selector (Section~\ref{sec:similarity-aware neighbor-selector}), and relation-aware neighbor aggregator (Section~\ref{sec:relation-aware neighbor-aggregator}).
In addition, we describe the overall algorithm and optimization in Section~\ref{sec:Proposed RemMrGNN}.

\subsection{Label-aware Neural Similarity Measure}
\label{sec:label-aware-similarity}
Compared with unsupervised similarity metrics like Cosine Similarity~\cite{yilmaz2020unsupervised} or Neural Networks~\cite{yilmaz2020unsupervised}, many practical problems like financial fraud, disease diagnosis, etc., require extra domain knowledge (e.g., high-fidelity data annotations) to identify anomaly instances.
To this end, we design a parameterized node similarity measure, i.e., label-aware neural similarity measure, using supervision signals from domain experts. 
AGCN~\cite{li2018adaptive} employs a Mahalanobis distance plus a Gaussian kernel, while DIAL-GNN~\cite{chen2019deep} adopts the parameterized cosine similarity.
NSN~\cite{liu2019neural} unitizes bilinear similarity based inner product and hyperspherical learning strategies. 
However, all of them have non-negligible time complexity $O(\overline{k}d)$, where $\bar{k}$ denotes the average degree of nodes, often very high in real-world graphs, and $d$ represents the feature dimension.
This leads to a loss of efficiency when discriminating the node representation learning.

Inspired by GraphMix~\cite{verma2019graphmix}, with a combination of Fully-Connected Network (FCN) and linear regularization, we adopt an FCN as the node label predictor at each layer of \RioGNN, and use the $l_1$-distance between the prediction results of two nodes as the measure of the in-between similarity.
It is a simple and efficient regularizer for semi-supervised node classification using GNNs.
At the $l$-th layer, when calculating the distance between one intermediate node $v$ and one of its neighbors $v^{\prime}$ under relation $r$, i.e., the edge $(v, v^{\prime})\in \mathcal{E}_{r}$, we take their embedding in the previous layer $\mathbf{h^{(l-1)}}$ as input, and apply the non-linear activation function $\sigma$ (we use $tanh$ in our work).
The distance between $v$ and $v^{\prime}$ is the $l_1$-distance of two embeddings:
\begin{equation}\label{eq:distance}
\mathcal{D}^{(l)}(v,v^{\prime})=||\sigma(FCN^{(l)}\mathbf{h}_{v}^{(l-1)})-\sigma(FCN^{(l)}\mathbf{h}_{v^{\prime}}^{(l-1)})||_{1}.
\end{equation}
\noindent Thus, the similarity of the two nodes can be defined as:
\begin{equation}\label{eq:similarity}
S^{(l)}(v,v^{\prime})=1-\mathcal{D}^{(l)}(v, v^{\prime}).
\end{equation}

The time complexity of our approach can be reduced from $O(\overline{k}d)$ to $O(d)$.
In general, the computational cost is low because for each node in the node set $\mathcal{V}$, we do not use the combined embedding of its $k$ neighbors with $d$-dimensional features to measure similarity like LAGCN~\cite{chen2020label}, but only consider the label predicted by FCN based on its own feature.

To train similarity measure with a direct supervision signal from the labels, we define the cross entropy loss of the FCN in the $l$-layer as:
\begin{equation}\label{eq:sim-loss}
\mathcal{L}_{Simi}^{(l)}=\sum_{v \in \mathcal{V}}-log(y_{v}\cdot\sigma(FCN^{(l)}(\mathbf{h}_{v}^{(l)}))).
\end{equation}
Further, we define the cross entropy loss of label-aware similarity measure for the entire network as:
\begin{equation}\label{eq:sim-loss-all}
\mathcal{L}_{Simi}=\sum_{l=1}^{L} \mathcal{L}_{Simi}^{(l)}.
\end{equation}
During the training process, the similarity measure parameters of FCNs are directly updated through the loss function. 
It ensures that similar neighbors can be quickly selected within a few batches and facilitates to regularize the GNN training process.

In this subsection, we propose a label-aware similarity detection method for the first challenge in Sec.~\ref{sec:challenges}. 
This method is based on node labels to effectively avoid interference caused by bad node camouflage in actual scenes, and reduces the complexity of similarity, which provides a stable basis for subsequent neighbor filtering.

\subsection{Similarity-aware Adaptive Neighbor Selector}
\label{sec:similarity-aware neighbor-selector}

To select appropriate neighbors adaptively, we design a similarity-aware neighbor selector to filter misbehaved nodes stemming from adversarial behaviors or inaccurate feature extraction.
More specifically, for each central node, the selector utilizes \emph{Top-p Sampling} along with adaptive filtering thresholds to construct similar neighbors under each relation.
Since the filter thresholds for different relations at different layers tend to be dynamically updated during the training phase, we propose \RSRL, a recursive and scalable reinforcement learning framework to optimize the filtering threshold for each relation in an efficient manner.

\subsubsection{Top-p Sampling}
\label{sec:top-p-sampling}

Before aggregating the information from both central node $v$ and its neighborhood, we perform a \emph{Top-p sampling} to filter the dissimilar neighbors according to different relations. 
A filtering threshold $p_{r}^{l}\in [0, 1]$ for relation $r$ at the $l$-th layer indicates the selection ratio from all neighbors.
For instance, all neighbor nodes under the relation $r$ are retained if $p_{r}$ is 1. 
More specifically, during the training phase, for a node $v$ in one batch under the relation $r$, we first calculate a set of similarity scores $\{S^{l} (v, v^{\prime})\}$ by using Eq.~\ref{eq:similarity} at $l$-th layer where the edge $(v, v^{\prime}) \in \mathcal{E}^{l}_{r}$.
We then rank the neighbors of each central node $v$ in descending order, based on $\{S^{l} (v, v^{\prime})\}$, and take the top $p_{r}^{l}\cdot |\{S^{l} (v, v^{\prime})\}|$ neighbors as the selected ones, i.e., $\mathcal{N}_{r}^{l}(v)$, at the $l$-th layer.
The residual nodes will be discarded at the current batch and not attend the following aggregation process within the layer.

\begin{figure}[h]
  \centering
  \includegraphics[width=\linewidth]{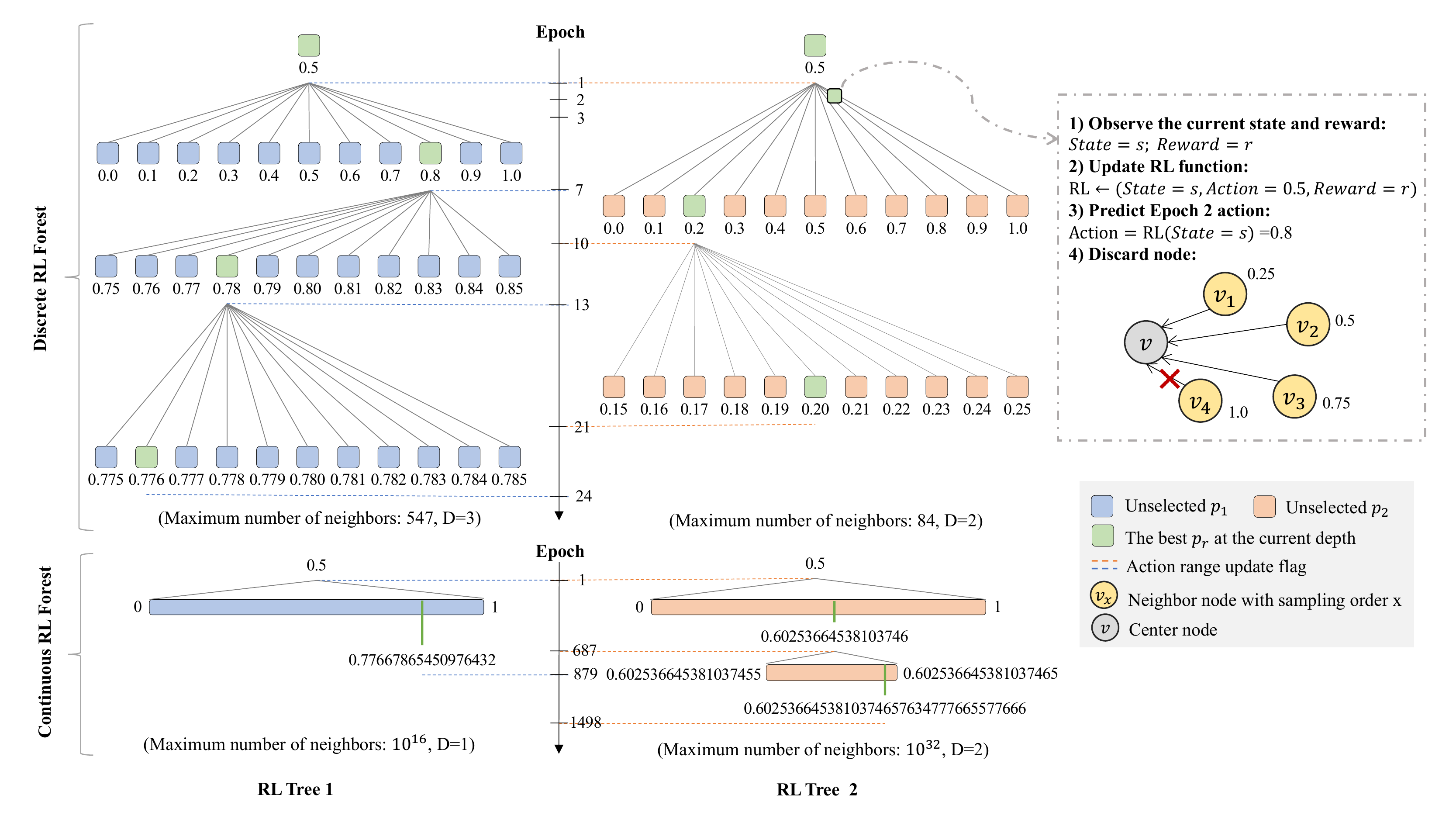}
  \caption{One layer Reinforcement Learning Forest.}
  \label{fig:RL-Framework}
\end{figure}

\subsubsection{\RSRL Framework}\label{sec:rsrl}

Previous work~\cite{Chen2019iterative,liu2020alleviating} regards the filtering threshold as a hyper-parameter, which is no longer valid for multi-relational graph with numerous noisy or misbehaved nodes.
To solve this, our preliminary work~\cite{dou2020enhancing} adopted the Bernoulli multi-armed bandit framework~\cite{sutton2018reinforcement} with a fixed learning strategy and dynamically updated the filtering threshold.
However, the effectiveness of this approach is largely impeded by the limited observation range of states and the manually-specified strategy.
As a result, the final convergence outcome of the filtering threshold tends to be local optimal.
In the face of larger-scale datasets, the maintenance of the prediction accuracy also needs to reduce the adjustment step size of the filtering threshold.
This process will increase the number of convergence epochs, and bring in an increase in the amount of calculation and a loss of accuracy.

\textbf{Overview}. To address these problems, we propose a novel Recursive and Scalable Reinforcement Learning framework \RSRL, upon traditional Reinforcement Learning based approaches~\cite{Feng2018rlclassification,Chen2019rldialogue,Lai2020policy}, to not only update strategies through the learning environment but also the recursive structure can be used to quickly and accurately meet the accuracy requirements of different relations. 
Figure.~\ref{fig:RL-Framework} depicts the forest-based learning architecture. 
The specific process of a tree in each epoch is shown on the right, where $s$ and $r$ respectively represent the state and reward after the previous epoch, and $a$ (the example in the figure is 0.5) represents the predicted action in the current epoch.
We formulate RSRL as an $L$-layer Reinforcement Learning (RL) Forest, and define the $l$-th layer forest as:
\begin{equation}\label{eq:RLF}
RLF^{(l)}=\{RLT_r^{(l)}\}|_{r=1}^{R}=\{\{RL_r^{(l)(d)}\}|_{d=1}^{D_r^{(l)}}\}|_{r=1}^{R},
\end{equation}
$RLF^{(l)}$ actually indicates the process of obtaining the best relational filtering threshold combination at the $l$-th layer. 
Each relation independently constructs a RL Tree $RLT_{r}^{(l)}$ with an adaptive depth $D_{r}^{(l)}=\lceil \log_{\alpha}k_r \rceil$ and a width $W_{r}^{(l)(d)}=\frac{1}{\alpha^d}$. $\alpha$ is the weight parameter of depth first and breadth first, and $k_r$ is the maximum number of neighbors contained in the node in relation $r$. 
$RLT_{r}^{(l)}$ performs a Reinforcement Learning $RL_{r}^{(l)(d)}$ for filtering the threshold with an accuracy $W_{r}^{(l)(d)}$ at each depth.
At the $l$-th layer, $RLT_{r}^{(l)}$ acquires the best filtering threshold $p_{r}^{(l)(d)}$ of neighbor nodes with higher accuracy than the previous depth of relation $r$ through multiple RL recursively, until the threshold for maximum accuracy requirements is found at the depth $D_{r}^{(l)}$. 

The $RLT_{r}^{(l)}$ recursive process is expressed as:
\begin{equation}\label{eq:recursive-process}
p_{r}^{(l)(d)} \xleftarrow[]{RL_{r}^{(l)(d)}}  \{p_{r}^{(l)(d-1)}-\frac{W_{r}^{(l)(d)}}{2},p_{r}^{(l)(d-1)}+\frac{W_{r}^{(l)(d)}}{2}\}.
\end{equation}
where $p_{r}^{(l)(d)}$ represents the optimal proportion of neighbor nodes in the relation $r$ to be discarded when the depth of the RL tree is $d$ in the $l$-th layer.
The learning range of the RL module of each depth is the value within $\pm\frac{W_{r}^{(l)}}{2}$ of the filter threshold $p_{r}^{(l)(d-1)}$ selected by the previous depth.
When the recursive process reaches the maximum depth $D_r^{(l)}$, we obtain the final filtering threshold $p_{r}^{(l)}$ of the relation $r$ in the $l$-th layer.
Considering that the complexity has a linear relationship with the size of the action space~\cite{dulacarnold2016d2c}, this process carries out a precision recursion on RL actions and can reduce the time from $O(k_{r})$ to $O({\alpha}\log_{\alpha}k_{r})$, where $\alpha \in (1,k_{r})$ and $k_{r}$ is the maximum node degree under relation $r$.

\textbf{Details of RL Process.}
We express an RL module as a Markov Decision Process $MDP<A, S, R, F>$ for the filtering threshold of a relation.
$A$ and $S$ are the action space and state space, respectively;
$R$ is the reward function, and $F$ is the iteration functions and termination conditions. 
To better deal with datasets with different sizes and diverse scenarios, we break down our solution into two distinct categories: Discrete Reinforcement Learning (D-RL) and Continuous Reinforcement Learning (C-RL). 

\begin{itemize}
\item \textbf{Action:} We define the action space $A$ of $RL_{r}^{(l)(d)}$ by collecting all actions $a_{r}^{(l)(d)}\in \{p_{r}^{(l)(d-1)}-\frac{W_{r}^{(l)(d)}}{2},p_{r}^{(l)(d-1)}+\frac{W_{r}^{(l)(d)}}{2}\}$ when the relation $r$ is at the depth $d$ of the $l$-th layer. 
The discrete scheme D-RL divides the action space into $\alpha$ discrete actions on average.
In the continuous scheme C-RL, the action space in $d$-th depth is a continuous floating point number with the width of $W_{r}^{(l)(d)}$.
The compatibility of the two types of action space can effectively adapt to a variety of reinforcement learning algorithms.
Due to the low precision requirements of the filtering threshold for small and medium-scale datasets, discretization of actions can reduce the number of action explorations~\cite{Anssi2020c2d} while meeting the basic accuracy requirements, which ensures efficient access to high-performance areas.
For the dataset with large-scale neighbors, a large number of discrete actions that meet the high-precision requirements will affect the learning effect~\cite{dulacarnold2016d2c}. We propose to generalize the filtering threshold to the continuous action space to improve the accuracy of large-scale data sets by reducing the spatial range multiple times.

\item \textbf{State:} Since it is impossible to directly perceive the classification loss of GNN as the environment state, we calculate the average node distance of each epoch as the state through the distance measure of label perception (Eq. (\ref{eq:distance})).
In the $l$-th layer of the $e$-th epoch, the state $s$ of relation $r$ in the $d$-th depth is:
\begin{equation}\label{eq:state-function}
s_{r}^{(l)(d)(e)}= \frac{\sum_{(v, v^{\prime}) \in \mathcal{E}_{r}^{(l)(d)(e)}}\mathcal{D}^{(l)}(v,v^{\prime})^{(e)}}{|\mathcal{E}_{r}^{(l)(d)(e)}|}, 
\end{equation}
where $\mathcal{E}_{r}^{(l)(d)(e)}$ is the set of edges filtered in the $l$-th layer and $d$-th depth of the $e$-th epoch under relation $r$.  
\item \textbf{Reward:}
For each relation, the goal is to ascertain a filtering threshold $p_{r}^{(l)(d)}$ so that the selected neighbor node and the central node are as close as possible.
We therefore use the similarity (Eq. (\ref{eq:similarity})) as the decisive factor within the reward function.
In the $l$-th layer of the $e$-th epoch, the reward $g$ of relation $r$ to $d$-th depth is:
\begin{equation}
\label{eq:reward-function}
g_{r}^{(l)(d)(e)}=\tau \cdot(\frac{\sum_{(v, v^{\prime}) \in \mathcal{E}_{r}^{(l)(d)(e)}}\mathcal{S}^{(l)}(v,v^{\prime})^{(e)}}{|\mathcal{E}_{r}^{(l)(d)(e)}|}),
\end{equation}
where $\tau$ is the weight parameter, and the meaning of $\mathcal{E}_{r}^{(l)(d)(e)}$ is the same as the definition in the state.

\item \textbf{Iteration and Termination:} 
Before starting each epoch, RL observes the state $s_{r}^{(l)(d)(e)}$ of the environment after the previous epoch of action $a_{r}^{(l)(d)(e-1)}$ and obtains a reward $g_{r}^{(l)(d)(e-1)}$.
They are then used to update the iterative function (broadly refers to the function of strategy iteration or value iteration process of reinforcement learning) $f$.
The iterative function of each $RL_{r}^{(l)(d)}$ is as follows:
\begin{equation}\label{eq:update-iterative-function}
f_{r}^{(l)(d)} \leftarrow s_{r}^{(l)(d)(e-1)}, s_{r}^{(l)(d)(e)}, a_{r}^{(l)(d)(e-1)}, g_{r}^{(l)(d)(e-1)}.
\end{equation}
Then we can use the iterative function to predict the action $a$ from the current state $s$, which is the filtering threshold:
\begin{equation}\label{eq:iterative-function}
p_{r}^{(l)(d)(e)}=a_{r}^{(l)(d)(e)}=f_{r}^{(l)(d)}(s_{r}^{(l)(d)(e)}).
\end{equation}
Here, for the output of the action, the activation function of the classification type is used to represent them in D-RL, such as softmax.
And in C-RL, we use the activation function of the return value type to represent them, such as $tanh$. 
Since what we propose is a general framework applicable to a variety of reinforcement learning algorithms, the specific definition of the iterative function depends on the actual algorithm.
We test the applicability of the \RSRL framework to various mainstream reinforcement learning algorithms in Sec.~\ref{sec:versatility}.
To improve the equalization efficiency, we assume the RL will be terminated as long as the same action appears three times in a row at the current accuracy $W_{r}^{(l)(d)}$. 
Specifically, in the $l$-th layer of the $e$-th epoch, the termination conditions of relation $r$ in the $d$-th depth are defined as follows:
\begin{equation}
\label{eq:termination-conditions}
\left\{
             \begin{array}{lr}
             \{p_{r}^{(l)(d)(i)}-p_{r}^{(l)(d)(i-1)}=0\}|_{i=e-1}^{e} & D-Rl, where\ e > 2.\\
             \\
             \{|p_{r}^{(l)(d)(i)}-p_{r}^{(l)(d)(i-1)}|<W_{r}^{(l)(d)}\}|_{i=e-1}^{e} & C-RL, where\ e > 2. 
             \end{array}
\right.
\end{equation}
We formally define this deep switching condition or termination condition as $deep\ switching\ number = 3$, and discuss parameter sensitivity in Section~\ref{sec:hyper-parameter}.
\end{itemize}

To acquire better training results, we use white-box methodology to test the results synchronously during the training process, and verify whether the convergence value is optimal for this round before starting a new round of RL for the same relation.
If the value is negative, the filtering threshold with better performance in the historical version will be reviewed in the new round of optimization as the basis for the new round of action range. 
For this backtracking mechanism, we will conduct a sensitivity experiment in Section~\ref{sec:hyper-parameter}.

In this subsection, we overcome the second and third challenges mentioned in Sec.~\ref{sec:challenges}. Specifically, we use the similarity measure in the previous subsection to perform Top-p sampling for neighbors of each relation.
Based on reinforcement learning, we use the agent to interact with the environment to make different relations to obtain different threshold combinations. 
This adaptive method is free from the help of data annotation.
Furthermore, in order to meet the accuracy requirements of different relations while ensuring accuracy, we propose a recursive framework for optimization.

\subsection{Relation-aware Weighted Neighbor Aggregator}\label{sec:relation-aware neighbor-aggregator}
Based on the selection of similar neighbors for each relation, the next step is to aggregate all these neighbor information among relations, for a comprehensive embedding. 
Previous methods employ attention mechanisms~\cite{liu2019b_geniepath,wang2019semi,gong2020attentional,hao2020pre} or weighting parameters~\cite{liu2018heterogeneous} to learn the relation weights during the aggregating procedure. 
To reduce the computational cost whilst retaining the relation importance information, we directly use the optimal filtering threshold $p^{(l)}_r$ learned by the \RSRL process as the inter-relation aggregation weights.
Formally, for central node $v$, under relation $r$ at the $l$-th layer, the intra-relation neighbor aggregation can be defined as follows:
\begin{equation}\label{eq:intra-relation-aggregation}
\textbf{h}_{v, r}^{(l)}=ReLU(AGG_{r}^{(l)}(\{\oplus\ \textbf{h}_{v'}^{(l-1)}: v'\in \mathcal{N}_{r}^{l}(v)\})),
\end{equation}
where $\oplus$ denotes the embedding bitwise summation operation for mean aggregator $AGG_{r}^{(l)}$, and $\mathcal{N}_{r}^{l}(v)$ refers to the set of top $p_r^{(l)}$ nodes obtained by Eq. (\ref{eq:update-iterative-function}) under relation $r$ at the $l$-th layer.
The purpose of the intra-relation neighbor aggregation for the central node $v$ is to aggregate all neighborhood information under the relation $r$ at the previous layer into the embedding vector ${h}_{v, r}^{(l)}$.

To follow up, we define inter-relation aggregation as below: 
\begin{equation}
\label{eq:inter-relation-aggregation}
\textbf{h}_{v}^{(l)}=ReLU(\textbf{h}_{v}^{(l-1)} \oplus AGG^{(l)}(\{\oplus\ (p_r^{(l)} \cdot \textbf{h}_{v, r}^{(l)})\}|_{r=1}^R)),
\end{equation}
where ${h}_{v, r}^{(l)}$ indicates the intra-relation neighbor embedding at the $l$-th layer and $AGG^{(l)}$ can be any type of aggregator. 
Here we directly use the $p_r^{(l)}$ optimized by the RSRL framework as the aggregation weight, and conduct experiments on other types of aggregation methods in Section~\ref{sec:fraud-accuracy} and Section~\ref{sec:diabetes-accuracy}.

In this subsection, in order to better deal with improper nodes to respond to Challenge $1$ in Sec.~\ref{sec:challenges}, we divide the aggregation process into inter-relation and intra-relation, and use filtered neighbor nodes and filter thresholds of different relations to strengthen the influence of benign nodes during aggregation.

\begin{algorithm}[htbp]
		  \caption{\textbf{\textsc{Rem}GNN:} Reinforced Neighborhood Selection Guided Multiple Relation GNN.}\label{alg:rio-gnn}
		\SetKwInOut{Input}{Require}\SetKwInOut{Output}{Ensure}
		\Input{An undirected multi-relational graph with node features and labels: $\mathcal{G}=\left\{\mathcal{V}, \mathbf{X}, \{\mathcal{E}_{r}\}|_{r=1}^{R}, Y\right\}$;\\
		Number of layers, batches, epochs: $L, B, E$; Parameterized similarity measures: $\{S^{(l)}(\cdot, \cdot)\}|_{l=1}^{L}$;\\
		Filtering thresholds: $P=\{p^{(l)}_{1}, \dots, p^{(l)}_{R}\}|_{l=1}^{L}$; Intra-R aggregators: $\{\textnormal{AGG}^{(l)}\}|_{r=1}^{R}, \forall l\in\{1,\dots,L\}$;\\
		Inter-R aggregators: $\{\textnormal{AGG}_{r}^{(l)}\}, \forall l\in\{1,\dots,L\}$;\\
		RL Module: $\{f_{r}^{(l)(d)}|_{d=1}^{D_{r}^{l}}\}, \forall r\in\{1,\dots,R\}, \forall l\in\{1,\dots,L\}$.
	}
		\Output{Vector representations $\mathbf{z}_{v}, \forall v \in \mathcal{V}_{train}$.}
		\BlankLine
		\tcp{Initialization}
		 $\mathbf{h}_{v}^{0} \leftarrow \mathbf{x}_{v}, \forall v\in \mathcal{V};$ $p_{r}^{(0)}=0.5, d_{r}^{(0)}=0, \mathcal{E}_{r}^{(0)} = \mathcal{E}, \forall r\in\{1,\dots,R\}$
		
		$\{p_{r}^{(l)(0)} \in [0,1]\}, \forall r\in\{1,\dots,R\}, \forall l\in\{1,\dots,L\}$
		
		\tcp{Training process of the proposed \textsc{Rio}GNN}
		
		\For{$e = 1, \cdots, E$}{
			\For{$b = 1, \cdots, B$}{
				\For{$l = 1, \cdots, L$}{
					$\mathcal{L}^{(l)}_{\textnormal{Simi}} \leftarrow$ Eq.~(\ref{eq:sim-loss}) \tcp{Cross entropy loss of label-aware similarity measure} \label{al:sim-loss}
					\For{$r = 1, \cdots, R$}{
					 $S^{(l)}(v, v^{\prime}) \leftarrow$ Eq.~(\ref{eq:similarity}) , $\forall (v, v^{\prime})\in \mathcal{E}_{r}^{(l-1)}$\; \label{al:sim}
					\tcp{Edge set under relation r at the l-th layer}
					$\mathcal{E}_{r}^{(l)} \leftarrow$ \textit{top-p} sampling (Section~\ref{sec:top-p-sampling})\; 
					$\mathbf{h}_{v,r}^{(l)} \leftarrow$ Eq.~(\ref{eq:intra-relation-aggregation}) $\forall v \in \mathcal{V}_{b}$ \tcp{Intra-relation aggregator} 	\label{al:agg-start}
					}
					$\mathbf{h}_{v}^{(l)} \leftarrow$ Eq.~(\ref{eq:inter-relation-aggregation}) $\forall v \in \mathcal{V}_{b}$; \tcp{Inter-relation aggregator}
				}
				
				$\mathbf{z}_{v} \leftarrow \mathbf{h}_{v}^{(L)}, \forall v \in \mathcal{V}_{b}$; \label{al:agg-end} \tcp{Batch node embeddings} 
				
				$\mathcal{L}_{\textnormal{GNN}} \leftarrow$ Eq.~(\ref{eq:gnn}); \tcp{Cross-entropy loss function of GNN} \label{al:gnn-loss}
				
				$\mathcal{L}_{\textnormal{\textsc{Rio}GNN}} \leftarrow$ Eq.~(\ref{eq:riognn}); \tcp{Final loss function of \textsc{Rio}GNN}
			} 
		 \tcp{RSRL Module: Markov Decision Process for filtering threshold}
	\For{$l = 1, \cdots, L$}{
		\For{$r = 1, \cdots, R$}{
			\If{$d_{r}^{(l)} < D{r}^{(l)}$}{ \label{al:rsrl-start}
			\tcp{Judgement of the termination condition}
				\If{\textnormal{Eq.~(\ref{eq:termination-conditions}) is False}} {
					$s_{r}^{(l)(d)(e)}, g_{r}^{(l)(d)(e-1)}  \leftarrow$ Eq. (\ref{eq:state-function}) and Eq. (\ref{eq:reward-function}) \tcp{Calculate state and reward} 
					$f_{r}^{(l)(d)(e)} \leftarrow$ Eq. (\ref{eq:update-iterative-function}); \tcp{Update RL iterative function}
					\tcp{the recursive process of optimal proportion of neighbor nodes}
					$p_{r}^{(l)(d)(e)} \leftarrow$ Eq. (\ref{eq:iterative-function}), $p_{r}^{(l)(d)}\in \{p_{r}^{(l)(d-1)}-\frac{W_{r}^{(l)(d)}}{2},p_{r}^{(l)(d-1)}+\frac{W_{r}^{(l)(d)}}{2}\}$
				}
				\Else{
					$d_{r}^{(l)}=d_{r}^{(l)}+1$;\tcp{Update the depth of Reinforcement Learning Tree} \label{al:rsrl-end}
				}
			}
		}    
	}
}
		
	\end{algorithm}

\subsection{Put Them Together}
\label{sec:Proposed RemMrGNN}

We denote the final embedding of node $v$ as $\mathbf{z}_{v}=\mathbf{h}_v^{(L)}$, which is the output of \RioGNN at the last layer.
We also define the loss function of GNN in node classification task as the cross-entropy loss function:
\begin{equation}\label{eq:gnn}
\mathcal{L}_{GNN}=\sum_{v \in \mathcal{V}}-log(y_{v}\cdot\sigma(MLP^{(l)}(\mathbf{z}_{v}))).
\end{equation}
Together with the loss function of node classification and the loss function of the similarity measure in Eq.~\ref{eq:sim-loss}, we define the final loss function of \RioGNN as follow:
\begin{equation}\label{eq:riognn}
\mathcal{L}_{\RioGNN}=\mathcal{L}_{GNN}+\lambda_{l}\sum_{l=1}^{L}\mathcal{L}_{Simi}^{(l)}+\lambda_{*}||\Theta||_{2},
\end{equation}
where $\lambda_{l}$ and $\lambda_{*}$ are the weight parameters, and $||\Theta||_{2}$ is the $L2$-norm of all model parameters.

Finally, based on the aforementioned study, Algorithm~\ref{alg:rio-gnn} outlines the training process of the proposed \RioGNN which takes any given input multiple relation graph built upon a real-world practical problem.
We employ the mini-batch training technique~\cite{goyal2017accurate} to cope with excessively large real-world graphs.  
We initialize the parameters of the label-aware similarity module, relation-aware neighbor select module, and relation-aware neighbor aggregator module, before training the \RioGNN model at each epoch. 
For each batch of nodes, we first compute the neighbor similarities using Eq.~(\ref{eq:sim-loss}) and then leverage the  \emph{top-p} sampling to filter the neighbors. 
Thereafter, we compute the intra-relation embeddings (by using Eq.~(\ref{eq:intra-relation-aggregation})) and inter-relation embeddings (by using Eq.~(\ref{eq:inter-relation-aggregation})), and define the loss functions (by using Eq.~(\ref{eq:riognn})) for the current batch.
In the \RSRL module, we allocate $H$ RL modules in sequence according to the maximum depth for each layer of each relation.
In each RL module, each epoch will observe the environment state via Eq.~(\ref{eq:state-function}) and get the reward via Eq.~(\ref{eq:reward-function}).
Then the algorithm updates the iterative function through Eq.~(\ref{eq:update-iterative-function}) and predict the filtering threshold via Eq.~(\ref{eq:iterative-function}) of the current epoch through the updated iterative function.
When an RL module reaches the convergence condition through Eq.~(\ref{eq:termination-conditions}) without targeting the maximum depth, we will recursively proceed to the next depth RL until all RL modules complete.

\textbf{Time Complexity of \RioGNN.}
The overall time complexity of Algorithm~\ref{alg:rio-gnn} is $O(|\mathcal{E}|\cdot\max(\{{\alpha}\log_{\alpha}k_{r}\}|^{R}_{r=1}))$, where $|\mathcal{E}|$ is the number of edges, $\alpha$ is the weight parameter of depth first and breadth first, and $k_r$ is the maximum number of neighbors contained in the node in relation $r$. 
Here, $O(|\mathcal{E}|)$ is the time complexity in one epoch.
Specifically, the similarity measure (Line~\ref{al:sim} in Algorithm~\ref{alg:rio-gnn}) and aggregation (Line~\ref{al:agg-start}-\ref{al:agg-end} in Algorithm~\ref{alg:rio-gnn}) take a total of $O(|\mathcal{E}|d+|\mathcal{V}|(\overline{k}d+d))=O(|\mathcal{E}|)$, where $\overline{k}$ is the average node degree and $|\mathcal{V}|$ is the number of nodes.
The RSRL module (Line~\ref{al:rsrl-start}-\ref{al:rsrl-end} in Algorithm~\ref{alg:rio-gnn}) takes $O(|\mathcal{E}|d)=O(|\mathcal{E}|)$.
Each cross-entropy loss function (Line~\ref{al:sim-loss} and Line~\ref{al:gnn-loss} in Algorithm~\ref{alg:rio-gnn}) takes $O(|V|)$.
In addition, the number of epochs is affected by the action space of reinforcement learning, and the number of epochs required to achieve convergence is related to the action space with the greatest demand among several relations $\max(\{{\alpha}\log_{\alpha}k_{r}\}|^{R}_{r=1})$.

%% file: 4-Experiment.tex
\section{Experimental Setup}
\label{sec:experimental-setup}

In the following two sections, we conduct experiments to evaluate and test \RioGNN. 
The experimental setup mainly revolves around the following six questions:
\begin{itemize}
\item \textbf{Q1:} How do we build multi-relational graphs in different scenarios (Section~\ref{sec:datasets})?
\item \textbf{Q2:} The effectiveness, efficiency and explainability of \RioGNN in fraud detection tasks (Section~\ref{sec:fraud-overall}).
\item \textbf{Q3:} The effectiveness, efficiency and explainability of \RioGNN in disease detection tasks (Section~\ref{sec:diabetes-overall}).
\item \textbf{Q4:} How do different task requirements match the universal RSRL framework (Section~\ref{sec:versatility})?
\item \textbf{Q5:} How does our model perform in clustering tasks and inductive learning (Section~\ref{sec:diabetes-accuracy}, Section~\ref{sec:fraud-accuracy} and Section~\ref{sec:inductive})?
\item \textbf{Q6:} Discussion of hyper-parameter sensitivity and effects on the model (Section~\ref{sec:hyper-parameter}).
\end{itemize}

\subsection{Experimental Settings}\label{sec:hardware-software}
We implement \RioGNN with Pytorch. 
All experiments are running on Python 3.7.1, and a NVIDIA V100 NVLINK GPU with 32GB RAM. 
The operating system is Ubuntu 20.04.2.
To improve the training efficiency and avoid overfitting, we employ mini-batch training and under-sampling techniques to train \RioGNN and other baselines. 
Specifically, under each mini-batch, we randomly sample the same number of negative instances as the number of positive instances.

\subsection{Datasets and Graph Construction}\label{sec:datasets}
We build different multi-relational graphs for experiments in \textit{two} task scenarios and \textit{three} datasets. 
Table~\ref{tab:dataset} lists various statistical information of different dataset nodes and relationships. 
In addition to the number of nodes and the proportion of noisy nodes (Fraud\%, Diabetes\%) in different scenarios, we also give the number of edges with different relations. 
For each relationship in each dataset, we calculate the feature similarity of adjacent nodes based on the Euclidean distance (range 0 to 1) of the feature vector of adjacent nodes, and normalize the average feature similarity. 
The last column of Table~\ref{tab:dataset} shows the average label similarity of each relationship, which is calculated based on whether two connected nodes have the same label.

\begin{table}[h]
    \setlength{\abovecaptionskip}{0.cm}
    \setlength{\belowcaptionskip}{-0.cm}
    \caption{Dataset and graph statistics.}\label{tab:dataset}
    \centering
    \scalebox{0.95}{
        \begin{tabular}{c|ccccc}
        \hline
        \multicolumn{1}{c|}{\multirow{2}*{Dataset}}&\multicolumn{1}{c}{\#Nodes}&\multirow{2}*{Relation}&\multirow{2}*{\#Edges}&\multicolumn{1}{c}{Avg. Feature}&\multicolumn{1}{c}{Avg. Label}\\
        \multicolumn{1}{c|}{}&\multicolumn{1}{c}{(Fraud\% / Diabetes\%)}&\multicolumn{1}{c}{}&\multicolumn{1}{c}{}&Similarity&Similarity\\
        \hline
        \multirow{4}*{Yelp}&\multicolumn{1}{c}{}&R-U-R&49,315&0.83&0.90\\
        \multicolumn{1}{c|}{}&45,954&R-T-R&573,616&0.79&0.05\\
        \multicolumn{1}{c|}{}&(14.5\%)&R-S-R&3,402,743&0.77&0.05\\
        \multicolumn{1}{c|}{}&\multicolumn{1}{c}{}&ALL&3,846,979&0.77&0.07\\
        \hline
        \multirow{4}*{Amazon}&\multicolumn{1}{c}{}&U-P-U&175,608&0.61&0.19\\
        \multicolumn{1}{c|}{}&11,944&U-S-U&3,566,479&0.64&0.04\\
        \multicolumn{1}{c|}{}&(9.5\%)&U-V-U&1,036,737&0.71&0.03\\
        \multicolumn{1}{c|}{}&\multicolumn{1}{c}{}&ALL&4,398,392&0.65&0.05\\
        \hline
        \hline
        \multirow{5}*{MIMIC-III}&\multicolumn{1}{c}{}&V-A-V&152,901,492&0.62&0.54 \\
        \multicolumn{1}{c|}{}&28,522&V-D-V&19,183,922&0.63&0.54 \\
        \multicolumn{1}{c|}{}&(49.9\%)&V-P-V&149,757,030&0.63&0.54 \\
        \multicolumn{1}{c|}{}&\multicolumn{1}{c}{}&V-M-V&15,794,101&0.65&0.51 \\
        \multicolumn{1}{c|}{}&\multicolumn{1}{c}{}&ALL&337,636,545&0.63&0.53 \\
        \hline
        \end{tabular}
    }
\end{table}

\subsubsection{Fraud Detection Task}\label{sec:fraud-detection-task}
We perform binary classification tasks, spam review detection and fraudulent user detection on the Yelp dataset and Amazon dataset, respectively.

\textbf{Yelp:} is collected from the internal dataset published by Yelp.com, the largest business reviewing site in the United States.
We use a subset of the YelpChi dataset collected and used by~\cite{mukherjee2013yelp}.
This subset contains 45,954 user reviews of hotels and restaurants in the Chicago area.
The reviews have been filtered (spam) and recommended (legal) by Yelp.
In addition to containing information about the relations between users and products, the dataset also contains various metadata, including the text content of the review, timestamp, and star rating.
We use 32 manual features including \emph{the ranking order of all product reviews}, \emph{the absolute rating deviation from the product average}, \emph{whether it is the only review of the user}, \emph{the percentage of all uppercase words}, \emph{the percentage of uppercase letters}, \emph{the length of the review}, \emph{the ratio of first-person pronouns}, \emph{the ratio of exclamatory sentences}, and \emph{subjective Word ratio}, \emph{ratio of objective words}, etc. used in~\cite{rayana2015collective} as the original node features of the Yelp dataset.
For specific relations, since previous research ~\cite{mukherjee2013yelp,rayana2015collective} has shown opinion fraudsters (i.e., spammers) are connected in terms of users, products, review text, and time, we use reviews as nodes in the graph and design three relations:
\begin{itemize}
\item R-U-R: it connects reviews posted by the same user.
\item R-T-R: it connects two reviews under the same product posted in the same month.
\item R-S-R: it connects reviews under the same product with the same star rating (1-5 stars).
\end{itemize}

\textbf{Amazon:} is a subset of Amazon's product dataset~\cite{mcauley2013amazon}.
The Amazon dataset contains more than 34,000 consumer reviews, from which we extracted 11,949 product reviews under the musical instrument category.
In addition, similar to ~\cite{zhang2020amazonfeatures}, we mark users with useful voting rates greater than 80\% as benign entities, and users with useful voting rates less than 20\% as fraudulent entities.
In terms of node feature selection, we use 25 manual features including \emph{the number of rated products}, \emph{the length of the user name}, \emph{the number and ratio of each rating level given by the user}, \emph{the ratio of positive and negative reviews, the user's rating}, \emph{the total number of useful and useless votes obtained by the user}, \emph{the ratio of useful votes and useless votes}, and \emph{Average value}, \emph{median of useful and useless votes}, \emph{minimum and maximum number of useful and useless votes}, \emph{number of days between the user‘s first and last rating}, \emph{same date indicator}, \emph{comment text sentiment}, etc. used in~\cite{zhang2020amazonfeatures} as the original node function of the Amazon data set.
For specific relations, we design three kinds of relations for the multi-relational graph, as shown below:
\begin{itemize}
\item U-P-U: it connects users reviewing at least one same product. 
\item U-S-V: it connects users having at least one same star rating within one week.
\item U-V-U: it connects users with the top 5\% mutual review text similarities (measured by TF-IDF) among all users.
\end{itemize}

\subsubsection{Diagnosis of diabetes mellitus task}
\label{sec:diabetes-detection-task}
We perform the binary classification task of diagnosis of diabetes mellitus on the processed MIMIC-III dataset resources.

\textbf{MIMIC-III:} is a publicly available dataset ~\cite{2000PhysioBank,0MIMIC} consisting of health records of 46,520 intensive care unit (ICU) patients over 11 years.
We have extracted a total of 28,522 patient visits, and each record contains information such as age, diagnosis, microbiology, procedures, corpus, etc.
We use a subset of the MIMIC-III dataset collected and used by~\cite{liu2020health}, and construct our multi-relational graph for our task based on that. 
For each patient and visit, there is a unique ID to track its corresponding information.
According to the diagnostic codes, we mark the medical records as diabetic or non-diabetic.
Ages are split into groups using threshold 15, 30 and 64 as suggested in ~\cite{info:doi/10.2196/medinform.6437}. 
Procedures and diagnoses are mapped into corresponding ICD-9-CM codes.
Microbiology tests with culture-positive results are mapped into the names of organisms.
We use the above four fields to form different relationships among visit nodes to construct a heterogeneous graph.
Based on previous medical representation learning, we obtain the feature representation of each node based on the medical corpus obtained from the admission record.
Due to the complexity of the medical knowledge field, we appropriately enlarge the selection range of the relationship to further test the filtering performance of our \RSRL Framework for camouflaged neighbors. 
For specific relations, since the previous work~\cite{cao2020multi} has described the concept of different fields in detail, we design four kinds of relations for the multi-relational graph based on it, as shown below:
\begin{itemize}
\item V-A-V: it connects visits in the same age category. 
\item V-D-V: it connects visits having the same diagnoses.
\item V-P-V: it connects visits with at least one same procedure code.
\item V-M-V: it connects visits with at least one same microbiology code.
\end{itemize}

\subsubsection{Evaluation of Datasets}\label{sec:x}
We measure different datasets from the four metrics of node distribution, edge distribution, and feature similarity and label similarity. 
It can be observed that in fraud detection tasks, Yelp and Amazon have only 14.5\% and 9.5\% of fraud nodes, while mimic has a more balanced ratio. 
In addition, the different relations of the three datasets all have different number levels and uneven edge distribution. 
In addition, in the Yelp and Amazon datasets, edges with different relations have balanced feature similarity, but existing edges have higher feature similarity and lower label similarity. 
For instance, the average feature similarity of R-T-R is 0.79, but the label similarity is only 0.05. 
This shows that the fraudsters successfully pretends to be in benign entities and needs a more effective way to identify it. 
In general, these characteristics challenge the model's ability to learn from data sets in different situations. 
Specifically, Yelp and Amazon focus on the ability to challenge uneven data sets. 
Mimic focuses more on the ability of filtering high-density neighbor nodes.
The relation matrix of MIMIC under each relation is denser, one order of magnitude higher than that of Yelp and Amazon.

\subsection{Baselines and Variations}\label{sec:baselines-and-variations}
\subsubsection{Baselines}\label{sec:baselines} 
To verify the effectiveness of \RioGNN in mitigating mutual interference between similar tasks and the model, we compare it with traditional and latest GNN baselines under semi-supervised learning settings.
For all baseline models, we use the open-source implementations.

The first three models are compared as traditional GNN baselines.
\begin{itemize}
\item \textbf{GCN}~\cite{kipf2017semi}: is a representative of the spectral graph convolution method, which sets up a simple and well-behaved hierarchical propagation rule for neural network models.
This rule runs directly on the graph, and uses the first-order approximation of the Chebyshev polynomial to complete an efficient graph convolution architecture.
\item \textbf{GAT}~\cite{velivckovic2018graph}: is a neural network architecture combined with the attention mechanism that runs on graph-structured data.
It uses masked self-attentional layers to give importance to the edges between nodes, help the model learn structural information, and assign different weights to different nodes in the neighborhood without expensive calculations and pre-definitions.
\item \textbf{Graph-SAGE}~\cite{hamilton2017inductive}: is a representative non-spectrogram method.
For each node, this method provides a general inductive framework that samples and aggregates its local neighbors' features to generate the embedding instead of training a separate embedding.
It improves the scalability and flexibility of GNNs.
\end{itemize}

The second ten baselines are the latest GNN models that handle multi-relational data or the datasets used in this article.
\begin{itemize}
\item \textbf{RGCN}~\cite{schlichtkrull2018modeling}: is a relational GCN model that uses Gaussian distribution as the hidden layer node feature representation, and relies on the attention mechanism to automatically assign the weight of each neighbor to aggregate neighbor information. 
\item \textbf{GeniePath}~\cite{liu2019b_geniepath}: is a scalable graph neural network model used to learn the adaptive receptive domain of neural networks defined on permutation invariant graph data.
Through the breadth and depth exploration of an adaptive path layer, the model can sense the importance of neighboring nodes and extract and filter signals gathered from the neighborhood.
\item \textbf{Player2Vec}~\cite{zhang2019key}: is an AHIN representation learning model that maps the attribute heterogeneous information network (AHIN) to a multi-view network, encodes the correlation between users described by different design meta-paths, and uses the attention mechanism to fuse embeddings from each view to form the final node representation.
\item \textbf{SemiGNN}~\cite{wang2019semi}: is a semi-supervised attention graph neural network, in which a hierarchical attention mechanism is designed. Neighborhood information is integrated through node-level attention, and multi-view data is integrated through view-level attention, resulting in better accuracy and interpretability.
\item \textbf{GAS}~\cite{li2019spam}: uses both a heterogeneous graph and a homogeneous graph to capture the local and global context of a comment, and is a meta-path based heterogeneous GCN model. 
In our scenario, meta-paths are enumerated from relations.
\item \textbf{FdGars}~\cite{wang2019fdgars}: constructs a single homogeneous graph based on multiple relations and employs GNNs to aggregate neighborhood information.
Compared with our work, this model lacks neighborhood selection.
\item \textbf{GraphConsis}~\cite{liu2020alleviating}: is a model that combines context embedding with nodes, filters inconsistent neighbors and generates corresponding sampling probabilities.
The embeddings of sampled nodes from each relation are fused using a relation attention mechanism.
\item \textbf{HAN}~\cite{wang2019heterogeneous}: is a hierarchical attention network aggregating neighbor information via different meta-paths.
Although the input data is heterogeneous, meta-paths are symmetrical in our scenario (i.e., the end nodes are of the same type), thus the model is regarded as a homogeneous model.
\item \textbf{GCT}~\cite{choi2020learning}: is a basic model for learning the implicit EHR structure using Transformer. Using statistical data to guide the structure learning process solves the problem that existing methods require complete docking structure information. Specifically, they use the attention mask and prior knowledge to guide self-attention to learn the hidden EHR structure, and they can learn the underlying structure of the EHR together even when the structural information is missing.
\item \textbf{HSGNN}~\cite{liu2020health}: is a heterogeneous medical graph-based semi-supervised graph neural network, which combines both meta-path instance based similarity matrices and self-attention mechanism. 
\end{itemize}

The third two baselines are the latest Reinforcement Learning guided GNN models.
We respectively use the raw heterogeneous graph and the proposed multi-relation graph as the input of these models.
\begin{itemize}
\item \textbf{GraphNAS}~\cite{gao2019graphnas}: enables automatic search of the suitable graph neural architecture via reinforcement learning. 
This model uses a recurrent network to generate variable-length strings that describe the architectures of graph neural networks, and then trains the recurrent network with reinforcement learning to maximize the expected accuracy of the generated architectures. 
\item \textbf{Policy-GNN}~\cite{Lai2020policy}: is a meta-policy framework that adaptively learns an aggregation policy to sample diverse iterations of aggregations for different nodes.
To accelerate the learning process, we also use a buffer mechanism to enable batch training and parameter sharing mechanism to decrease the training cost.
\end{itemize}

The last baseline is from the preliminary version of this article.
\begin{itemize}
\item \textbf{CARE-GNN}~\cite{dou2020enhancing}: A layer of label-perceived similarity measure is used to find information-rich neighboring nodes.
Then the Bernoulli Multi-armed Bandit (BMAB) mechanism is used to explore the optimal number of neighbors for each relationship.
\end{itemize}

Among those baselines, GCN, GAT, GraphSAGE, and GeniePath runs on the homogeneous graph (i.e., Relation ALL in Table~\ref{tab:dataset}), where all relations are merged together. 
GraphNAS$^{H}$ and Policy-GNN$^{H}$ runs on the raw heterogeneous graph. 
Other models run on the multi-relational graph.
On the multi-relational graph, they process information from different relations in their methods.

\subsubsection{Variations}\label{sec:variations}
We have implemented many variants of the \RioGNN model.
The configurations of different variants are shown in Table~\ref{tab:fraud_variants_setup}.
In detail, the setting of the model variants mainly revolves around the key mechanisms of the three modules: Label-aware Similarity Measure, Similarity-aware Neighbor Selector, and Relation-aware Neighbor Aggregator.

\begin{table}[t]
    \setlength{\abovecaptionskip}{0.cm}
    \setlength{\belowcaptionskip}{-0.cm}
    \caption{Comparison of main functions of different variants.}\label{tab:fraud_variants_setup}
    \centering
    \scalebox{0.95}{
        \begin{tabular}{c|cccccccc}
            \hline
            \multicolumn{1}{c|}{Models}&\multicolumn{1}{c}{\textbf{Multi-layer}}&\multicolumn{1}{c}{\textbf{RL Module}}&\multicolumn{1}{c}{\textbf{Action Space}}&\multicolumn{1}{c}{\textbf{Recursion}}&\multicolumn{1}{c}{\textbf{Inter-AGG}}\\
            \hline
            \multicolumn{1}{c|}{\RioGNN$_{2l}$}&\checkmark&AC&Discrete&\checkmark&Threshold\\
            \hline
            \multicolumn{1}{c|}{BIO-GNN}&$\times$&BMAB&Discrete&\checkmark&Threshold\\
            \multicolumn{1}{c|}{ROO-GNN}&$\times$&AC&Discrete&$\times$&Threshold\\
            \hline
            \multicolumn{1}{c|}{RIO-Att}&$\times$&AC&Discrete&\checkmark&Attention\\
            \multicolumn{1}{c|}{RIO-Weight}&$\times$&AC&Discrete&\checkmark&Weight\\
            \multicolumn{1}{c|}{RIO-Mean}&$\times$&AC&Discrete&\checkmark&Mean\\
            \hline
            \multicolumn{1}{c|}{\RioGNN}&$\times$&AC&Discrete&\checkmark&Threshold\\
            \hline
        \end{tabular}
    }
\end{table}

The first one given is a variation of the unexpanded multi-layer version of the Label-aware Similarity Measure section.
\begin{itemize}
\item \textbf{\RioGNN$_{2l}$:} It uses the Actor-Critic (AC) algorithm with a discrete strategy to recursively select the filter thresholds of different relationships, and uses the filter thresholds as relation weights to aggregate neighbors between different relations.
But the Label-aware Similarity Measure uses a 2-layer structure for neighbor selection.
\end{itemize}

The next two methods are variants according to the Similarity-aware Neighbor Selector module.
\begin{itemize}
\item \textbf{BIO-GNN:} This variant is a method with the single-layer similarity-aware neighbor selection and uses filtering thresholds for aggregation between relations.
But the Bernoulli Multi-armed Bandit (BMAB) algorithm of discrete strategy is used to recursively select the filtering threshold of relations.

\item \textbf{ROO-GNN:} This variant is a method with the single-layer similarity-aware neighbor selection and uses filtering thresholds for aggregation between relations.
But the Actor-Critic algorithm of discrete strategy is used to directly select the filter threshold of relationships.
This method can be regarded as a non-recursive (one-depth) version of \RioGNN.
\end{itemize}

The difference between the following three variants is reflected in the aggregation method between different relations of Relation-aware Neighbor Aggregator.
\begin{itemize}
\item \textbf{RIO-Att:} This variant uses single-layer similarity perception for neighbor selection, and uses the Actor-Critic algorithm with a discrete strategy to recursively select the filter thresholds of different relations.
But it chooses the method of Attention~\cite{velivckovic2018graph} when aggregating neighbors between different relations.
\item \textbf{RIO-Weight:} This variant uses single-layer similarity perception for neighbor selection, and uses the Actor-Critic algorithm with a discrete strategy to recursively select the filter thresholds of different relations.
But it chooses the method of Weight~\cite{liu2018heterogeneous} when aggregating neighbors between different relations.
\item \textbf{RIO-Mean:} This variant uses single-layer similarity perception for neighbor selection, and uses the Actor-Critic algorithm with a discrete strategy to recursively select the filter thresholds of different relations.
But it chooses the method of Mean~\cite{hamilton2017inductive} when aggregating neighbors between different relations.
\end{itemize}

\subsubsection{RL Variations}\label{sec:rl-variations}
In order to better discuss the adaptability of the framework of this article to a variety of reinforcement learning algorithms, we conducted experiments on two action spaces (discrete and continuous) of different reinforcement learning algorithms.

The first three reinforcement learning models are based on discrete action spaces. That is, in the process of constructing the reinforcement learning forest, a discrete filtering threshold is used as the action type of reinforcement learning.
\begin{itemize}
\item \textbf{AC~\cite{konda2000actor}:} Actor-Critic (AC) method combines the advantages of the value-based method and policy-based method.
The value-based method is used to train the Q function to improve the sample utilization efficiency.
The policy-based method is used to train the strategy, which is suitable for discrete and continuous action spaces.
This kind of method can be regarded as an extension of the value-based method in the continuous action space, or as an improvement of the policy-based method to reduce the sampling variance.

\item \textbf{DQN~\cite{mnih2015dqn}:} Deep Q-Learning (DQN) is a temporal-difference, value-based and off-policy reinforcement learning method. DQN approximately solves the dimensional disaster problem of Q-Learning method in the face of high-dimensional state and action through the function approximation. In addition, the traditional Q-Learning method uses samples with time series for single-step update, and the $Q$ value is updated by the sample continuity. DQN uses random data for gradient descent due to trial and error to collect a large number of samples, which can break the correlation between data.

\item \textbf{PPO~\cite{schulman2017proximal}:} Proximal Policy Optimization (PPO) restricts the update step size on the basis of Policy Gradient (PG) to prevent policy collapse and make the algorithm rise more steadily.
\end{itemize}

The next four reinforcement learning models are based on continuous action spaces. That is, in the process of constructing the reinforcement learning forest, a continuous filtering threshold is used as the action type of reinforcement learning.

\begin{itemize}
\item \textbf{AC~\cite{konda2000actor}:} We use the AC method to conduct variation experiments in both discrete action space and continuous action space.

\item \textbf{DDPG~\cite{lillicrap2019ddpg}:} Deep Deterministic Policy Gradient (DDPG) is an off-policy algorithm for continuous control developed by DeepMind, which is more sample efficient than PPO.
DDPG trains a deterministic policy, that is, only one optimal action is considered in each state. 

\item \textbf{SAC~\cite{he2019hetespaceywalk}:} Soft Actor-Critic (SAC) is an off-policy algorithm developed for Maximum Entropy Reinforcement learning.
Compared with DDPG, Soft Actor-Critic uses stochastic policy, which has certain advantages over deterministic policies.
Soft Actor-Critic has achieved outstanding results in the public benchmark and can be directly applied to real robots.

\item \textbf{TD3~\cite{scott2018td3}:} Twin Delayed Deep Deterministic policy gradient (TD3) is a temporal-difference, policy based and policy gradient reinforcement learning method.
TD3 is an optimized version of DDPG. It uses two sets of networks to estimate the $Q$ value, and the relatively smaller one is used as the update target.

\end{itemize}

\subsection{Model Training}\label{sec:model-training}
We use unified embedding size (64), batch size (1024 for Yelp, 256 for Amazon and MIMIC-III), learning rate (0.01), the similarity loss weight ($\lambda_{1} = 2$), L2 regularization weight ($\lambda_{2} = 0.001$) for all the models. 
Except for the comparative experiments specifically explained, other experiments that are not explained all use a 40\% training ratio, under-sampling ratio $1:1$, deep switching number 3, weight parameter of depth first and breadth first $\alpha$ as 10, and a single-layer similarity perception structure with backtracking.
For the reinforcement learning model, we use gamma (0.95), learning rate (0.001) as unified parameters and buffer capacity (5), batch size (1) as parameters for DDPG, DQN, SAC and TD3.
We conduct the sensitivity study for deep switching number, backtracking setting, under-sampling ratio in Section~\ref{sec:hyper-parameter}. In addition, we also discuss weight parameter of depth first and breadth first in Section~\ref{sec:versatility}, and training ratio in Section~\ref{sec:fraud-overall} and Section~\ref{sec:diabetes-overall}.

\subsection{Evaluation Metrics}\label{exp:evaluation-metrics}

We utilize ROC-AUC (AUC) \cite{tom2006auc} and Recall to evaluate the overall performance of all classifiers.
AUC is computed based on the relative ranking of prediction probabilities of all instances, which could eliminate the influence of imbalanced classes.
The Recall is defined as:
\begin{equation}\label{eq:auc}
Recall=\frac{TP}{TP+FN},
\end{equation}
where $TP$ is True Positive, $FN$ is False Negative.
The AUC is defined as:
\begin{equation}\label{eq:recall}
AUC = \frac{1}{2}\sum_{i=1}^{m-1}(x_{i+1}-x_{i})(y_{i}+y_{i+1}),
\end{equation}
where $y$ is True Posotive Rate ($TPR=\frac{TP}{TP+FN}$), $x$ is False Positive Rate ($FPR=\frac{FP}{FP+TN}$). And $FP$ is False Positive, $TN$ is True Negative.

In the clustering task, we use Normalized Mutual Information (NMI) and Adjusted Rand Index (ARI) as the performance indicators.
The ARI is defined as:
\begin{equation}\label{eq:recall}
ARI = \frac{\sum_{ij}(^{n_{ij}}_{2})-[\sum_{i}(^{a_{i}}_{2})\sum_{j}(^{b_{j}}_{2})]/(^{n}_{2})}{\frac{1}{2}[\sum_{i}(^{a_{i}}_{2})-\sum_{j}(^{b_{j}}_{2})]-[\sum_{i}(^{a_{i}}_{2})\sum_{j}(^{b_{j}}_{2})]/(^{n}_{2})},
\end{equation}
where each $n_{ij}$ represents the number of nodes located in class$_{i}$ and cluster$_{j}$ at the same time, $a_{i}$ is the number of the nodes in class$_{i}$ and $b_{j}$ is the number of the nodes in cluster$_{i}$.
The NMI is defined as:
\begin{equation}\label{eq:recall}
NMI = \frac{I(\omega;C)}{[H(\omega)+H(C)]/2},
\end{equation}
where $I$ is mutual information, $H$ is entropy.

In inductive learning, in order to better measure the effectiveness, we add the F1 indicator to measure after the two indicators of AUC and Recall. F1 is defined as:
\begin{equation}\label{eq:auc}
F1=2\cdot\frac{Precision\cdot Recall}{Precision+Recal},
\end{equation}
where $Precision=\frac{TP}{TP+FP}$, $TP$ is True Positive, $FP$ is False Positive.

%% file: 5-Result.tex
\section{Result and discussion}
\label{sec:result}

\subsection{Overall Evaluation of Fraud Detection Task}\label{sec:fraud-overall}
\begin{sidewaystable}[thp]
    \setlength{\abovecaptionskip}{-16.cm}
    \centering
    \scalebox{0.92}{
        \begin{tabular}{c|cccc|cccc||cccc|cccc}
            \hline
            \multirow{4}*{Models}&\multicolumn{8}{c||}{\multirow{2}*{\textbf{Yelp}}}&\multicolumn{8}{c}{\multirow{2}*{\textbf{Amazon}}}\\
            &&&&\multicolumn{1}{c}{}&&&&&&&&\multicolumn{1}{c}{}&&&&\\
            \cline{2-17}
            &\multicolumn{4}{c|}{\textbf{AUC}}&\multicolumn{4}{c||}{\textbf{Recall}}&\multicolumn{4}{c|}{\textbf{AUC}}&\multicolumn{4}{c}{\textbf{Recall}}\\
            &5\%&10\%&20\%&40\%&5\%&10\%&20\%&40\%&5\%&10\%&20\%&40\%&5\%&10\%&20\%&40\%\\
            \hline
            \multirow{2}*{\textbf{GCN}}&\multirow{2}*{54.98}&\multirow{2}*{50.94}&\multirow{2}*{53.15}&\multirow{2}*{52.47}&\multirow{2}*{53.12}&\multirow{2}*{51.10}&\multirow{2}*{53.87}&\multirow{2}*{50.81}&\multirow{2}*{74.44}&\multirow{2}*{75.25}&\multirow{2}*{75.13}&\multirow{2}*{74.34}&\multirow{2}*{65.54}&\multirow{2}*{67.81}&\multirow{2}*{66.15}&\multirow{2}*{67.45}\\
            \multirow{2}*{\textbf{GAT}}&\multirow{2}*{56.23}&\multirow{2}*{55.45}&\multirow{2}*{57.69}&\multirow{2}*{56.24}&\multirow{2}*{54.68}&\multirow{2}*{52.34}&\multirow{2}*{53.20}&\multirow{2}*{54.52}&\multirow{2}*{73.89}&\multirow{2}*{74.55}&\multirow{2}*{72.10}&\multirow{2}*{72.16}&\multirow{2}*{63.22}&\multirow{2}*{65.84}&\multirow{2}*{67.13}&\multirow{2}*{65.51}\\
            \multirow{2}*{\textbf{GraphSAGE}}&\multirow{2}*{53.82}&\multirow{2}*{54.20}&\multirow{2}*{56.12}&\multirow{2}*{54.00}&\multirow{2}*{54.25}&\multirow{2}*{52.23}&\multirow{2}*{52.69}&\multirow{2}*{52.86}&\multirow{2}*{70.71}&\multirow{2}*{73.97}&\multirow{2}*{73.97}&\multirow{2}*{75.27}&\multirow{2}*{69.09}&\multirow{2}*{69.36}&\multirow{2}*{70.30}&\multirow{2}*{70.16}\\
            &&&&&&&&&&&&&&&&\\
            \hline
            \multirow{2}*{\textbf{RGCN}}&\multirow{2}*{50.21}&\multirow{2}*{55.12}&\multirow{2}*{55.05}&\multirow{2}*{53.38}&\multirow{2}*{50.38}&\multirow{2}*{51.75}&\multirow{2}*{50.92}&\multirow{2}*{50.43}&\multirow{2}*{75.12}&\multirow{2}*{74.13}&\multirow{2}*{75.58}&\multirow{2}*{74.68}&\multirow{2}*{64.23}&\multirow{2}*{67.22}&\multirow{2}*{65.08}&\multirow{2}*{67.68}\\
            \multirow{2}*{\textbf{GeniePath}}&\multirow{2}*{56.33}&\multirow{2}*{56.29}&\multirow{2}*{57.32}&\multirow{2}*{55.91}&\multirow{2}*{52.33}&\multirow{2}*{54.35}&\multirow{2}*{54.84}&\multirow{2}*{50.94}&\multirow{2}*{71.56}&\multirow{2}*{72.23}&\multirow{2}*{71.89}&\multirow{2}*{72.65}&\multirow{2}*{65.56}&\multirow{2}*{66.63}&\multirow{2}*{65.08}&\multirow{2}*{65.41}\\
            \multirow{2}*{\textbf{Player2Vec}}&\multirow{2}*{51.03}&\multirow{2}*{50.15}&\multirow{2}*{51.56}&\multirow{2}*{53.65}&\multirow{2}*{50.00}&\multirow{2}*{50.00}&\multirow{2}*{50.00}&\multirow{2}*{50.00}&\multirow{2}*{76.86}&\multirow{2}*{75.73}&\multirow{2}*{74.55}&\multirow{2}*{56.94}&\multirow{2}*{50.00}&\multirow{2}*{50.00}&\multirow{2}*{50.00}&\multirow{2}*{50.00}\\
            \multirow{2}*{\textbf{SemiGNN}}&\multirow{2}*{53.73}&\multirow{2}*{51.68}&\multirow{2}*{51.55}&\multirow{2}*{51.58}&\multirow{2}*{52.28}&\multirow{2}*{52.57}&\multirow{2}*{52.16}&\multirow{2}*{50.59}&\multirow{2}*{70.25}&\multirow{2}*{76.21}&\multirow{2}*{73.98}&\multirow{2}*{70.35}&\multirow{2}*{63.29}&\multirow{2}*{63.32}&\multirow{2}*{61.28}&\multirow{2}*{62.89}\\
            \multirow{2}*{\textbf{GraphConsis}}&\multirow{2}*{61.58}&\multirow{2}*{62.07}&\multirow{2}*{62.31}&\multirow{2}*{62.07}&\multirow{2}*{62.60}&\multirow{2}*{62.08}&\multirow{2}*{62.35}&\multirow{2}*{62.08}&\multirow{2}*{85.46}&\multirow{2}*{85.29}&\multirow{2}*{85.50}&\multirow{2}*{85.50}&\multirow{2}*{85.49}&\multirow{2}*{85.38}&\multirow{2}*{85.59}&\multirow{2}*{85.53}\\
            \multirow{2}*{\textbf{GAS}}&\multirow{2}*{54.43}&\multirow{2}*{52.58}&\multirow{2}*{52.51}&\multirow{2}*{52.60}&\multirow{2}*{53.40}&\multirow{2}*{53.26}&\multirow{2}*{53.37}&\multirow{2}*{51.61}&\multirow{2}*{71.40}&\multirow{2}*{77.49}&\multirow{2}*{74.51}&\multirow{2}*{71.03}&\multirow{2}*{64.31}&\multirow{2}*{64.57}&\multirow{2}*{62.08}&\multirow{2}*{63.74}\\
            \multirow{2}*{\textbf{FdGars}}&\multirow{2}*{61.77}&\multirow{2}*{62.15}&\multirow{2}*{62.81}&\multirow{2}*{62.66}&\multirow{2}*{62.83}&\multirow{2}*{62.16}&\multirow{2}*{62.73}&\multirow{2}*{62.40}&\multirow{2}*{85.58}&\multirow{2}*{85.41}&\multirow{2}*{85.88}&\multirow{2}*{85.81}&\multirow{2}*{85.83}&\multirow{2}*{85.73}&\multirow{2}*{85.84}&\multirow{2}*{85.93}\\
            &&&&&&&&&&&&&&&&\\
            \hline
            \multirow{2}*{\textbf{GraphNAS$^{H}$}}&\multirow{2}*{52.93}&\multirow{2}*{54.69}&\multirow{2}*{56.73}&\multirow{2}*{54.46}&\multirow{2}*{52.40}&\multirow{2}*{54.15}&\multirow{2}*{55.69}&\multirow{2}*{56.16}&\multirow{2}*{71.01}&\multirow{2}*{72.48}&\multirow{2}*{73.52}&\multirow{2}*{76.05}&\multirow{2}*{69.17}&\multirow{2}*{69.48}&\multirow{2}*{70.35}&\multirow{2}*{70.16}\\
            \multirow{2}*{\textbf{GraphNAS}}&\multirow{2}*{53.26}&\multirow{2}*{55.31}&\multirow{2}*{57.15}&\multirow{2}*{55.59}&\multirow{2}*{53.69}&\multirow{2}*{55.47}&\multirow{2}*{56.04}&\multirow{2}*{57.00}&\multirow{2}*{72.41}&\multirow{2}*{73.04}&\multirow{2}*{73.58}&\multirow{2}*{76.25}&\multirow{2}*{70.36}&\multirow{2}*{70.53}&\multirow{2}*{71.73}&\multirow{2}*{71.88}\\
            \multirow{2}*{\textbf{Policy-GNN$^{H}$}}&\multirow{2}*{54.04}&\multirow{2}*{55.73}&\multirow{2}*{59.30}&\multirow{2}*{60.60}&\multirow{2}*{53.08}&\multirow{2}*{55.35}&\multirow{2}*{58.75}&\multirow{2}*{59.99}&\multirow{2}*{72.20}&\multirow{2}*{73.30}&\multirow{2}*{74.11}&\multirow{2}*{77.20}&\multirow{2}*{70.10}&\multirow{2}*{71.20}&\multirow{2}*{73.08}&\multirow{2}*{74.44}\\
            \multirow{2}*{\textbf{Policy-GNN}}&\multirow{2}*{55.75}&\multirow{2}*{56.29}&\multirow{2}*{60.01}&\multirow{2}*{61.52}&\multirow{2}*{54.15}&\multirow{2}*{56.16}&\multirow{2}*{58.95}&\multirow{2}*{60.33}&\multirow{2}*{73.69}&\multirow{2}*{74.06}&\multirow{2}*{75.29}&\multirow{2}*{78.85}&\multirow{2}*{71.34}&\multirow{2}*{72.46}&\multirow{2}*{74.55}&\multirow{2}*{76.70}\\
            &&&&&&&&&&&&&&&&\\
            \hline
            \multirow{2}*{\textbf{CARE-GNN}}&\multirow{2}*{71.26}&\multirow{2}*{73.31}&\multirow{2}*{74.45}&\multirow{2}*{75.70}&\multirow{2}*{67.53}&\multirow{2}*{67.77}&\multirow{2}*{68.60}&\multirow{2}*{71.92}&\multirow{2}*{89.54}&\multirow{2}*{89.44}&\multirow{2}*{89.45}&\multirow{2}*{89.73}&\multirow{2}*{88.34}&\multirow{2}*{88.29}&\multirow{2}*{88.27}&\multirow{2}*{88.48}\\
            &&&&&&&&&&&&&&&&\\
            \hline
            \multirow{2}*{\textbf{\RioGNN}}&\multirow{2}*{\textbf{81.97}}&\multirow{2}*{\textbf{83.72}}&\multirow{2}*{\textbf{82.31}}&\multirow{2}*{\textbf{83.54}}&\multirow{2}*{\textbf{75.33}}&\multirow{2}*{\textbf{75.78}}&\multirow{2}*{\textbf{75.51}}&\multirow{2}*{\textbf{76.19}}&\multirow{2}*{\textbf{95.44}}&\multirow{2}*{\textbf{95.41}}&\multirow{2}*{\textbf{95.63}}&\multirow{2}*{\textbf{96.19}}&\multirow{2}*{\textbf{90.17}}&\multirow{2}*{\textbf{89.48}}&\multirow{2}*{\textbf{89.51}}&\multirow{2}*{\textbf{89.82}}\\
            &&&&&&&&&&&&&&&&\\            
            \hline
        \end{tabular}
    }
    \caption{Fraud Detection results ($\%$) compared to the baselines.}\label{tab:fraud_baseline}
\end{sidewaystable}

\subsubsection{Accuracy Analysis}\label{sec:fraud-accuracy}
In this section, we conduct experiments to evaluate the accuracy of the fraud detection task on Yelp and Amazon datasets.
We report the best test results of \RioGNN, baselines and variants in five hundred epochs.
It can be observed from the results that \RioGNN performs better than other baselines and variants under most training ratios or indicators. 
This indicates the feasibility of \RioGNN in fraud detection scenarios. 

\textbf{Single-relations vs. Multi-relations. }
Table~\ref{tab:fraud_baseline} shows the results of baseline experiments built on different types of graphs for the fraud detection task.
To solve the diversity and heterogeneity of complex networks in actual fine-grained applications, we consider introducing a neural network with a multi-relational graph structure instead of a single relation structure.
However, from the results of some of the baselines in the table, although GCN, GAT, GraphSAGE and GeniePath models run on a single relation graph, they are better than RGCN, Player2Vec and SemiGNN in terms of accuracy of the Yelp dataset, which are run on the multi-relational graphs.
The observation above shows that the previous multi-relational GNNs are not suitable for  constructing multi-relational graphs in the fraud detection task.
Similar phenomena manifest in the Amazon dataset.
Besides, GraphConsis, FdGars, CARE-GNN and \RioGNN significantly outperform other models by 8.60\%-32.78\% over Yelp and Amazon datasets. 
This is because these four models sample the neighbors according to node features before aggregating them, indicating that the impurity neighbors will interfere with the aggregation process and the fraud detection task has a strong demand for neighbor sampling optimization.
CARE-GNN and \RioGNN have improved the AUC of 3.51\%-21.65\%, compared with the performance of GraphConsis and FdGars.
This is because \RioGNN can better use internal relations to solve downstream application problems through parameterized similarity measures and adaptive sampling thresholds, which shows that automated sampling has a significant improvement effect on fraud detection tasks.
The more remarkable result is that the proposed \RioGNN model improves the accuracy of 5.90\% and 10.41\% compared with CARE-GNN.
It verifies the advantage of combining the label-aware neighbor similarity measure and the Recursive and Scalable Reinforcement Learning framework, which can effectively break through the limitations of the CARE-GNN state observation range and manually specified strategies.
Meanwhile, \RioGNN has a promising effectiveness on fraud detection tasks.

\textbf{Heterogeneous vs. Multi-relation. }
In order to further analyze the accuracy of the multi-relational graph, we conduct heterogeneous graph experiments and multi-relational graph experiments on the latest GNN model guided by reinforcement learning. 
From the results of the heterogeneous graph model GraphNAS$^{H}$, Policy-GNN$^{H}$ and the multi-relationship graph model GraphNAS and Policy-GNN in Table~\ref{tab:fraud_baseline}, it can be found that the multi-relational graph compared with the heterogeneous graph brings an AUC improvement of $0.33\%$-$1.71\%$ in the Yelp and Amazon datasets. 
This confirms that in other similar models, the multi-relational graph we construct still has obvious advantages. 
In addition, we found that GraphNAS and Policy-GNN, which are also based on reinforcement learning guidance and use the multi-relational graph, have no significant advantages in AUC and Recall. 
This is because GraphNAS and Policy-GNN do not adaptively sample different relations, which causes them to be limited by the complexity of the relationship between Yelp and Amazon datasets. 
And for Policy-GNN, since the one-hop neighbor information of Yelp and Amazon is already rich enough, more multi-hop strategies cannot bring significant benefits.

\textbf{Training Percentage. }
To measure the impact of the training ratio on the classification accuracy, we use four different ratios of 5\% to 40\% for experiments.
It can be seen from Table~\ref{tab:fraud_baseline} that most of the baseline performance changes are not necessarily related to the increase in training percentage.
It indicates that the semi-supervised learning approach leveraging a small number of supervised signals is enough to train a good model.
Moreover, in the four different training ratios of the Yelp dataset, the AUC fluctuation range of \RioGNN is only within 1.57\% compared to the 4.44\% of CARE-GNN.
In Amazon, both models have good stability.
This is because the Amazon node features provide enough information to distinguish fraudsters, which is of higher quality than the Yelp dataset.
It also verifies from another result that \RioGNN has better stability and adaptability under complicated environments.

\begin{table}[t]
    \setlength{\abovecaptionskip}{0.cm}
    \setlength{\belowcaptionskip}{-0.cm}
    \caption{Fraud Detection classification results ($\%$) compared to \RioGNN variants.}\label{tab:fraud_variants}
    \centering
    \scalebox{1}{
        \begin{tabular}{p{0.5cm}p{2.5cm}|p{1.5cm}<{\centering}p{1.5cm}<{\centering}|p{1.5cm}<{\centering}p{1.5cm}<{\centering}}
            \hline
            &\multicolumn{1}{p{2.5cm}|}{\multirow{2}*{Models}}&\multicolumn{2}{c|}{\textbf{Yelp}}&\multicolumn{2}{c}{\textbf{Amazon}}\\
            \cline{3-6}
            &\multicolumn{1}{p{2.5cm}|}{}&\textbf{AUC}&\textbf{Recall}&\textbf{AUC}&\textbf{Recall}\\
            \hline
            &\multicolumn{1}{p{2.5cm}|}{\RioGNN$_{2l}$}&76.01&63.15&91.28&72.46\\
            \hline
            &\multicolumn{1}{p{2.5cm}|}{BIO-GNN}&78.67&71.21&95.47&88.35\\
            &\multicolumn{1}{p{2.5cm}|}{ROO-GNN}&\textbf{83.59}&    \textbf{75.56}&95.58&89.22\\
            \hline
            &\multicolumn{1}{p{2.5cm}|}{RIO-Att}&78.65&71.69&93.97&83.78\\
            &\multicolumn{1}{p{2.5cm}|}{RIO-Weight}&80.40&72.83&\textbf{96.25}&\textbf{89.61}\\
            &\multicolumn{1}{p{2.5cm}|}{RIO-Mean}&77.84&71.43&94.57&89.47\\
            \hline
            &\multicolumn{1}{p{2.5cm}|}{\RioGNN}&83.54&75.55&96.19&88.66\\
            \hline
        \end{tabular}
    }
\end{table}

\textbf{\RioGNN Variants in Classification. }
To measure the positive impact of the newly added mechanism on the classification accuracy of fraud detection tasks, we compare several variants of \RioGNN under the discrete strategy.
The experiment sets the training data ratio to 40\%, and the other settings are the same as Section~\ref{sec:hardware-software}.
We show the experimental results of spam review classification in the Yelp dataset and suspicious user classification in the Amazon dataset as shown in Table~\ref{tab:fraud_variants}.
From the results, the performance of all variants is better than the baseline model. Next, we will discuss the effects of different variants from three aspects.

Firstly, in the two datasets, except for the \RioGNN$_{2l}$ variant, all other variants only use the Label-aware Similarity Measure with a one-layer structure.
From the results in Table~\ref{tab:fraud_variants}, the \RioGNN$_{2l}$ variant is lower than all other variants, but it performs better than all baselines.
It can be found that in the fraud detection task, the increase in the number of layers does not bring about a significant increase in classification accuracy.
This is limited by the dataset size of Yelp and Amazon, and the importance of information in multi-hop neighbors is low.
We will continue to explore more multi-layer effects in Section~\ref{sec:fraud-efficiency}.

\begin{figure}[t]
\centering
\subfigure[AUC of Rio-GNN, BIO-GNN and ROO-GNN on Yelp.]{\label{fig:fraud-efficiency-a}
\begin{minipage}[t]{1\linewidth}
\centering
\includegraphics[width=15.3cm]{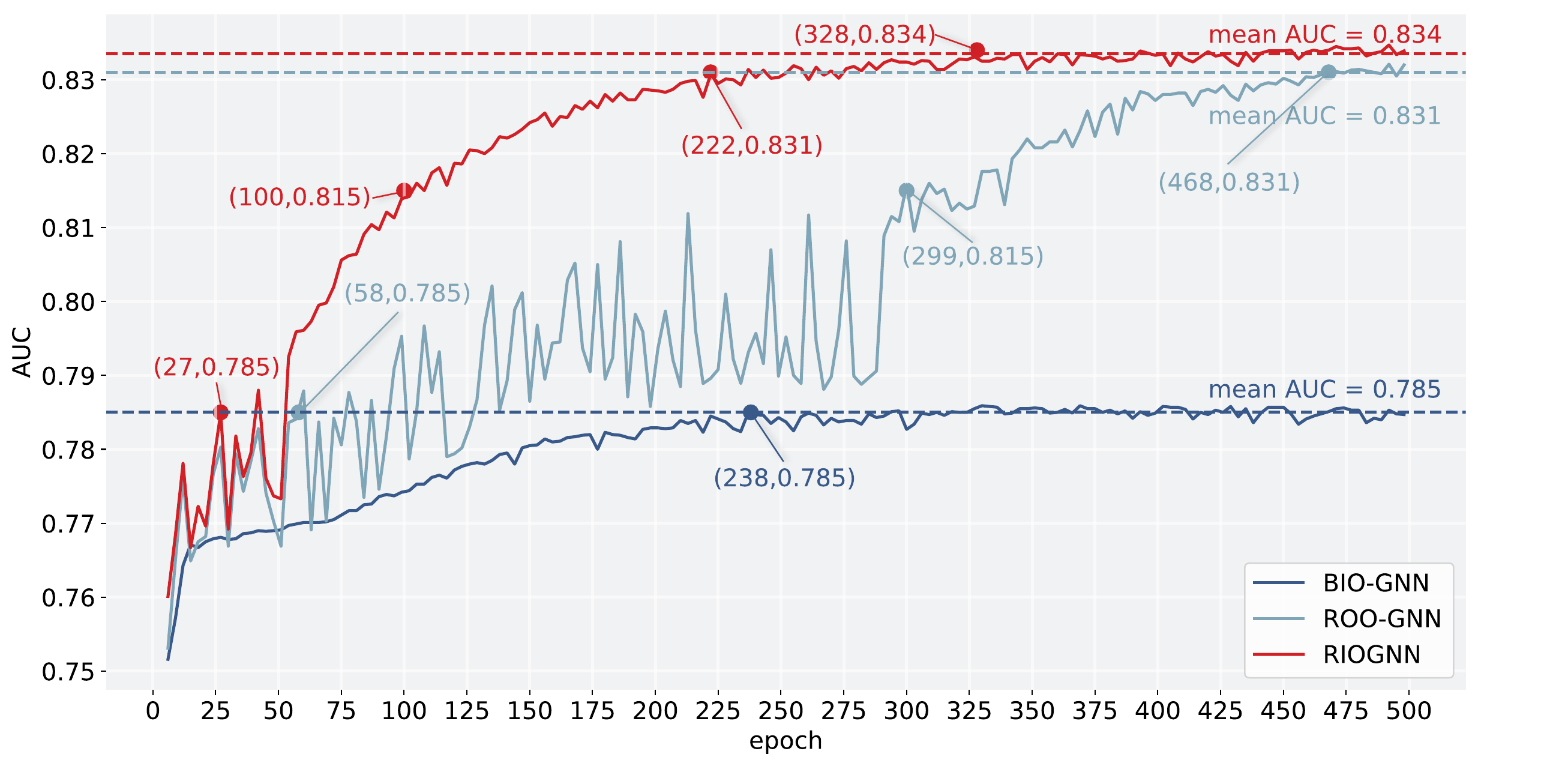}
\end{minipage}%
}%

\subfigure[AUC of \RioGNN and \RioGNN$_{No \ Recursion}$ on Amazon.]{\label{fig:fraud-efficiency-b}
\begin{minipage}[t]{1\linewidth}
\centering
\includegraphics[width=15.3cm]{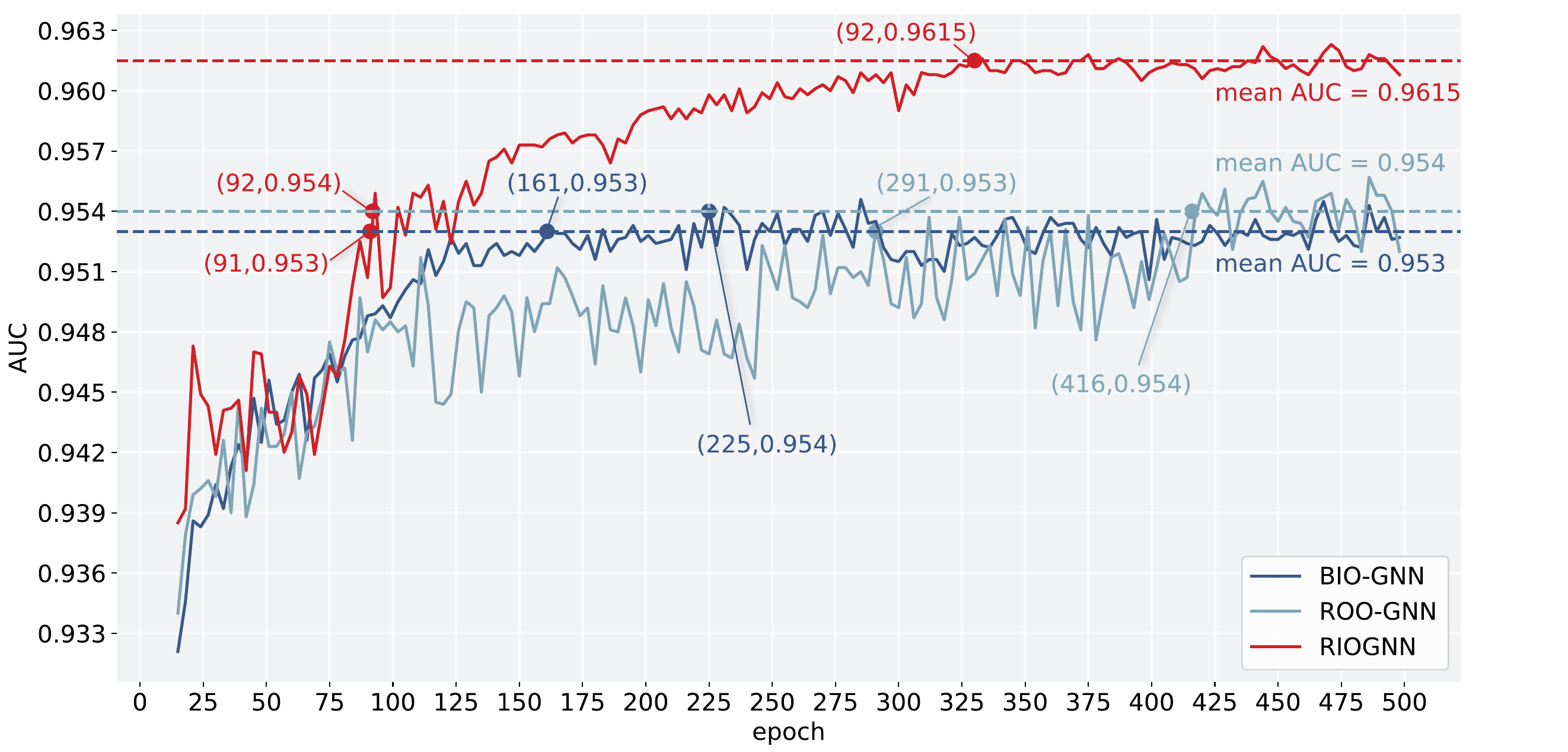}
\end{minipage}%
}%
\centering
\caption{The impact of recursive framework on computational efficiency.}\label{fig:fraud-efficiency}
\end{figure}

Secondly, for the similarity-aware neighbor selector part, we observe the comparison results of the variant BIO-GNN without adaptive strategy optimization reinforcement learning algorithm and the single-depth structure variant ROO-GNN without the recursive framework for the \RSRL framework.
The second part of Table~\ref{tab:fraud_variants} gives evidence of partial optimization. 
The results of the BIO-GNN variant on the \RioGNN model show that the automatic strategy optimization in Yelp and Amazon effectively improves the classification accuracy of 4.65\% and 0.89\%, and the BIO-GNN variant is also far better than most baseline models.
This shows that the Markov Decision Process has a positive effect on searching for the filtering threshold of the aggregation process.
Moreover, the reinforcement learning algorithm with dynamic iterative function and full action space learning process breaks the limitation of fixed strategy and observation range and obtains a better threshold selection effect, which also confirms the conjecture in Section~\ref{sec:rsrl}. 
Besides, combining Figure~\ref{fig:fraud-efficiency} with Table~\ref{tab:fraud_variants}, the accuracy of the ROO-GNN variant has little change compared with \RioGNN, that is, the multi-depth structure of \RioGNN can converge much quicker than the single-depth variant ROO-GNN whilst maintaining a higher accuracy rate.
This also implies the stability of the recursive framework in terms of classification accuracy.

Finally, from the results of the variation of the aggregation methods between the different relations in the third part of the Table~\ref{tab:fraud_variants}, \RioGNN has apparent advantages over the other three on the Yelp dataset, and both RIO-Weight and \RioGNN on the Amazon dataset have good results.
It confirms that \RioGNN does not need to train additional attention weights, and using the filtering threshold as an inter-relation aggregation weight can improve the performance of GNN and reduce the complexity of the model.
Therefore, it can get the best performance compared with other variants.
For the three variants, RIO-Weight has better results than the other two, but \RioGNN can maintain better accuracy in the dataset of different quality and structure and has a certain degree of adaptability.

\begin{table}[t]
    \setlength{\abovecaptionskip}{0.cm}
    \setlength{\belowcaptionskip}{-0.cm}
    \caption{Fraud detection clustering results ($\%$) compared to \RioGNN variants.}\label{tab:fraud_cluster}
    \centering
    \scalebox{1}{
        \begin{tabular}{c|c|c|cc|ccc|c}
            \hline
            \multicolumn{1}{c|}{\multirow{1}*{Dataset}}&\multicolumn{1}{c|}{\multirow{1}*{Metric}}&\multirow{1}*{\RioGNN$_{2l}$}&BIO-GNN&ROO-GNN&RIO-Att&RIO-Weight&RIO-Mean&\multirow{1}*{\RioGNN}\\
            \hline
            \multirow{2}*{\textbf{Yelp}}&\textbf{NMI}&3.18&9.36&\textbf{12.39}&9.80&12.05&8.39&12.22\\
            &\textbf{ARI}&6.12&11.84&\textbf{16.61}&11.88&15.88&8.80&16.45\\
            \hline
            \multirow{2}*{\textbf{Amazon}}&\textbf{NMI}&58.87&59.83&57.81&55.76&58.76&58.72&\textbf{61.26}\\
            &\textbf{ARI}&76.53&77.38&76.09&76.54&76.73&76.51&\textbf{78.40}\\
            \hline
        \end{tabular}
    }
\end{table}

\textbf{\RioGNN Variants in Clustering. }
In order to explore the effectiveness of \RioGNN in clustering tasks, we conduct clustering experiments on \RioGNN and its variant models. 
The experiment set a fixed training rate of $40\%$. 
We cluster the node representations learned by \RioGNN through K-Means. 
The results are shown in Table~\ref{tab:fraud_cluster}. 
We respectively count the best values of NMI and ARI indicators within 500 epochs. 
It can be seen from the results that compared with RIO-GNN$_{2l}$ and BIO-GNN, \RioGNN's NMI and ARI indicators in the Yelp dataset increase at least $9.04\%$ and $10.33\%$ respectively. 
Similarly, the Amazon dataset has risen by at least $2.39\%$ and $1.87\%$. 
This phenomenon is the same as the classification result, and is affected by the limitation of the size of the dataset, the choice of the action space, and the dynamic iterative function. 
In addition, the NMI and ARI of ROO-GNN in the Yelp dataset achieved the best results, while the \RioGNN in the Amazon dataset exceeded the NMI and ARI of ROO-GNN by $3.45\%$ and $2.31\%$. 
This is because the Yelp dataset is smaller than Amazon, so the optimization space of the recursive framework is also smaller. 
It shows that the recursive framework has better advantages in dense datasets. 
What is more noteworthy is that \RioGNN in the clustering experiment exceeds the effect of all GNN aggregation variants. 
NMI and ARI exceed $0.17\%$-$3.83\%$ and $0.57\%$-$7.65\%$ in the Yelp dataset, and exceed $2.5\%$-$5.5\%$ and $1.67\%$-$1.89\%$ in the Amazon dataset. 
This shows that directly using the filtering threshold as the weight of the aggregation has obvious advantages in clustering tasks.

\begin{figure}[t]
\centering
\subfigure[Scores of \RioGNN on Yelp.]{\label{fig:fraud-training-RioGNN-a}
\begin{minipage}[t]{0.5\linewidth}
\centering
\includegraphics[width=8.3cm]{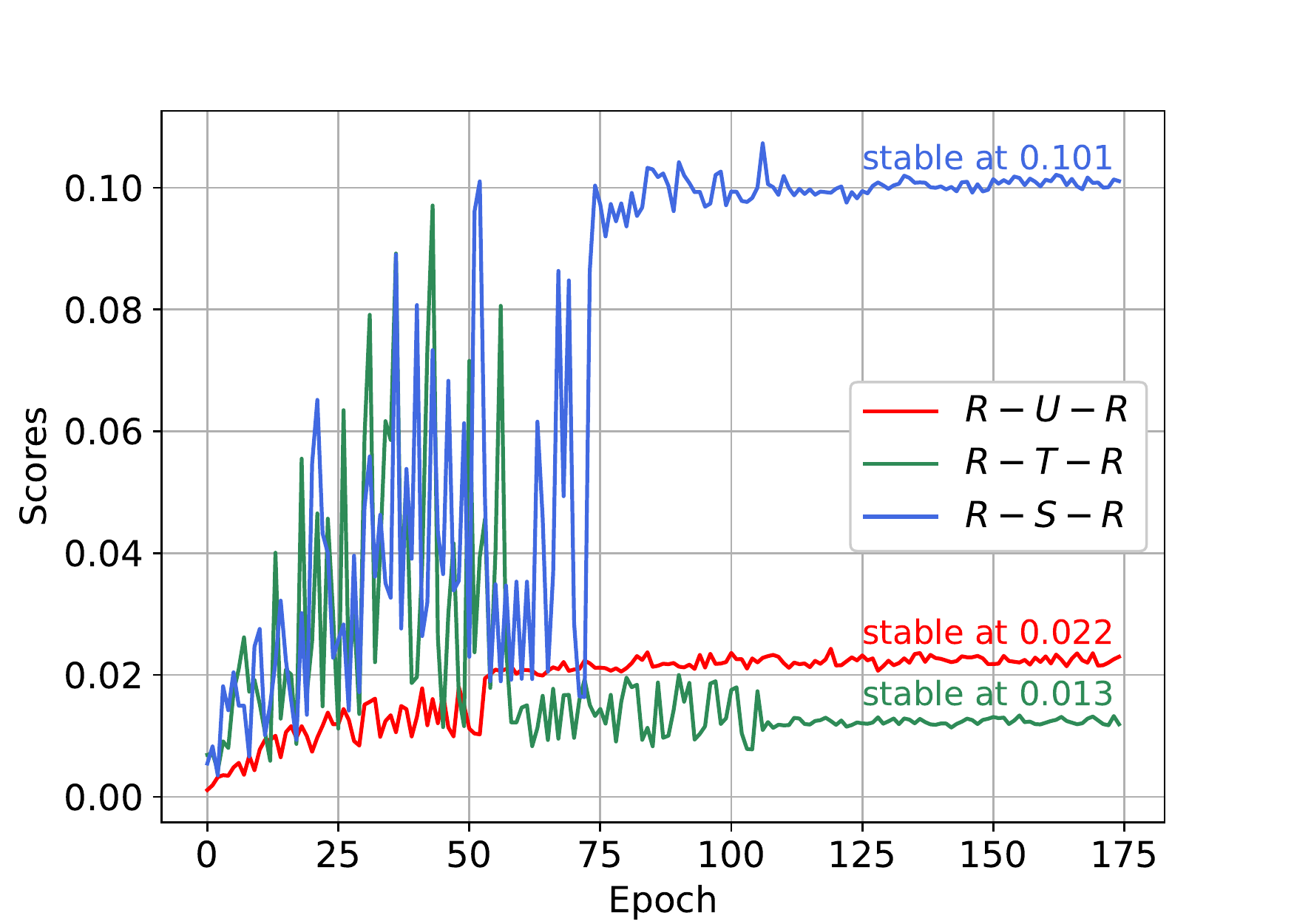}
\end{minipage}%
}%
\subfigure[Thresholds of \RioGNN on Yelp.]{\label{fig:fraud-training-RioGNN-b}
\begin{minipage}[t]{0.5\linewidth}
\centering
\includegraphics[width=8.3cm]{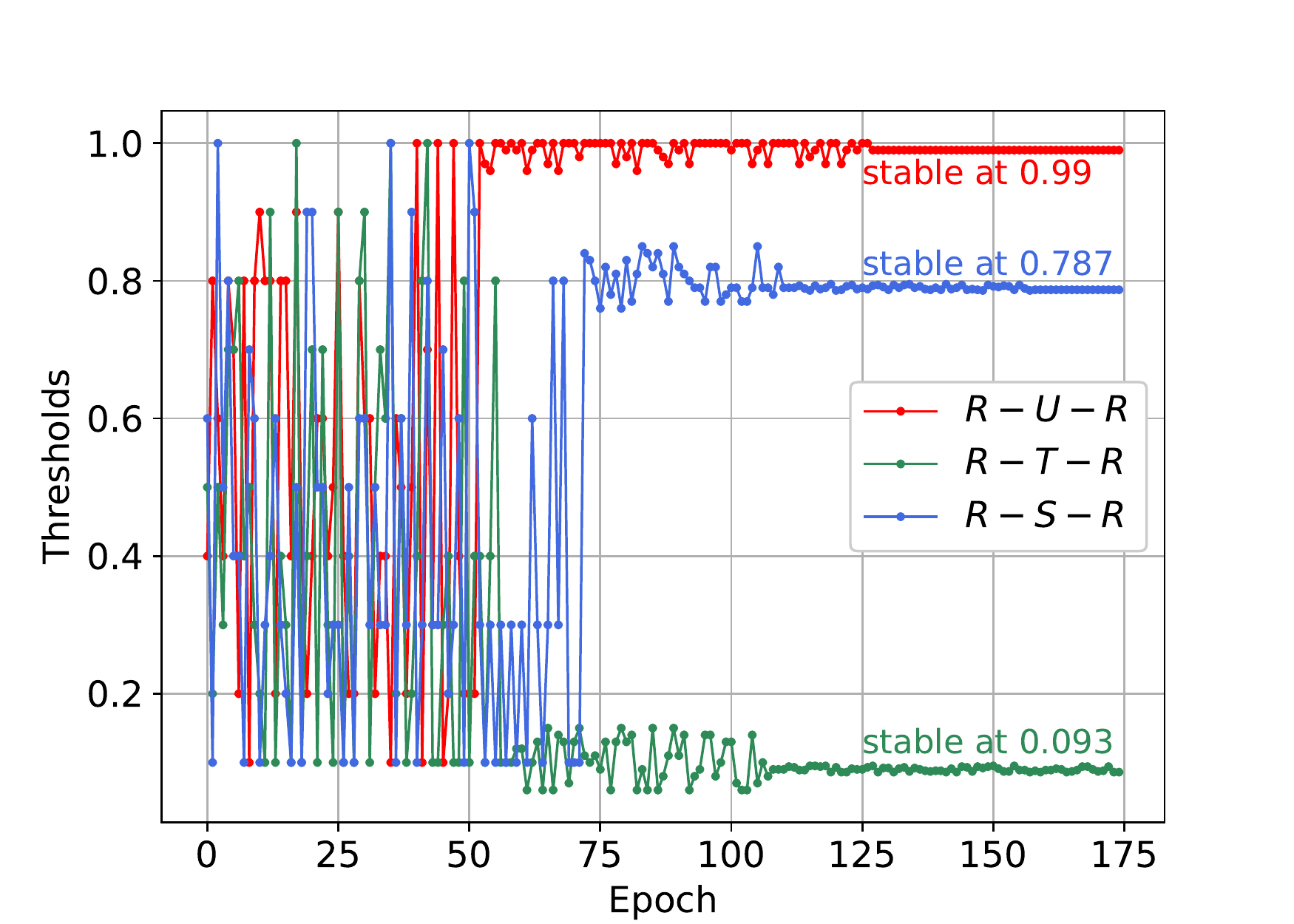}
\end{minipage}%
}%

\centering
\subfigure[Scores of \RioGNN on Amazon.]{\label{fig:fraud-training-RioGNN-c}
\begin{minipage}[t]{0.5\linewidth}
\centering
\includegraphics[width=8.3cm]{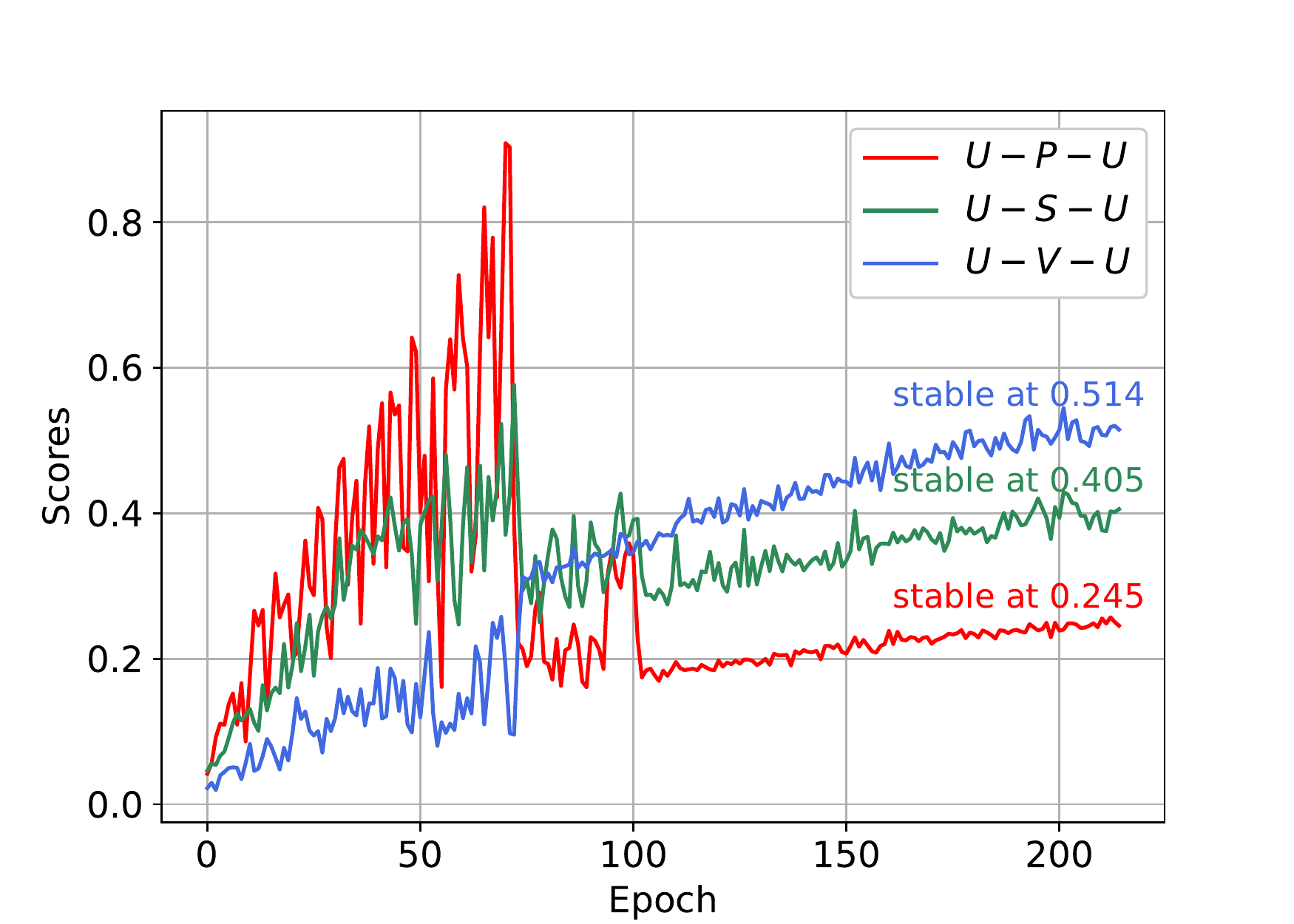}
\end{minipage}%
}%
\subfigure[Thresholds of \RioGNN on Amazon.]{\label{fig:fraud-training-RioGNN-d}
\begin{minipage}[t]{0.5\linewidth}
\centering
\includegraphics[width=8.3cm]{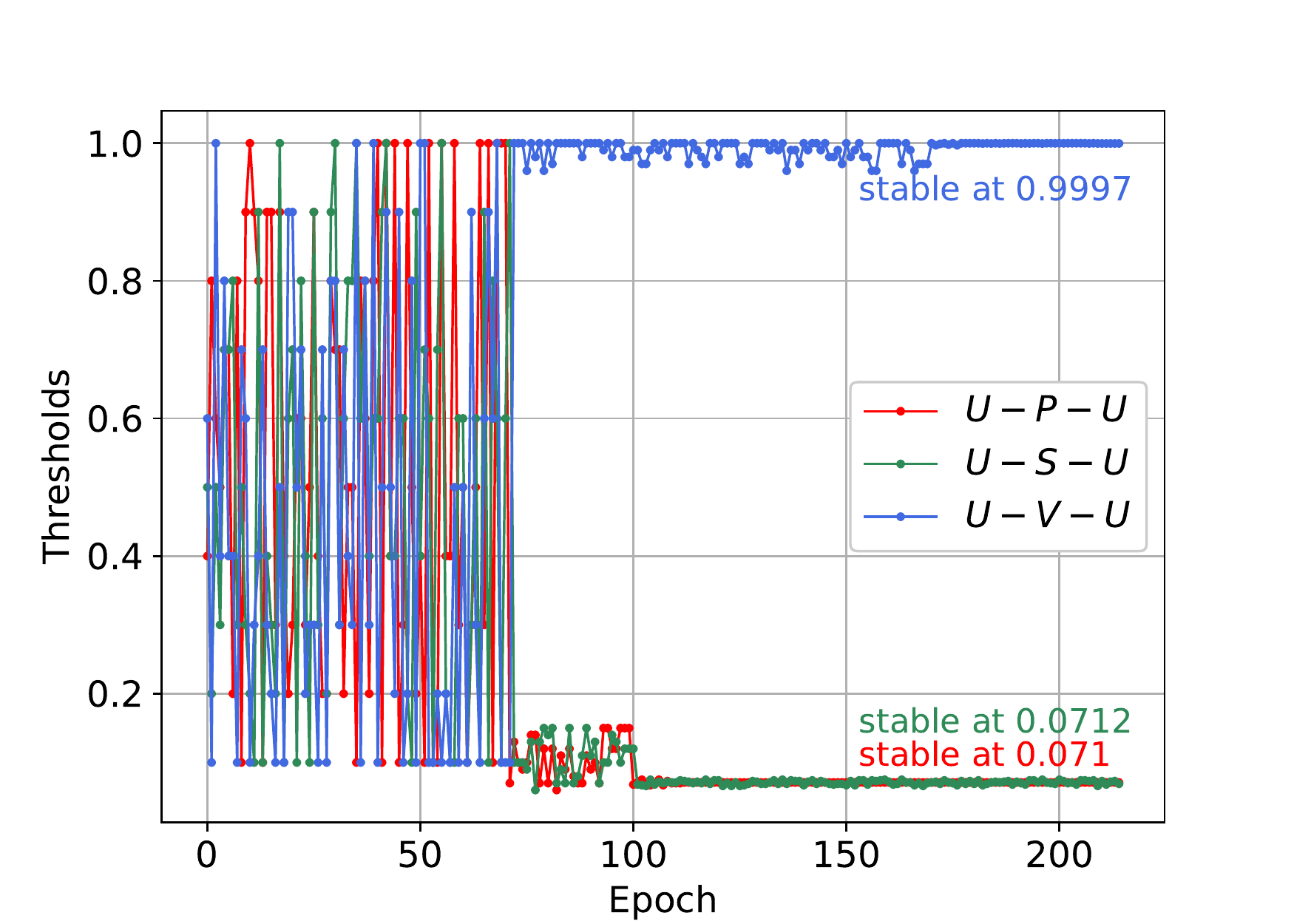}
\end{minipage}%
}%
\centering
\caption{The training scores and thresholds of \RioGNN on Yelp and Amazon.}\label{fig:fraud-training-RioGNN}
\end{figure}

\begin{figure}[h]
\centering
\subfigure[Scores of ROO-GNN on Yelp.]{\label{fig:fraud-training-varitants-a}
\begin{minipage}[h]{0.5\linewidth}
\centering
\includegraphics[width=8.3cm]{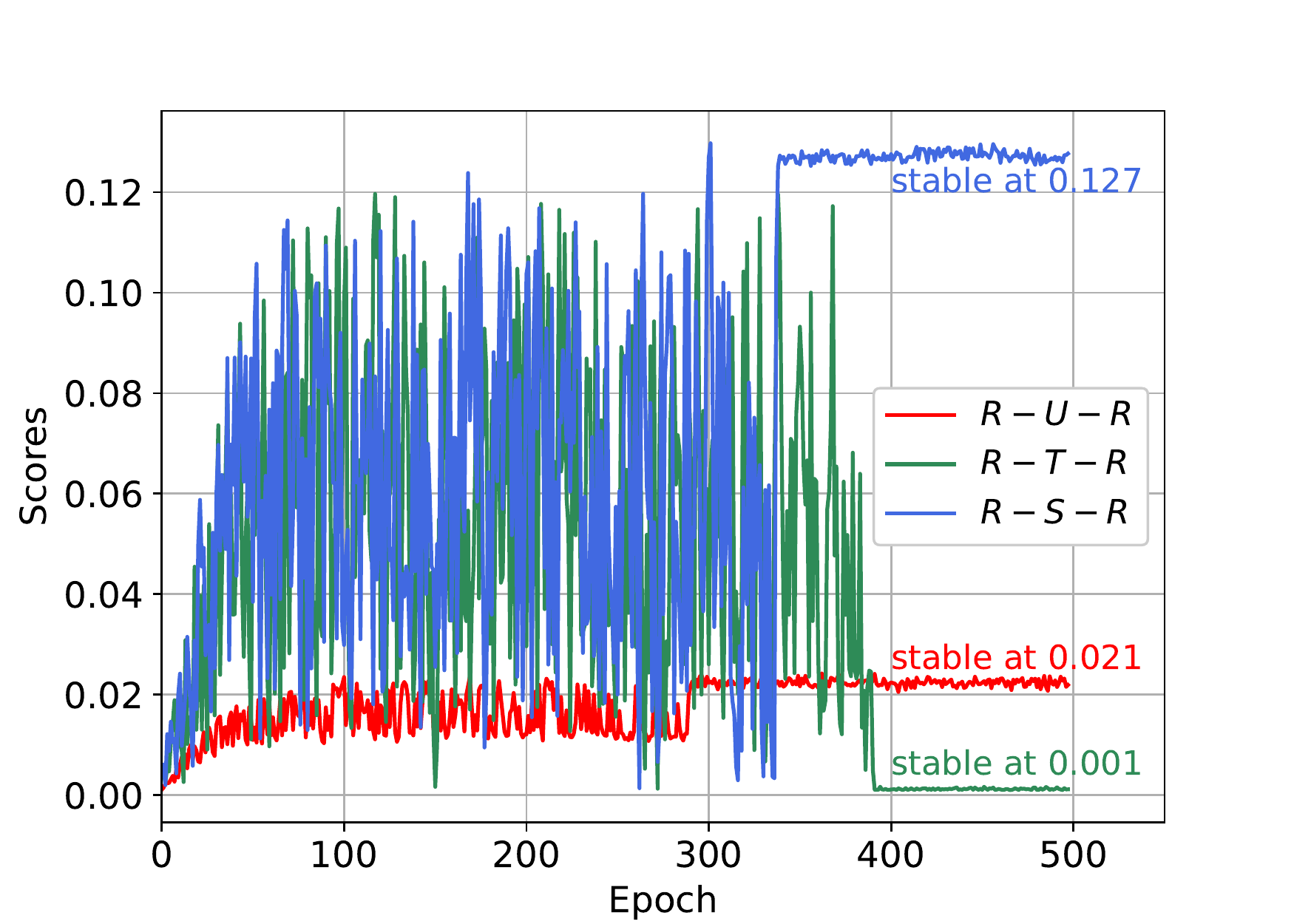}
\end{minipage}%
}%
\subfigure[Thresholds of ROO-GNN on Yelp.]{\label{fig:fraud-training-varitants-b}
\begin{minipage}[h]{0.5\linewidth}
\centering
\includegraphics[width=8.3cm]{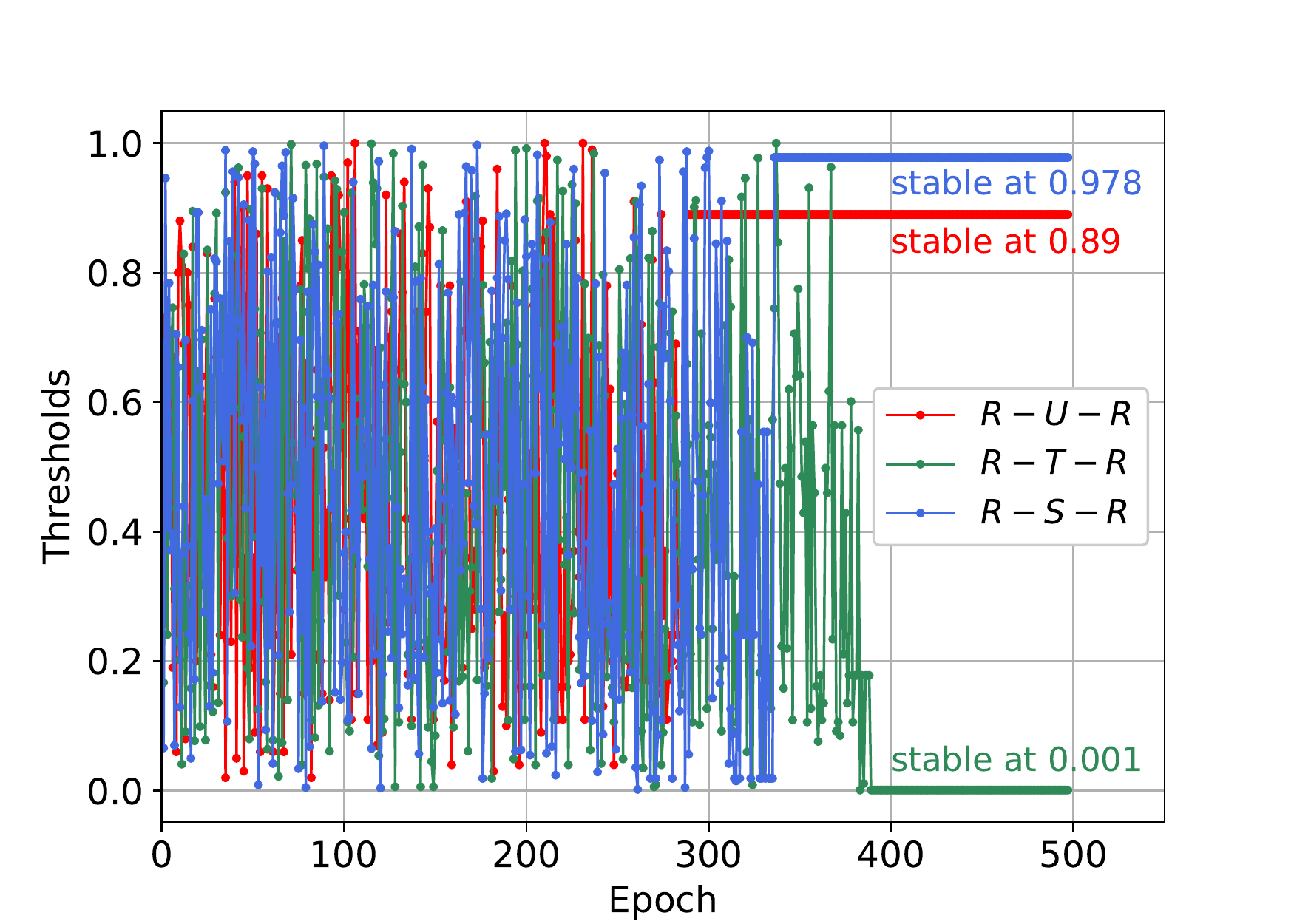}
\end{minipage}%
}%

\subfigure[Scores of BIO-GNN on Yelp.]{\label{fig:fraud-training-varitants-c}
\begin{minipage}[h]{0.5\linewidth}
\centering
\includegraphics[width=8.3cm]{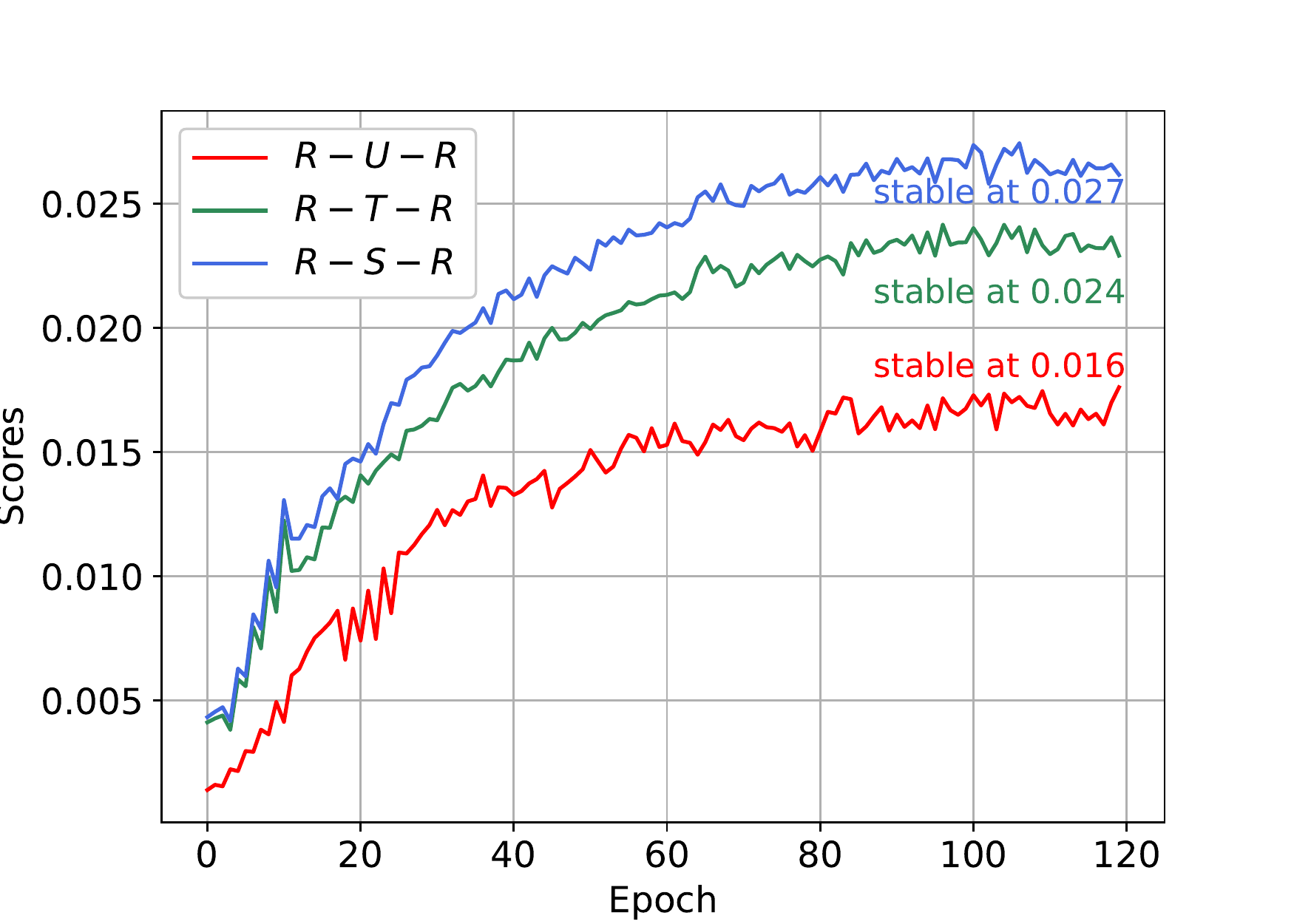}
\end{minipage}%
}%
\subfigure[Thresholds of BIO-GNN on Yelp.]{\label{fig:fraud-training-varitants-d}
\begin{minipage}[h]{0.5\linewidth}
\centering
\includegraphics[width=8.3cm]{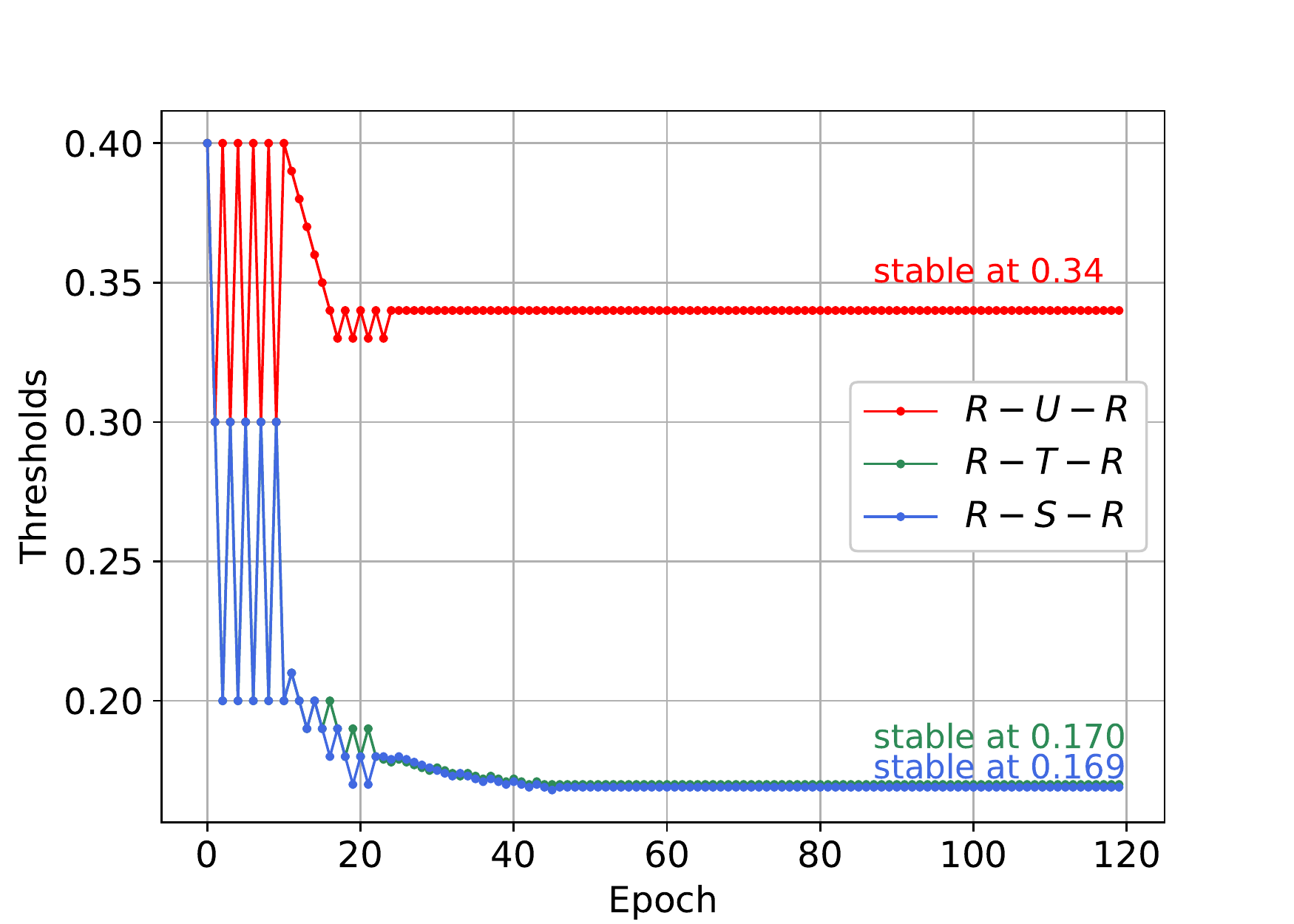}
\end{minipage}%
}%
\centering
\caption{The training scores and thresholds of \RioGNN variants on Yelp.}\label{fig:fraud-training-varitants}
\end{figure}

\subsubsection{Explainable RSRL Training Process}\label{sec:fraud-explainable}
This section focuses on the \RSRL framework and discusses the explainability of the reinforcement learning process in detail.
We show the change process of the \RioGNN's filtering threshold and similarity score of the different relations before convergence during the training process on the Yelp and Amazon datasets.
For a better comparison, we also conduct a similar analysis on ROO-GNN and BIO-GNN variants.

\textbf{Filter Thresholds. }
In Table~\ref{tab:dataset}, we observe that the average feature similarity for most relations in Yelp and Amazon are very high.
However, there is a relation such as R-T-R, which has a low label similarity of $0.05$, which indicates that fraudsters successfully disguised themselves.
For this reason, we propose to filter the lower-ranked neighbors in the Top-p sampling through the filtering threshold.
It can be seen from Figure~\ref{fig:fraud-training-RioGNN-b} and Figure~\ref{fig:fraud-training-RioGNN-c} that the filtering thresholds of the three relations of the Yelp dataset are stable at $[0.99, 0.093, 0.787]$, and the Amazon dataset converges to $[0.071, 0.0712, 0.9997]$.
It shows that the filtering thresholds for different relations eventually converge to different values.
The reason is that the label similarity and feature similarity of different relations are different in the same dataset.
This result also can be verified from Table~\ref{tab:dataset}.
For instance, the label similarity difference between R-U-R and R-T-R is $0.85$, but the feature similarity difference is only $0.04$.
The proposed framework builds a reinforcement learning tree for each relation, uses the similarity between the relations as a reward, and independently finds the appropriate filtering threshold.
Due to the mutual influence between relations, a set of Nash equilibrium filtering thresholds will eventually be obtained, which is a sampling scheme that can best eliminate the interference of fraudsters~\cite{dou2020robust}.
In the Nash equilibrium, it is impossible for all agents to obtain greater rewards only by changing their own strategies.
In addition, in Figure~\ref{fig:fraud-training-RioGNN-b} and Figure~\ref{fig:fraud-training-varitants-b}, Figure~\ref{fig:fraud-training-varitants-d}, we also observe that different models are used under the same dataset, and they converge and combine with different filtering thresholds $[0.99, 0.093, 0.787]$, $[0.89, 0.001, 0.978]$, $[0.34, 0.17, 0.169]$, respectively.
Since the result of reinforcement learning is obtained by the connection strategy of all agents, there may be many different filter threshold combinations at each time the game between relations reaches the Nash equilibrium~\cite{junling1998rlmultinash,pozo2011multinash}.

Figure~\ref{fig:fraud-training-RioGNN-a} and Figure~\ref{fig:fraud-training-RioGNN-c} show the changes in rewards obtained during the filtering threshold learning process for different relations.
We observe that in the Yelp dataset, the relations R-S-R and R-U-R achieve better reward growth in the interest competition of the three relations. 
In contrast, the relation R-T-R eventually stabilizes at a relatively low reward. 
This is consistent with the actual scenario. 
Comments with the same star rating in the same product are important considerations for dividing spam comments, and comments by the same user usually have the same tendency. 
However, reviews published in the same month have a lower impact factor. 
The U-P-U and U-S-U of the similar Amazon dataset are more meaningful than the relation U-V-U. 
This means that users who post similar content are more closely connected, and are considered an important observation factor when making user judgments.

\textbf{Recursive Framework.}
In the right column of Figure~\ref{fig:fraud-training-varitants}, we show the changes in the filtering thresholds of the two variants of \RSRL, ROO-GNN and BIO-GNN on the Yelp dataset.
Comparing \RioGNN in Figure~\ref{fig:fraud-training-RioGNN-b} with ROO-GNN in Figure~\ref{fig:fraud-training-varitants-b}, it can be seen that \RioGNN with the recursive framework of Figure~\ref{fig:fraud-training-RioGNN-b} performs a more accurate filtering threshold search for each depth.
For example, the relation R-U-R is first explored in the interval $[0,1]$ with an accuracy of $0.1$ to obtain a convergence of $0.9$.
Then it searches for the convergence between $[0.85,0.95]$ to obtain the highest accuracy $0.01$ of the relation R-U-R, and get the final convergence $0.99$. 
The difference from \RioGNN is the ROO-GNN model in Figure~\ref{fig:fraud-training-varitants-b}, where only one deep reinforcement learning is performed.
For example, R-U-R directly explores the $[0,1]$ interval with the highest accuracy of $0.01$ and obtains the final convergence value of $0.89$.
In addition to the difference mentioned above, we can also see whether there is a recursive reward change shown in Figure~\ref{fig:fraud-training-RioGNN-a} and Figure~\ref{fig:fraud-training-varitants-a}.
For \RioGNN with recursion, all relations converge fully at $110$ epochs, while ROO-GNN without recursion converges at $390$ epochs.
Furthermore, the rewards obtained by the two methods are basically the same.
This solves the challenge we pose in Section~\ref{sec:challenges}, and explains that reducing the number of actions for each reinforcement learning can effectively accelerate the convergence of the model.
The experimental results also show that the addition of recursion does not bring about a significant loss of accuracy, which is significant.

\begin{figure}[t]
\centering
\subfigure[R-U-R.]{\label{fig:fraud-2layer-a}
\begin{minipage}[h]{0.3333\linewidth}
\centering
\includegraphics[width=5.7cm]{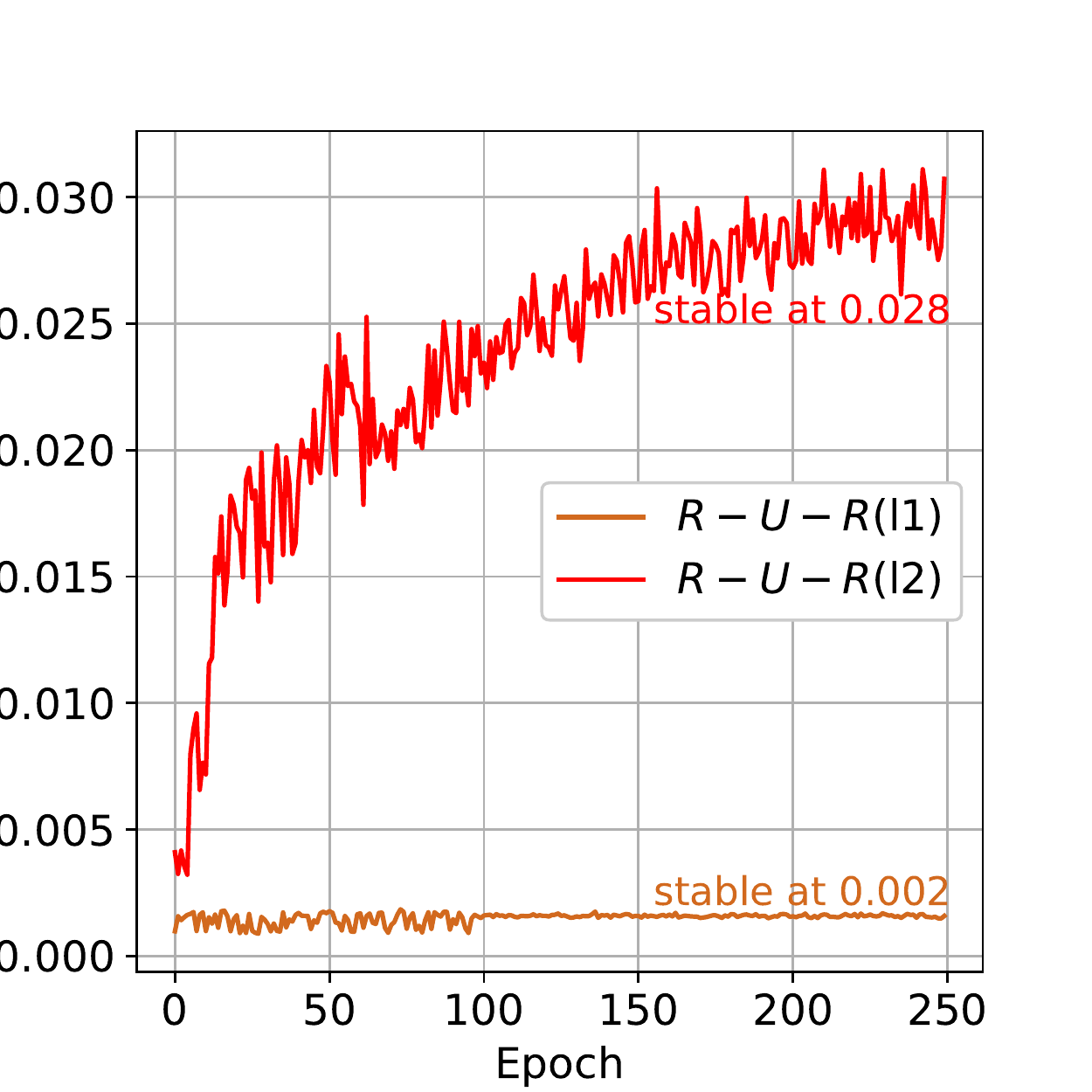}
\end{minipage}%
}%
\subfigure[R-T-R.]{\label{fig:fraud-2layer-b}
\begin{minipage}[h]{0.3333\linewidth}
\centering
\includegraphics[width=5.7cm]{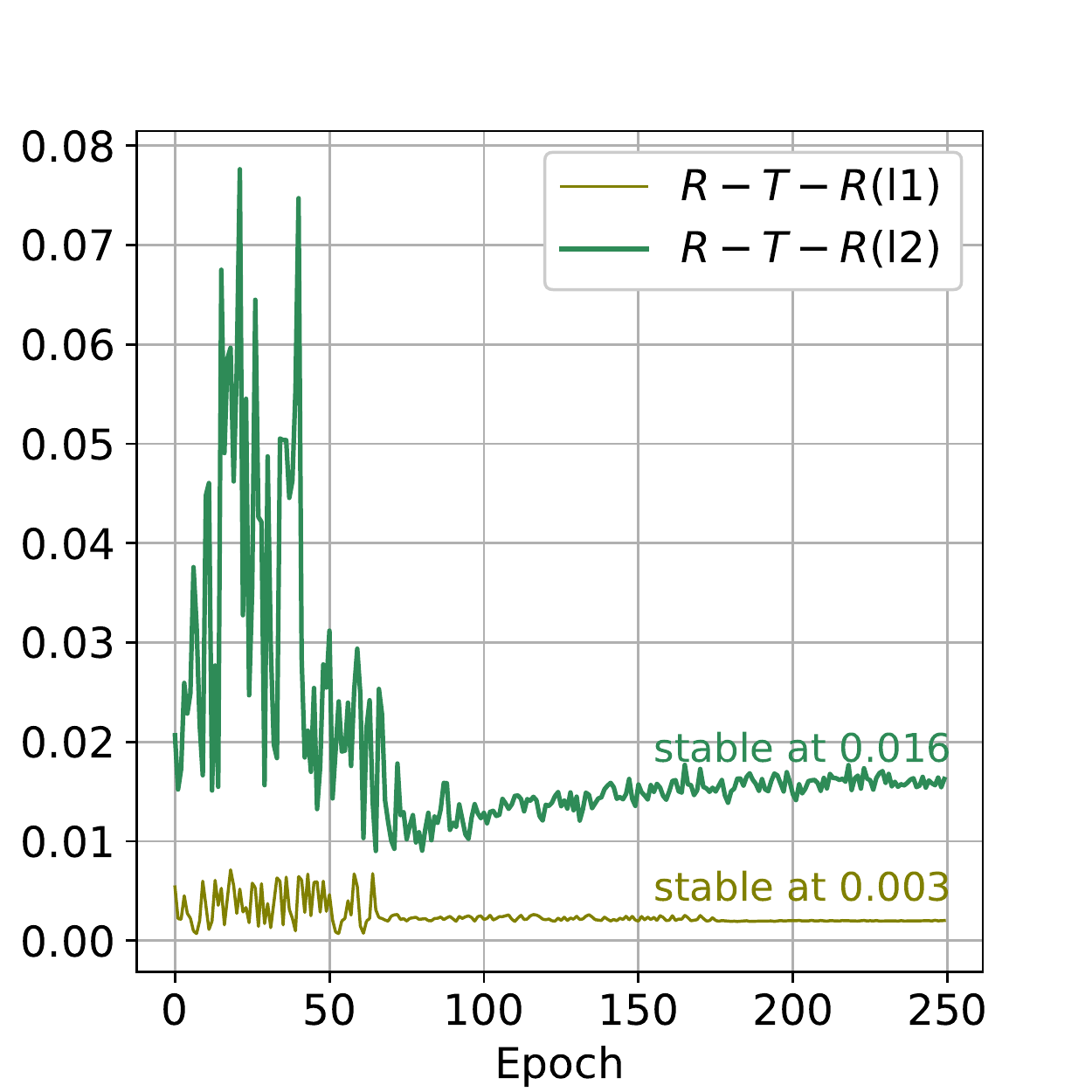}
\end{minipage}%
}%
\subfigure[R-S-R.]{\label{fig:fraud-2layer-c}
\begin{minipage}[h]{0.3333\linewidth}
\centering
\includegraphics[width=5.7cm]{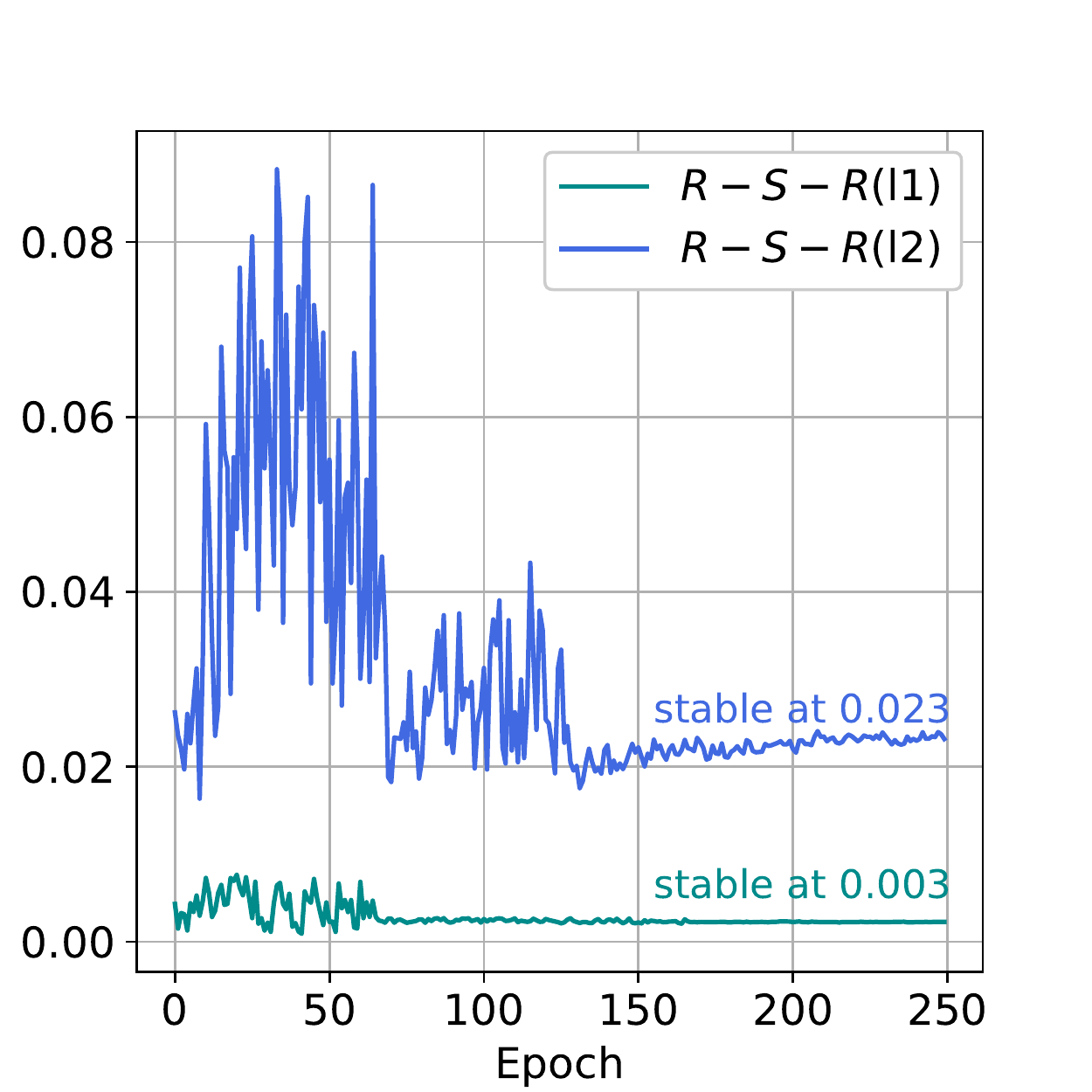}
\end{minipage}%
}%
\centering
\caption{Scores of Multi-Layer \RioGNN on Yelp.}\label{fig:fraud-2layer-scores}
\end{figure}

\textbf{Action Space and Iterative Function. }
In Figure~\ref{fig:fraud-training-RioGNN-b} and Figure~\ref{fig:fraud-training-varitants-d}, we present the BIO-GNN variant and \RioGNN with dynamic iterative function and full action space learning process.
It can be seen that, similar to \RioGNN, BIO-GNN also recursively converges to $0.4$ with an accuracy of $0.1$ for the R-U-R relation and then converges to $0.33$ with an accuracy of $0.01$.
But the difference from \RioGNN is that this method has an average similarity score of $0.045$ for the three relations, which is lower than \RioGNN's $0.021$.
This means that the maximum filtering effect has not been achieved. 
In other respects, compared with the Figure~\ref{fig:fraud-training-RioGNN-a} and Figure~\ref{fig:fraud-training-varitants-a} that have obtained better performance, the R-T-R relation of Figure~\ref{fig:fraud-training-varitants-c} has obtained a higher reward ratio in the competition. That is, the behaviors published in the same month are considered by BIO-GNN to be more important, which is contrary to the reality. So this explains the reason for the lower performance of BIO-GNN.
Meanwhile, these results demonstrates that dynamic learning can be observed globally, thereby obtaining more effect, which proves the conjecture in Section~\ref{sec:rsrl}.

\subsubsection{Effectiveness and Efficiency Evaluation}\label{sec:fraud-efficiency}
Next, we introduce the effectiveness and efficiency of multi-layer similarity perception modules and recursive neighbor selectors in fraud detection tasks on the two datasets.
We use \RioGNN with a two-layer perception structure to train for $500$ epochs on the Yelp dataset, and select the first $250$ epochs to show the changes in the scores of each layer in Figure~\ref{fig:fraud-2layer-scores}.
For the recursive framework, we analyze the AUC change trend of ROO-GNN, BIO-GNN variants and \RioGNN on two datasets within 500 epochs (Figure~\ref{fig:fraud-efficiency}).
The dotted line in the figure indicates the average AUC of each variant after reaching stability.
The specially marked points in the figure show the epoch numbers for different models to reach a certain AUC.

\textbf{Multi-layer Analysis.}
We provide a multi-layer label-aware similarity neighbor measurement scheme to deal with data collections with more complex structures in the future. 
For the datasets in this article, the increase of multiple layers is limited by the datasets (one-hop neighbor information is enough for sampling), and the AUC gain brought by it is limited. 
However, we notice in Figure~\ref{fig:fraud-2layer-scores} that as the number of layers increases, Layer $2$ has a significant improvement in the similarity score of each relation compared to Layer $1$. 
Among them, the similarity score of the relation R-U-R is $14.00$ in the second level compared with the first level. 
In addition, the R-T-R and R-S-R are improved in $5.33$ and $7.66$.
This is because we take the embedding of the neighbors of the previous layer as the input of the second layer, embedding more hops of neighbor relations. 
Rich neighbor information makes the performance of the similarity module better. This also shows that multi-layer joining can be a new deployment scheme in some datasets with insufficient embedded information of one-hop neighbors.

\textbf{Multi-depth Analysis.}
In order to effectively speed up the optimization speed of the filtering threshold of each relation while ensuring the accuracy rate, we propose a recursive reinforcement learning learning framework. 
It can be seen from Figure~\ref{fig:fraud-efficiency} that the \RioGNN model has obvious advantages over ROO-GNN and BIO-GNN in both datasets. 
First of all, we observe the final convergence AUC size.
In the Yelp dataset, \RioGNN and BIO-GNN models are about $4.75\%$ higher than ROO-GNN. 
In the Amazon dataset, \RioGNN is better than the other two models by about $0.8\%$. 
And BIO-GNN finally converged AUC is generally low. 
In addition, in terms of computational efficiency, compared with the ROO-GNN model without recursion, \RioGNN maintains a stable and rapid increase in AUC in both datasets. 
However, ROO-GNN has greater fluctuations. 
In the first $50$ epochs of Yelp and the first $75$ epochs of Amazon, there is no significant difference between them. 
In the Yelp dataset, the speedup of \RioGNN compared to ROO-GNN is $2.14$ when AUC reaches $78.5\%$, $2.99$ when AUC reaches $81.5$, and $2.10$ when AUC reaches $83.1\%$. 
In Amazon, the speedup ratio is $3.19$ when the AUC reaches $95.30\%$, and the speedup ratio is $4.52$ when AUC reaches $95.4\%$. 
Compared with the BIO-GNN with limited action space and fixed strategy, Yelp and Amazon are also observed $8.81$ and $1.71$ times time savings at $78.5\%$ and $95.3\%$ AUC. 
This shows that the proposed recursive framework can achieve good efficiency while maintaining accuracy. And generally, the higher the AUC demand, the better the efficiency. 
The broader and flexible action space and the iterative function that automatically updates have more significant advantages in terms of efficiency and accuracy. 
Finally, we find that different optimization structures have different impact factors for different datasets. 
Due to the smaller scale and lower accuracy requirements of the Yelp dataset, whether it has a recursive structure has little effect on the final convergence AUC. 
But in Amazon, which has a larger scale and higher precision requirements, placing all actions at one depth causes a loss of accuracy. 
This also confirms the conjecture in Section~\ref{sec:rsrl} about the loss of accuracy caused by the excessively large action space, and confirms the advantages of the recursive structure in large-scale datasets.

\subsection{Overall Evaluation of Diagnosis of Diabetes Mellitus Task}\label{sec:diabetes-overall}

\subsubsection{Accuracy Analysis}\label{sec:diabetes-accuracy}
In this section, we conduct experiments to evaluate the accuracy of diagnosis of diabetes mellitus on the MIMIC-III dataset.
As presented in Table~\ref{tab:diabetes_base}, we report the best test results of \RioGNN and various baselines and variants in seven hundred epochs.
It can be observed from the results that \RioGNN performs better than other baselines and variants under most training ratios and indicators.

\textbf{Single-relation vs. Multi-relation.}
In order to further prove the effectiveness of \RioGNN in processing the multi-relational graph, we perform a baseline comparison on the more challenging task of diagnosis of diabetes mellitus.
Table~\ref{tab:diabetes_base} shows the comparative results of \RioGNN and the three types of baselines. 
Compared with the first type of baseline running on a single-relational graph: GCN, GAT and GraphSAGE, the model \RioGNN is significantly better than them on the MIMIC-III dataset by 13.32\%-18.30\%.
This result fully confirms that the fine-grained division of multi-relational graph and hierarchical aggregation based on different relations are very conducive to the completion of the node classification task. 
This point is completely consistent with the experimental conclusions on the fraud detection task. 
Furthermore, in order to show the performance of \RioGNN when dealing with the multi-relational graph, we carry out the second type of baseline comparison experiment.
Although GCT, HSGNN and HAN all run on heterogeneous graphs, and propose different ideas for processing heterogeneous relations, their accuracy rates are at least 8.77\% lower than \RioGNN. 
Due to the high density of neighbor nodes under each relation, the inability to effectively filter interfering nodes also brings difficulties to the final diagnosis task. 
Compared with \RioGNN, it is obvious that they fail to filter out interfering neighbor nodes and produce a sufficiently strong positive effect on the final diagnosis. 
Interestingly, comparing the two types of baselines, we find that although the second type of baselines divides the heterogeneous relations to a certain extent, they are in most cases even worse than the first type of baselines running on a single-relational graph. 
The above results show that when dealing with the multi-relational graph, how to effectively select the appropriate neighbor nodes is particularly important, while \RioGNN achieves this well through parameterized similarity measures and adaptive sampling thresholds. 
When it comes to the third type of baseline, CARE-GNN, we find that its accuracy is significantly higher than the first two types of baselines by 11.73\%-17.70\%, which means that fine-grained and multi-relational division of heterogeneous graphs is very necessary.
Although CARE-GNN realizes automatic filtering and sampling of neighbor nodes to a certain extent, its diagnostic effect is lower than \RioGNN due to the inability to adaptively select the best filtering threshold under each relation. 
The above shows that the proposed RSRL framework finds the optimal filtering threshold under each relation in the recursive process successfully, thus it performs outstandingly in downstream tasks.

\begin{table}[t]
    \setlength{\abovecaptionskip}{0.cm}
    \setlength{\belowcaptionskip}{-0.cm}
    \caption{Diabetes Detection results ($\%$) compared to the baselines.}\label{tab:diabetes_base}
    \centering
    \scalebox{1}{
        \begin{tabular}{p{3cm}<{\centering}|p{1cm}<{\centering}p{1cm}<{\centering}p{1cm}<{\centering}p{1cm}<{\centering}|p{1cm}<{\centering}p{1cm}<{\centering}p{1cm}<{\centering}p{1cm}<{\centering}}
            \hline
            \multirow{4}*{Models}&\multicolumn{8}{c}{\multirow{2}*{\textbf{MIMIC-III}}}\\
            &&&&\multicolumn{1}{c}{}&&&&\\
            \cline{2-9}
            &\multicolumn{4}{c|}{\textbf{AUC}}&\multicolumn{4}{c}{\textbf{Recall}}\\
            &5\%&10\%&20\%&40\%&5\%&10\%&20\%&40\%\\
            \hline
            \multirow{1}*{\textbf{GCN}}&\multirow{1}*{65.37}&\multirow{1}*{65.39}&\multirow{1}*{64.93}&\multirow{1}*{65.13}&\multirow{1}*{60.85}&\multirow{1}*{61.24}&\multirow{1}*{60.31}&\multirow{1}*{60.99}\\
             \multirow{1}*{\textbf{GAT}}&\multirow{1}*{63.67}&\multirow{1}*{63.09}&\multirow{1}*{63.11}&\multirow{1}*{64.26}&\multirow{1}*{62.13}&\multirow{1}*{62.75}&\multirow{1}*{63.12}&\multirow{1}*{63.45}\\
             \multirow{1}*{\textbf{GraphSAGE}}&\multirow{1}*{65.91}&\multirow{1}*{66.28}&\multirow{1}*{65.82}&\multirow{1}*{65.34}&\multirow{1}*{63.80}&\multirow{1}*{63.88}&\multirow{1}*{63.82}&\multirow{1}*{63.99}\\
             \hline
             \multirow{1}*{\textbf{GCT}}&\multirow{1}*{62.97}&\multirow{1}*{63.45}&\multirow{1}*{64.33}&\multirow{1}*{65.14}&\multirow{1}*{60.08}&\multirow{1}*{61.23}&\multirow{1}*{61.39}&\multirow{1}*{62.25}\\
             \multirow{1}*{\textbf{HSGNN}}&\multirow{1}*{63.27}&\multirow{1}*{65.68}&\multirow{1}*{65.03}&\multirow{1}*{67.87}&\multirow{1}*{61.24}&\multirow{1}*{63.87}&\multirow{1}*{64.08}&\multirow{1}*{65.39}\\
            \multirow{1}*{\textbf{HAN}}&\multirow{1}*{62.79}&\multirow{1}*{63.26}&\multirow{1}*{64.94}&\multirow{1}*{65.13}&\multirow{1}*{61.15}&\multirow{1}*{61.27}&\multirow{1}*{62.38}&\multirow{1}*{63.02}\\
            \hline
             \multirow{1}*{\textbf{GraphNAS$^{H}$}}&\multirow{1}*{64.03}&\multirow{1}*{65.28}&\multirow{1}*{65.08}&\multirow{1}*{65.59}&\multirow{1}*{62.18}&\multirow{1}*{63.94}&\multirow{1}*{64.05}&\multirow{1}*{65.14}\\
             \multirow{1}*{\textbf{GraphNAS}}&\multirow{1}*{65.76}&\multirow{1}*{67.40}&\multirow{1}*{67.79}&\multirow{1}*{68.05}&\multirow{1}*{64.28}&\multirow{1}*{66.88}&\multirow{1}*{67.31}&\multirow{1}*{67.93}\\
            \multirow{1}*{\textbf{Policy-GNN$^{H}$}}&\multirow{1}*{66.30}&\multirow{1}*{67.19}&\multirow{1}*{67.70}&\multirow{1}*{67.98}&\multirow{1}*{63.28}&\multirow{1}*{66.05}&\multirow{1}*{67.12}&\multirow{1}*{68.18}\\
            \multirow{1}*{\textbf{Policy-GNN}}&\multirow{1}*{67.45}&\multirow{1}*{68.55}&\multirow{1}*{69.01}&\multirow{1}*{69.59}&\multirow{1}*{64.73}&\multirow{1}*{67.11}&\multirow{1}*{68.32}&\multirow{1}*{69.02}\\
            \hline
            \multirow{1}*{\textbf{CARE-GNN}}&\multirow{1}*{77.64}&\multirow{1}*{80.22}&\multirow{1}*{80.81}&\multirow{1}*{81.26}&\multirow{1}*{69.17}&\multirow{1}*{71.58}&\multirow{1}*{72.08}&\multirow{1}*{72.68}\\
            \hline
            \multirow{1}*{\textbf{\RioGNN}}&\multirow{1}*{\textbf{79.23}}&\multirow{1}*{\textbf{80.92}}&\multirow{1}*{\textbf{81.23}}&\multirow{1}*{\textbf{82.56}}&\multirow{1}*{\textbf{71.23}}&\multirow{1}*{\textbf{72.64}}&\multirow{1}*{\textbf{72.93}}&\multirow{1}*{\textbf{74.01}}\\
            \hline
        \end{tabular}
    }
\end{table}

\textbf{Heterogeneous vs. Multi-relation. }
In synchronization with the fraud detection task, we also analyze the performance of GraphNAS and Policy-GNN models in heterogeneous graph and multi-relational graph in the disease detection task. 
From Table~\ref{tab:diabetes_base}, the same phenomenon as the fraud detection task can be obtained, that is, the introduction of the multiple relational graph on the Mimic dataset also has certain advantages compared with the heterogeneous graph. 
However, unlike Yelp and Amazon, GraphNAS and Policy-GNN are generally better than other baselines in terms of accuracy. 
We believe that the cause is the balanced features and labels of the Mimic dataset with different relations, and the higher aggregation requirements of the large-scale dataset for neighbor information.

\textbf{Training Percentage.}
In order to measure the impact of the training ratio on the classification accuracy in the diagnosis of diabetes mellitus task, we still set four different training ratios ranging from 5\% to 40\%. 
It can be seen from Table~\ref{tab:diabetes_base} that the classification accuracy of \RioGNN shows a steady upward trend as the training ratio increases. 
This shows that the training process of \RioGNN has a very positive effect on the final classification accuracy. 
It can be seen intuitively, \RioGNN has successfully achieved a good performance improvement in the process of reinforcement learning recursion by supervised signals, which is also in line with expectations. 
However, the accuracy of some baselines changes unrelated to the training percentage, which means that their training methods have great limitations in this task and fail to improve the performance of their models by increasing the supervised signal learning process. 
It also verifies that \RioGNN has better explainability and can continuously improve the accuracy of diagnosis through learning more node features. 
Consistent with its performance under the previous fraud detection task, \RioGNN still maintains strong stability and explainability and its accuracy rate surpasses the others at each training ratio.

\begin{table}[t]
    \setlength{\abovecaptionskip}{0.cm}
    \setlength{\belowcaptionskip}{-0.cm}
    \caption{Diabetes diagnosis classification results ($\%$) compared to \RioGNN variants.}\label{tab:diabetes-variants}
    \centering
    \scalebox{1}{
        \begin{tabular}{p{3cm}<{\centering}|p{2cm}<{\centering}p{2cm}<{\centering}}
            \hline
            \multicolumn{1}{p{3cm}<{\centering}|}{\multirow{2}*{Models}}&\multicolumn{2}{p{4cm}<{\centering}}{\textbf{MIMIC-III}}\\
            \cline{2-3}
            \multicolumn{1}{p{3cm}<{\centering}|}{}&\textbf{AUC}&\textbf{Recall}\\
            \hline
            \multicolumn{1}{p{3cm}<{\centering}|}{\RioGNN$_{2l}$}&81.06&72.28\\
            \hline
            \multicolumn{1}{p{3cm}<{\centering}|}{BIO-GNN}&81.29&72.75\\
            \multicolumn{1}{p{3cm}<{\centering}|}{ROO-GNN}&81.01&72.34\\
            \hline
            \multicolumn{1}{p{3cm}<{\centering}|}{RIO-Att}&80.96&72.16\\
            \multicolumn{1}{p{3cm}<{\centering}|}{RIO-Weight}&81.04&72.58\\
            \multicolumn{1}{p{3cm}<{\centering}|}{RIO-Mean}&80.31&77.42\\
            \hline
            \multicolumn{1}{p{3cm}<{\centering}|}{\RioGNN}&\textbf{81.36}&\textbf{72.84}\\
            \hline
        \end{tabular}
    }
\end{table}

\begin{figure}[t]
\centering
\subfigure[AUC of \RioGNN,ROO-GNN and BIO-GNN on MIMIC-III.]{
\begin{minipage}[t]{1\linewidth}
\centering
\includegraphics[width=15.3cm]{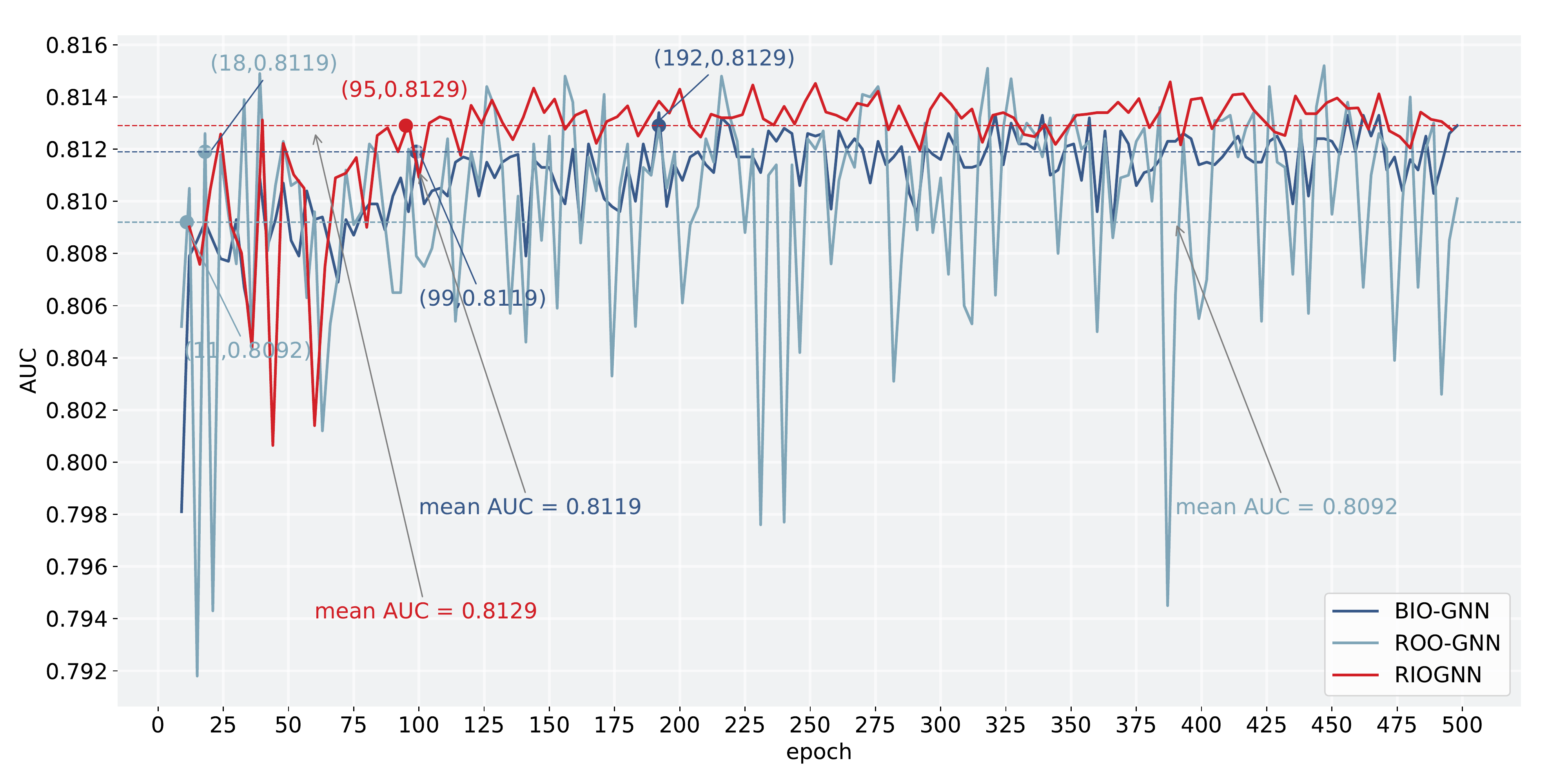}
\end{minipage}%
}%
\centering
\caption{The effectiveness and necessity of the RSRL framework.}\label{fig:RSRL provement_mic}
\end{figure}

\textbf{\RioGNN Variants in Classification.}
In order to further verify the impact of the proposed new mechanism on different tasks, we compare different performances of several variants of \RioGNN under the discrete strategy in the context of diabetes diagnosis. 
We show the experimental results of diagnosis of diabetes mellitus on the MIMIC-III dataset in Table~\ref{tab:diabetes-variants}.
First of all, we compare three variants based on different aggregation methods on the MIMIC-III dataset. 
The results show that \RioGNN is better than the other three variants, while RIO-Weight has better results than the other two, which is consistent with the results on Yelp dataset.
We find that the choice of aggregation method is usually based on a specific dataset. 
As for different dataset, different relations have different ways of influencing the results, which lead to different aggregation methods.
However, whether it is for Yelp, Amazon or MIMIC-III, the results show that \RioGNN is superior to all aggregation variants, while RIO-Weight is second only.
This point has a certain degree of universality.
Next, we also focus on the performance of the \RioGNN variant with two-layer structure, \RioGNN$_{2l}$.
Consistent with the previous results on Yelp and Amazon, the two-layer architecture doesn't bring very good performance improvements to the model on MIMIC-III. 
Compared with \RioGNN, it can be found that the increase in the number of layers does not improve the final accuracy, indicating that the use of a single-layer structure based on the label-aware similarity measure for neighbor selection is optimal. 
However, in conjunction with Table~\ref{tab:dataset}, it is noted that MIMIC-III dataset is denser than Yelp and Amazon, thus the effect of \RioGNN$_{2l}$ is very close to that of \RioGNN. 
This shows that for denser relations, the second-order neighbors found by \RioGNN$_{2l}$ are more effective in the final result. 
To some extent, for too dense multi-relation graphs, second-order neighbors can be used as supplementary information for first-order neighbors.
Finally, for the Label-aware Similarity Measure section, we observe the variant BIO-GNN without adaptive strategy optimization reinforcement learning algorithm and the single-depth structure variant ROO-GNN without the recursive framework for the RSRL framework again on MIMIC-III. 
It can be found from Table~\ref{tab:diabetes-variants} and Table~\ref{tab:diabetes_base} that ROO-GNN performs significantly better than most baselines and variants, second only to \RioGNN. 
This shows that Bernoulli Multi-armed Bandit (BMAB) algorithm of discrete strategy has strong adaptability in the process of recursively selecting the filtering threshold of relations. 
In contrast, the accuracy rate of ROO-GNN is 0.28\% lower than BIO-GNN. 
Apart from that, Figure~\ref{fig:RSRL provement_mic} shows that as the training epoch increases, ROO-GNN fluctuates greatly in the range of 79.48\%-81.43\%, and can never be stable in a fixed smaller interval as \RioGNN or BIO-GNN. 
That is to say, the multi-depth structure of RioGNN brings better convergence speed and excellent stability than the single-depth variant ROO-GNN while maintaining a higher accuracy rate.

\begin{table}[h]
    \setlength{\abovecaptionskip}{0.cm}
    \setlength{\belowcaptionskip}{-0.cm}
    \caption{Diabetes diagnosis clustering results ($\%$) compared to \RioGNN variants.}\label{tab:diabetes-cluster}
    \centering
    \scalebox{1}{
        \begin{tabular}{c|c|c|cc|ccc|c}
            \hline
            \multicolumn{1}{c|}{\multirow{1}*{Dataset}}&\multicolumn{1}{c|}{\multirow{1}*{Metric}}&\multirow{1}*{\RioGNN$_{2l}$}&BIO-GNN&ROO-GNN&RIO-Att&RIO-Weight&RIO-Mean&\multirow{1}*{\RioGNN}\\
            \hline
            \multirow{2}*{\textbf{MIMIC-III}}&\textbf{NMI}&19.01&19.81&19.13&17.17&\textbf{20.22}&19.86&20.10\\
            &\textbf{ARI}&7.15&8.27&8.11&6.24&9.01&7.51&\textbf{10.03}\\
            \hline
        \end{tabular}
    }
\end{table}

\textbf{\RioGNN Variants in Clustering. }
Similar to fraud detection, we also perform cluster analysis in the task of disease diagnosis. 
The best results of NMI and ARI within $700$ epochs are recorded in Table~\ref{tab:diabetes-cluster}. 
It can be seen that compared with BIO-GNN and ROO-GNN, \RioGNN brings at least $0.29\%$ and $1.76\%$ increase in NMI and ARI respectively. 
This proves that the RSRL framework also brings accuracy optimization for dense datasets in the clustering task. 
In addition, RioGNN has better performance on ARI indicators for variants that use different methods for aggregation. 
For the NMI indicator, the RIO-Weight effect has been improved. 
We believe this is because the MIMIC-III dataset has a smaller difference in weight between the relationships compared with Yelp and Amazon. 
Weight can distinguish different classes better than \RioGNN, which directly uses the filtering threshold as the aggregation parameter.

\begin{figure}[h]
\centering
\subfigure[Scores of \RioGNN on MIMIC-III.]{\label{fig:MIMIC-training-RioGNN-a}
\begin{minipage}[t]{0.5\linewidth}
\centering
\includegraphics[width=8.3cm]{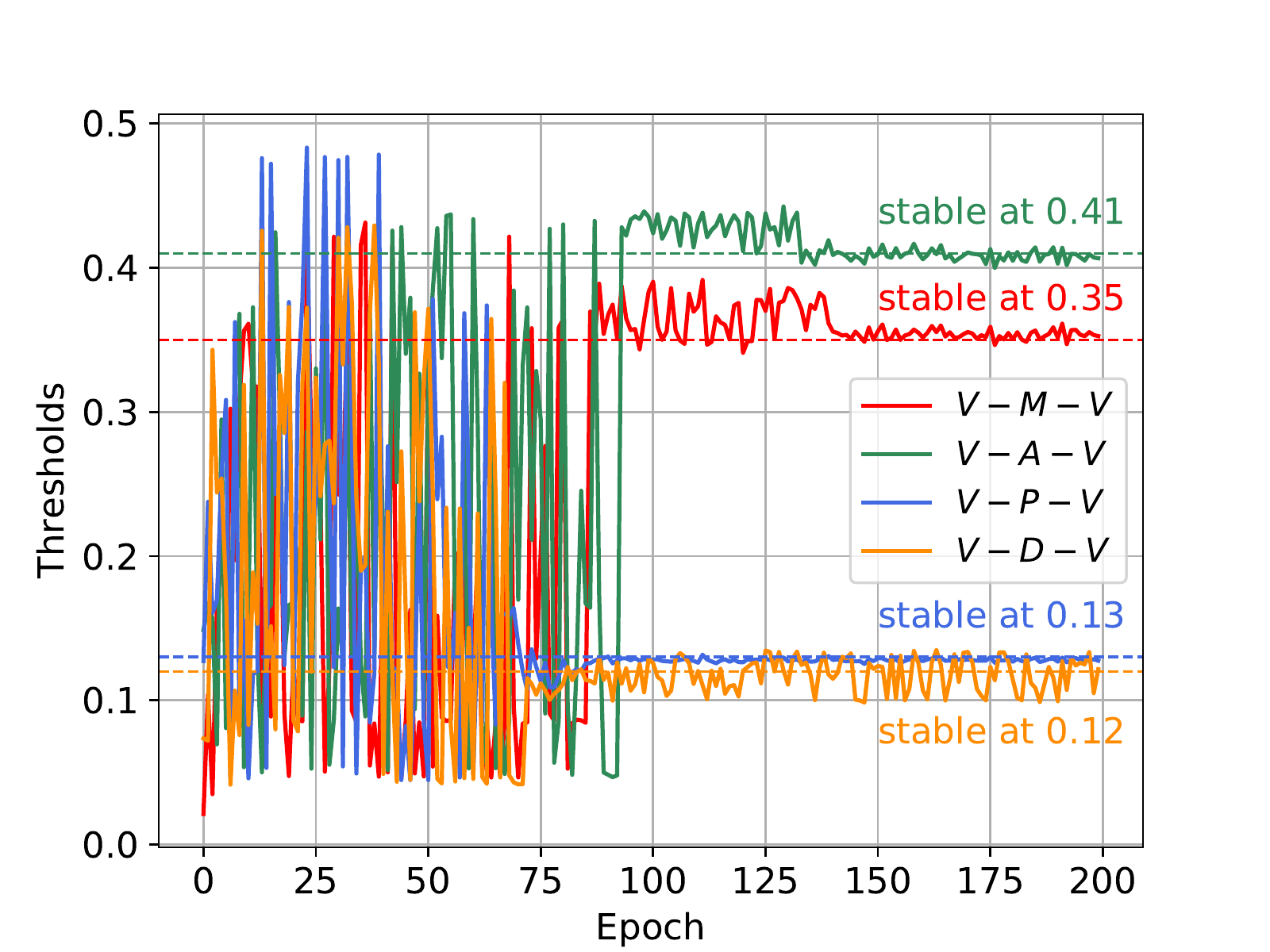}
\end{minipage}%
}%
\subfigure[Thresholds of \RioGNN on MIMIC-III.]{\label{fig:MIMIC-training-RioGNN-b}
\begin{minipage}[t]{0.5\linewidth}
\centering
\includegraphics[width=8.3cm]{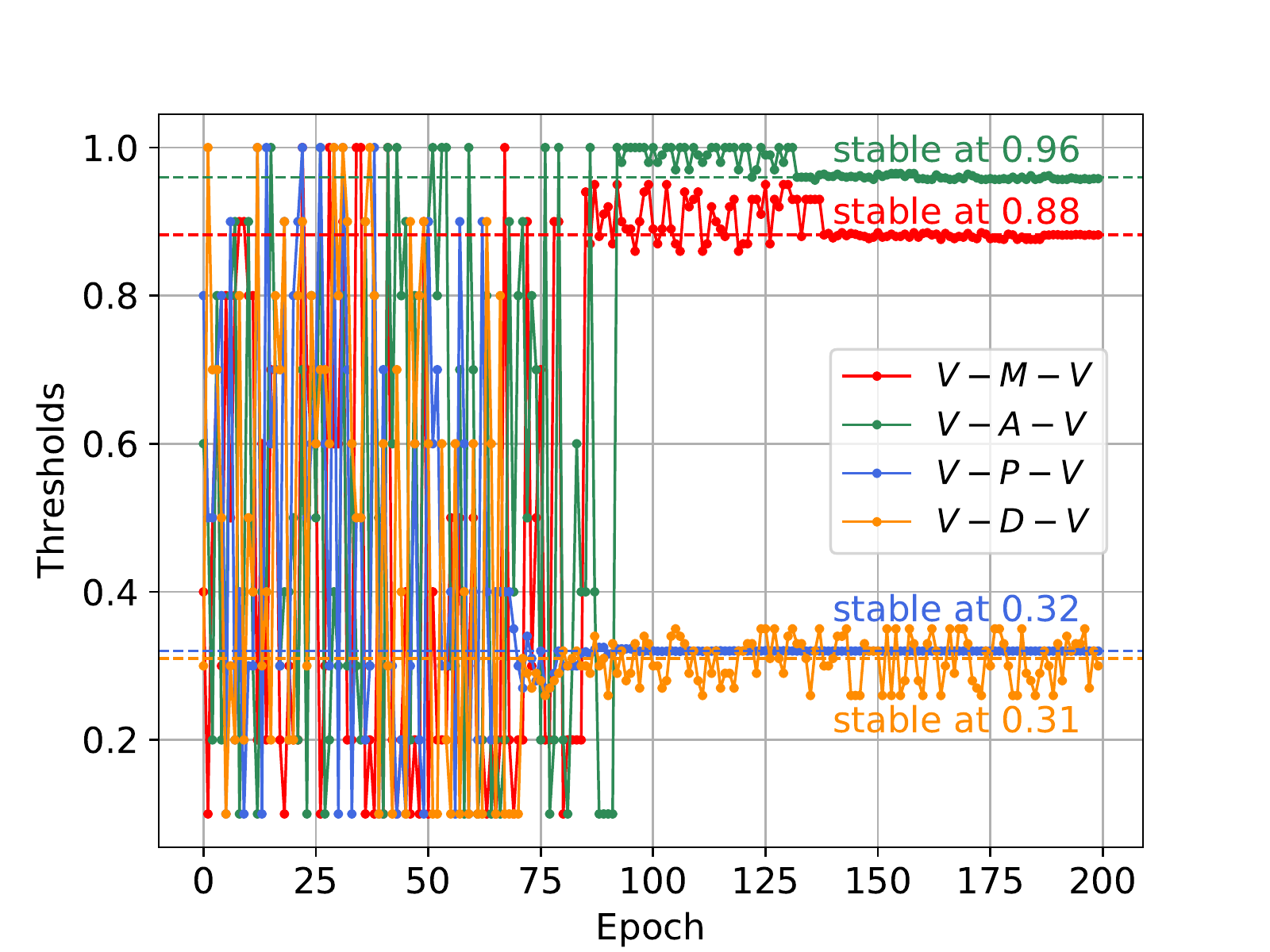}
\end{minipage}%
}%

\subfigure[Scores of BIO-GNN on MIMIC-III.]{\label{fig:MIMIC-training-RioGNN-c}
\begin{minipage}[t]{0.5\linewidth}
\centering
\includegraphics[width=8.3cm]{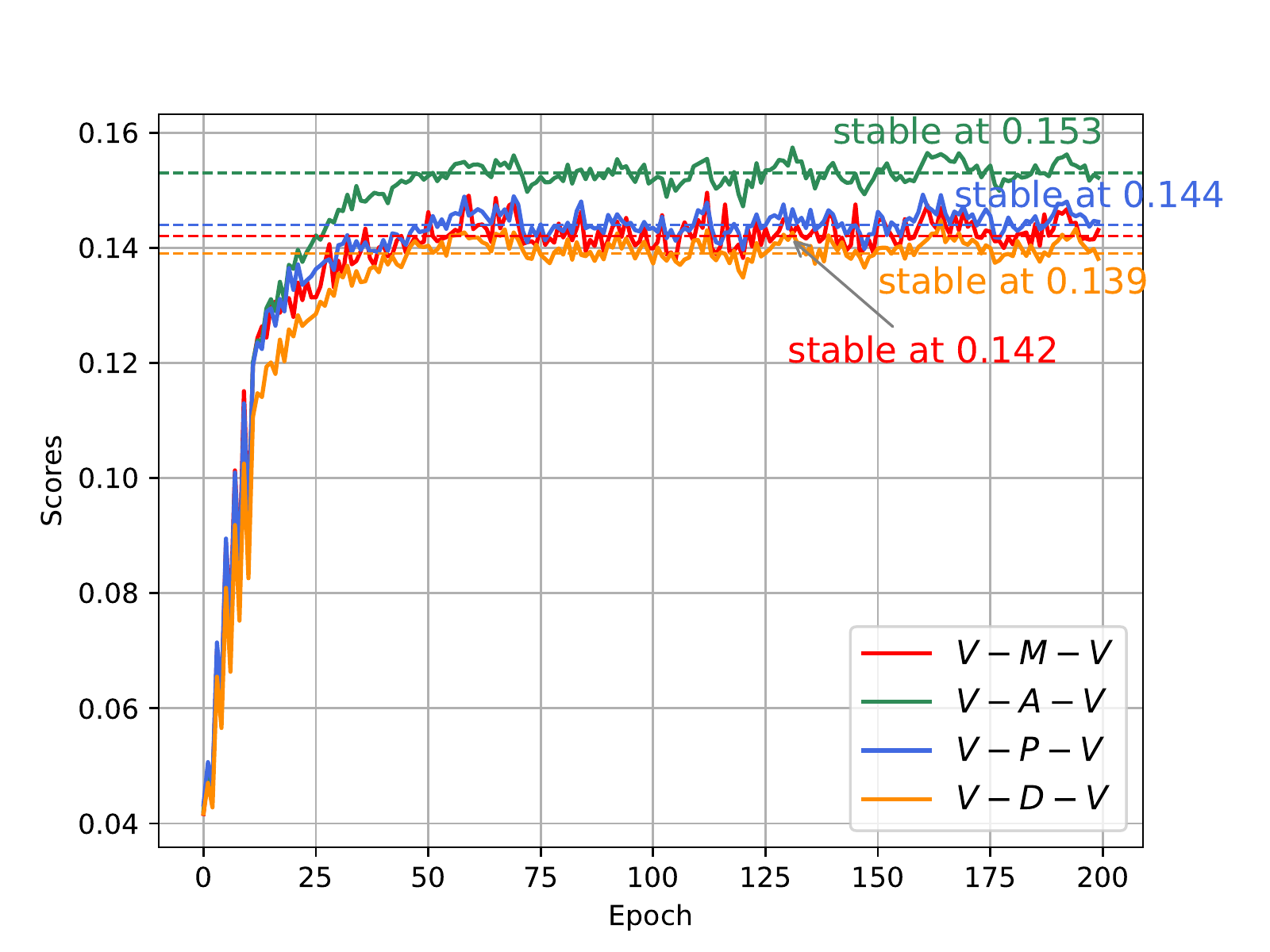}
\end{minipage}%
}%
\subfigure[Thresholds of BIO-GNN on MIMIC-III.]{\label{fig:MIMIC-training-RioGNN-d}
\begin{minipage}[t]{0.5\linewidth}
\centering
\includegraphics[width=8.3cm]{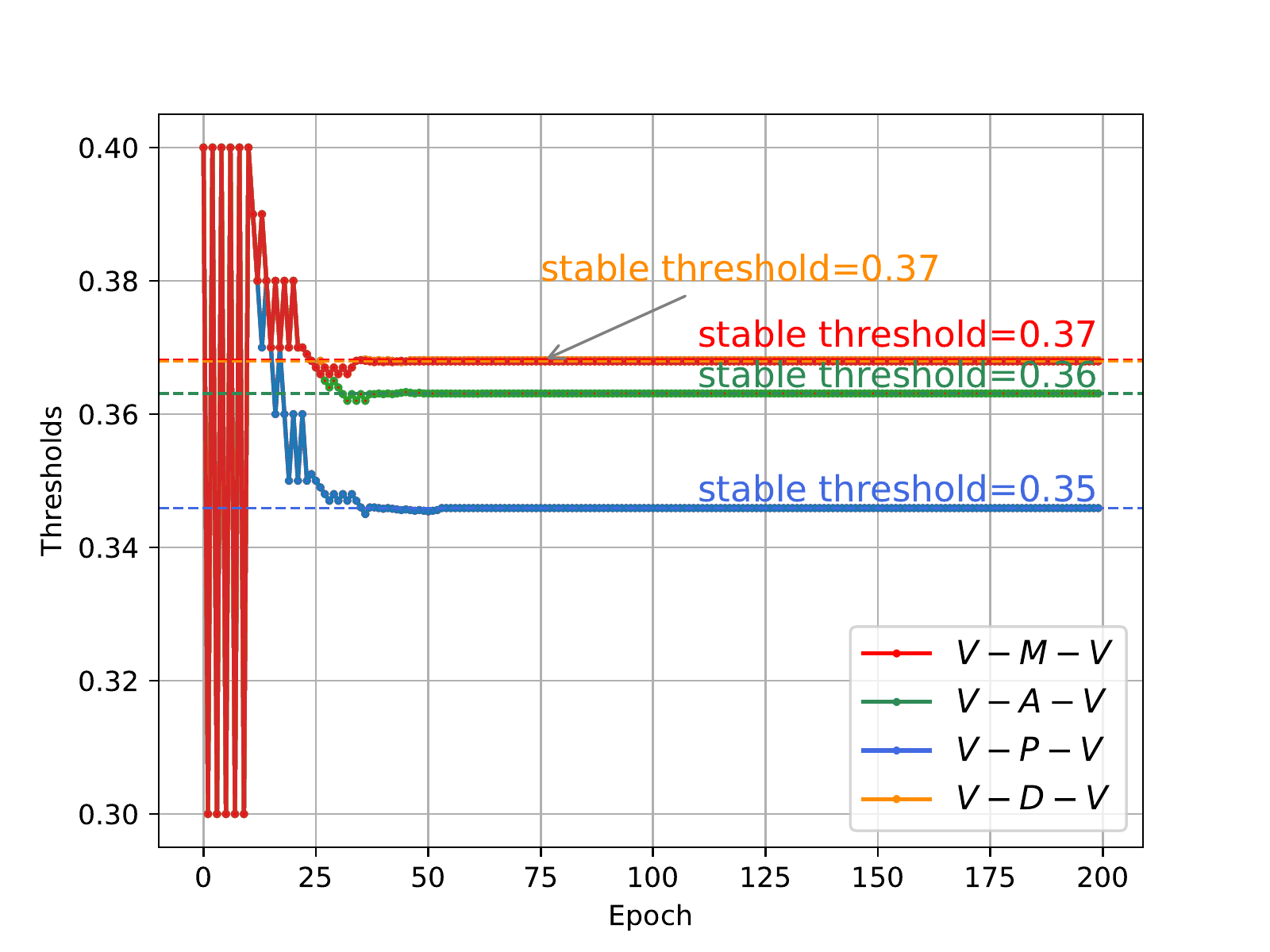}
\end{minipage}%
}%
\centering
\caption{The training scores and thresholds of \RioGNN vs BIO-GNN on MIMIC-III.}\label{fig:MIMIC-training-RioGNN}
\end{figure}

\subsubsection{Explainable Reinforcement Learning Training Process}
This section focuses on the process of reinforcement learning and explains the convergence process of proposed \RioGNN on the MIMIC-III dataset.
We also present a comparative analysis of different variants to further explain the applicability of \RioGNN.

\textbf{The Effectiveness and Necessity of the \RSRL Framework.}
This part demonstrates the validity and necessity of the proposed framework by comparing \RioGNN with variants and its preliminary version CARE-GNN.
It can be seen from the Figure~\ref{fig:RSRL provement_mic} that \RioGNN performs better than CARE-GNN in almost every epoch, which proves that the \RSRL framework has a positive effect on the final classification results on MIMIC-III.
\RioGNN can effectively filter suspected nodes by building a reinforcement learning tree for each relation and identify suspected nodes more accurately. 
Compared with \RioGNN, the accuracy of ROO-GNN in Figure~\ref{fig:MIMIC-training-RioGNN} fluctuates significantly, and it is difficult to converge to a stable range, which proves the necessity to establish a recursive process.
Through the \RSRL process, the classification accuracy can be maintained in a relatively stable range. 
The experimental results show that \RioGNN with the recursive framework performs better through a more accurate filtering threshold search for each depth.
While ROO-GNN without recursion not only fails to converge within a finite number of epochs but also causes a loss of accuracy.

\textbf{Filter Thresholds. }
To further test the proposed model's filtering performance against suspected neighbors, we deliberately extract four denser relations when designing the MIMIC-III dataset.
Each relation is at least an order of magnitude higher than the Yelp or Amazon dataset, which means that the MIMIC-III dataset is more challenging for \RioGNN's filtering performance.
It can be seen from Figure~\ref{fig:MIMIC-training-RioGNN-b} and Figure~\ref{fig:MIMIC-training-RioGNN-d} that the filtering thresholds of the four relations of \RioGNN on the MIMIC-III dataset are stable at [0.88, 0.96, 0.32, 0.26], while BIO-GNN are stable at [0.35, 0.37, 0.36, 0.37].
Considering that the label similarity and feature similarity of different relations are different, the model's filtering strength for different relations is not the same. 
It is worth noting that there is a certain commonality between the relation filtering strength on \RioGNN and BIO-GNN.
Under RioGNN, the convergence thresholds of V-A-V and V-M-V are relatively similar, while V-P-V and V-D-V are relatively similar. 
Generally speaking, relations with high filtering thresholds can bring more positive guidance to the diagnosis result and are more explainable. 
From another perspective, a higher filtering threshold can prove the explainability and correctness of the selected relation, which is more conducive to us intuitively judging whether the choice of the relation is appropriate enough. 
It is clear from Figure~\ref{fig:MIMIC-training-RioGNN} that these four relations of \RioGNN converge to a stable value within 100 epochs.
This indicates that the algorithm can obtain a set of Nash equalization filter thresholds in a very limited epoch through RSRL framework, and thus has good stability and high efficiency.

\begin{table}[t]
    \setlength{\abovecaptionskip}{0.cm}
    \setlength{\belowcaptionskip}{-0.cm}
    \caption{Results ($\%$) compared to different RL algorithms and strengthening strategies.}\label{tab:RL_algorithms}
    \centering
    \scalebox{1}{
        \begin{tabular}{p{0.5cm}<{\centering}|p{2cm}<{\centering}||p{1.5cm}<{\centering}p{1.5cm}<{\centering}|p{2cm}<{\centering}}
            \hline
            \multicolumn{2}{c||}{Methods}&\multirow{1}*{\textbf{Yelp}}&\multirow{1}*{\textbf{Amazon}}&\multirow{1}*{\textbf{MIMIC-III}}\\
            \hline
            \multirow{3}*{\rotatebox{90}{Discrete}}&AC~\cite{konda2000actor}&83.54&\textbf{96.19}&81.36\\
            &DQN~\cite{mnih2015dqn}&84.08&95.13&80.96\\
            &PPO~\cite{schulman2017proximal}&80.52&94.99&80.98\\
            \hline
            \multirow{4}*{\rotatebox{90}{Continuous}}&AC~\cite{konda2000actor}&81.31&94.72&80.98\\
            &DDPG~\cite{lillicrap2019ddpg}&83.80&95.39&81.17\\
            &SAC~\cite{haarnoja2018soft}&80.42&94.76&80.87\\
            &TD3~\cite{scott2018td3}&\textbf{84.18}&95.11&\textbf{81.51}\\
            \hline
        \end{tabular}
    }
\end{table}

\begin{figure}[t]
\centering
\subfigure[Yelp.]{\label{fig:alapha-a}
\begin{minipage}[t]{0.333\linewidth}
\centering
\includegraphics[width=5.3cm]{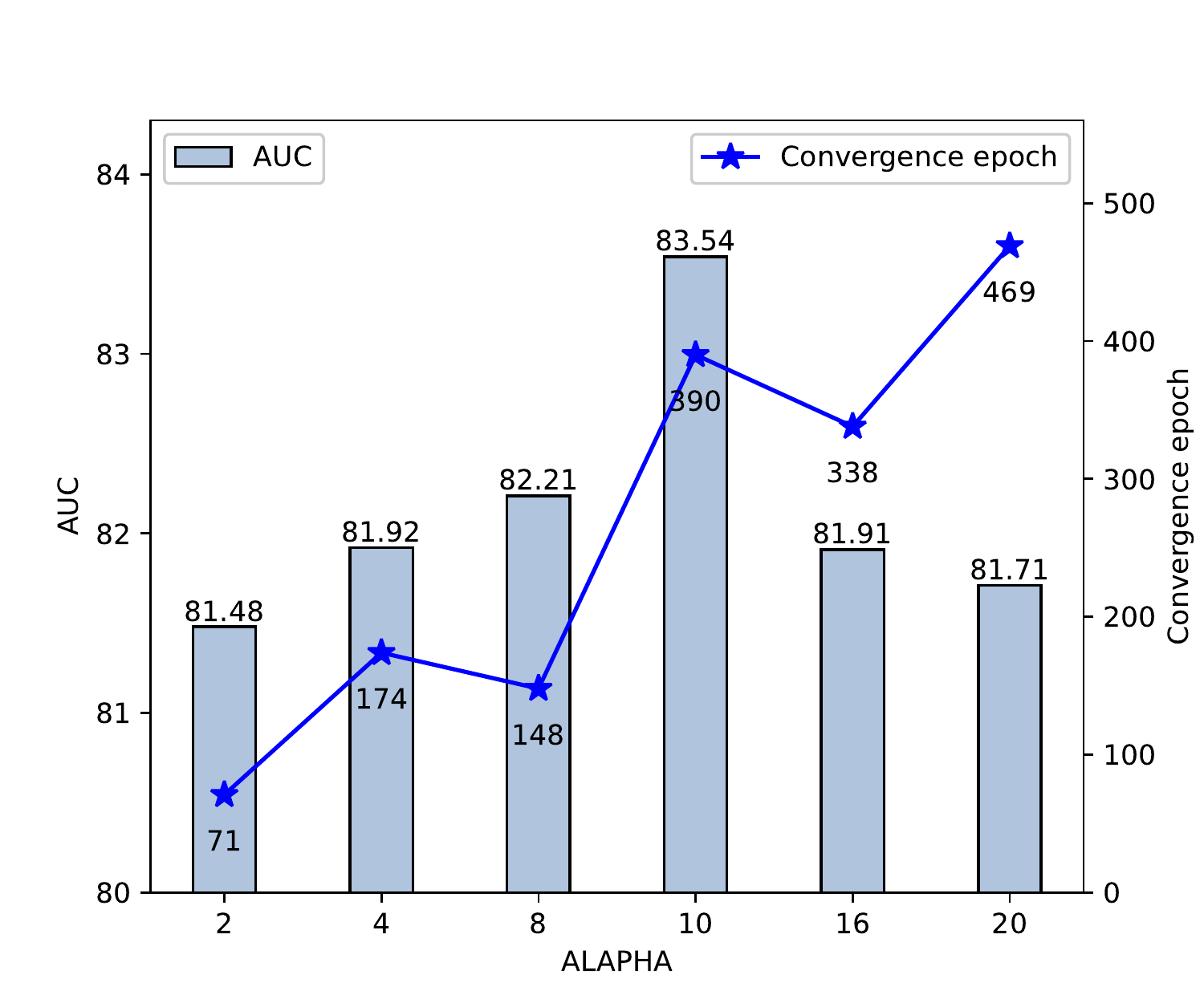}
\end{minipage}%
}%
\subfigure[Amazon.]{\label{fig:alapha-b}
\begin{minipage}[t]{0.333\linewidth}
\centering
\includegraphics[width=5.3cm]{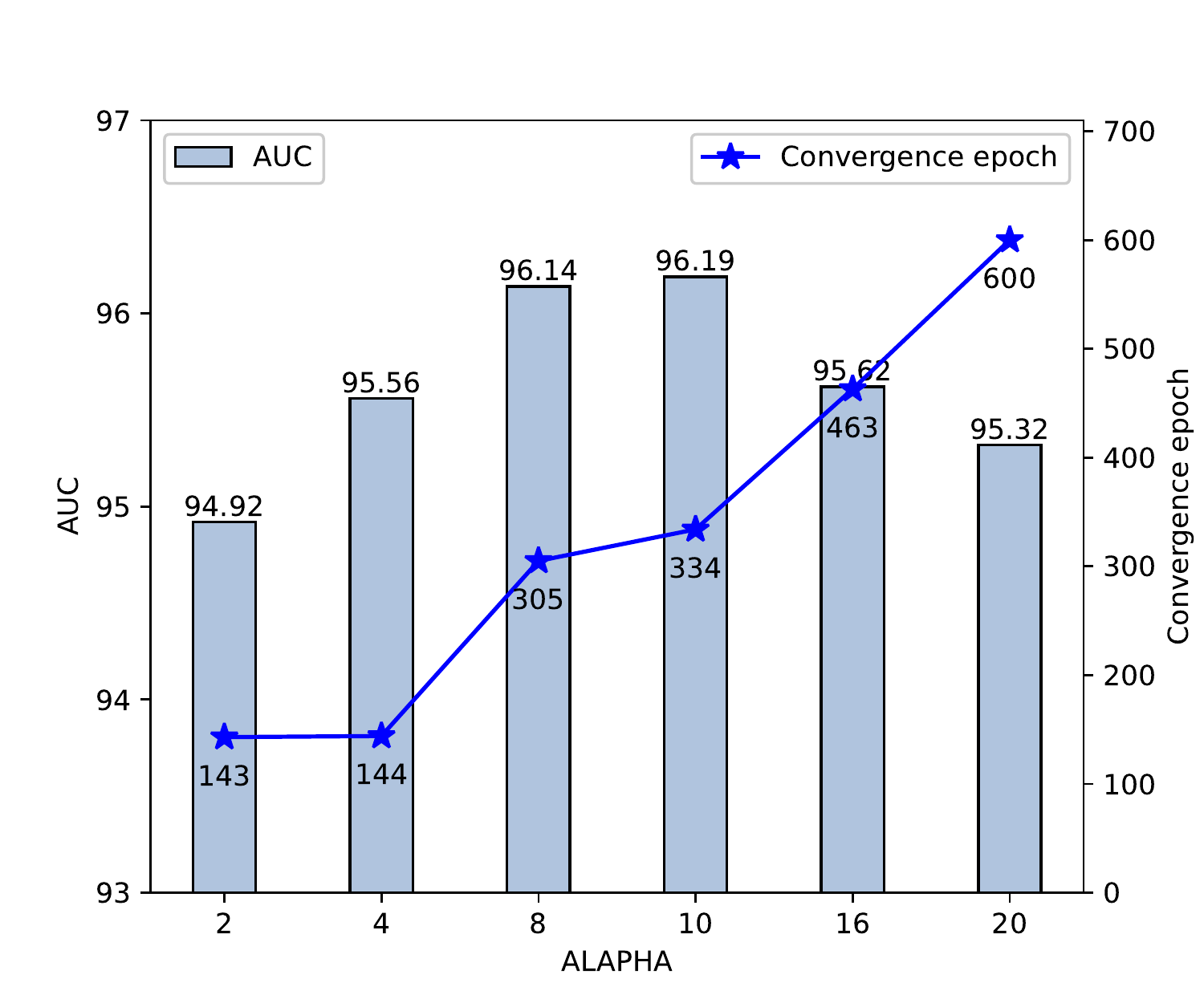}
\end{minipage}%
}%
\centering
\subfigure[MIMIC-III.]{\label{fig:alapha-c}
\begin{minipage}[t]{0.333\linewidth}
\centering
\includegraphics[width=5.3cm]{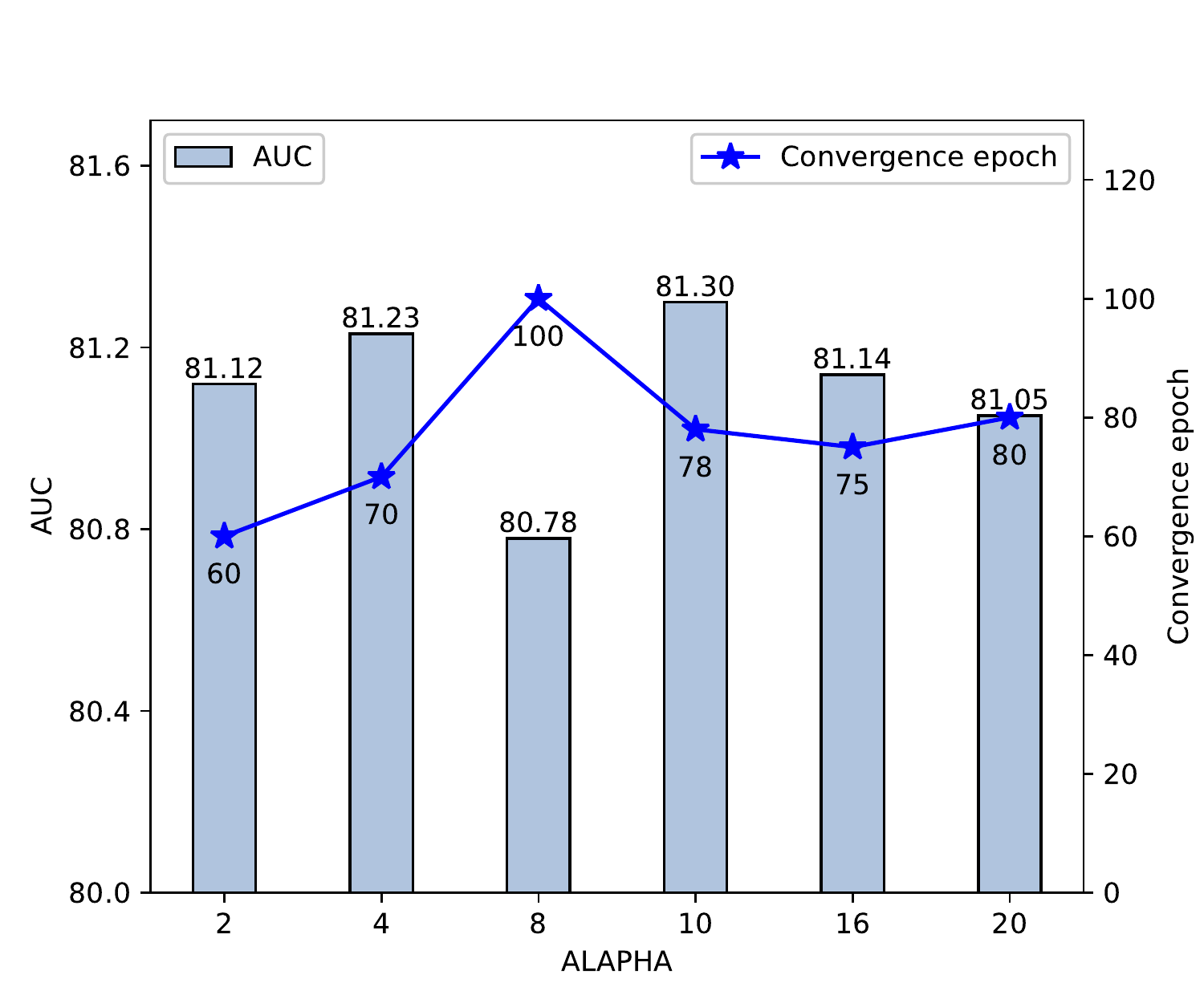}
\end{minipage}%
}%
\centering
\caption{Depth and Width for Different Task Scenarios.}\label{fig:alapha}
\end{figure}

\subsection{Versatility Analysis of \RSRL Framework}\label{sec:versatility}
To better adapt to many task-driven scenarios, we implement a general \RSRL reinforcement learning framework. 
In dealing with datasets of different sizes and types, different reinforcement learning algorithms and action space types can be flexibly matched. 
The depth and width of the reinforcement learning tree can be adaptively estimated for each relation.

\textbf{Algorithms and Action Space for Different Task Scenarios. }
In Section~\ref{sec:fraud-overall} and Section~\ref{sec:diabetes-overall}, we analyze the function variants (Table~\ref{tab:fraud_variants} and Table~\ref{tab:diabetes-variants}) and the difference between reinforcement learning algorithms under \RSRL framework and traditional reinforcement learning (Figure~\ref{fig:fraud-efficiency} and Figure~\ref{fig:RSRL provement_mic}) in terms of accuracy and efficiency through a classical reinforcement learning algorithm, Actor-Critic.
Here, in order to better explore the versatility of \RioGNN for different reinforcement learning algorithms and the applicability of different reinforcement learning algorithms for different tasks, in Table~\ref{tab:RL_algorithms} we select the current mainstream reinforcement learning algorithms for experiments, and record the best AUC value within $500$ epochs. 
In addition, for the \RSRL framework, we proposed two different action spaces for the construction of reinforcement learning forests in Section~\ref{sec:rsrl} to adapt to different task requirements.
Different from the discrete action space, the reinforcement learning framework with continuous action space has continuous precision (that is, the highest floating-point number of the processor) in every action selection of reinforcement learning. 
This difference makes it have a better exploration effect in large-scale dataset.
From the experimental results, PD3 algorithm continuous action space and two sets of networks to update the Q value achieves the best results in both Yelp and MIMIC-III dataset. 
In the Amazon dataset, the best result is obtained from the more basic discrete action space AC. SAC and PPO are at a low level in the three datasets. 
In general, \RioGNN is well-adapted to most reinforcement learning algorithms, and is a versatile framework for different types of dataset and task scenarios.

\textbf{Depth and Width for Different Task Scenarios. }
In Section~\ref{sec:rsrl}, we define a depth and width adaptive parameter $\alpha$ to adjust the size of the action space of each layer of the relation and the depth of the entire relation tree. 
In the previous experiment, we fixedly chose 10 as $\alpha$.
In this section, in order to discuss the impact of the depth and width adaptive parameter on the accuracy and efficiency of the \RioGNN model, we compare and analyze the AUC and convergence epoch sizes of the three dataset under different settings.
As shown in Figure~\ref{fig:alapha}, we set the six $\alpha$ values of $2$, $4$, $8$, $10$, $16$, $20$, and respectively record the maximum AUC and the corresponding epoch serial number obtained in $500$ epochs. 
In the Yelp dataset, AUC achieves the maximum value when $\alpha$ is $10$, which is at least $1.33\%$ better than other parameters. 
But in this case, it takes longer to reach this value. 
Therefore, we suggest that Yelp can be adaptively chosen $\alpha$ to $8$ or $10$ for efficiency priority and accuracy priority.
On the other hand, Amazon achieves better accuracy when $\alpha$ is $8$ or $10$, and achieves better efficiency when $\alpha$ is $2$ or $4$. 
The difference from the previous two is that the MIMIC-III dataset obtains a loss of accuracy and efficiency when $\alpha$ is $8$.
Judging from these situations, \RioGNN can be adapted to a dataset of different scales and different accuracy and efficiency biases by adjusting $\alpha$. 
This represents the versatility of the dataset size and task requirements.

\begin{table}[h]
    \setlength{\abovecaptionskip}{0.cm}
    \setlength{\belowcaptionskip}{-0.cm}
    \caption{Inductive learning results ($\%$) compared to \RioGNN variants.}\label{tab:inductive}
    \centering
    \scalebox{1}{
        \begin{tabular}{p{3cm}<{\centering}|p{1cm}<{\centering}p{1cm}<{\centering}p{1cm}<{\centering}|p{1cm}<{\centering}p{1cm}<{\centering}p{1cm}<{\centering}|p{1cm}<{\centering}p{1cm}<{\centering}p{1cm}<{\centering}}
            \hline
            \multicolumn{1}{c|}{\multirow{2}*{Models}}&\multicolumn{3}{c|}{\textbf{Yelp}}&\multicolumn{3}{c|}{\textbf{Amazon}}&\multicolumn{3}{c}{\textbf{MIMIC-III}}\\
            \cline{2-10}
            \multicolumn{1}{c|}{}&\textbf{AUC}&\textbf{Recall}&\textbf{F1}&\textbf{AUC}&\textbf{Recall}&\textbf{F1}&\textbf{AUC}&\textbf{Recall}&\textbf{F1}\\
            \hline
            \multicolumn{1}{c|}{GAT}&55.94&51.79&47.25&72.33&65.86&60.17&63.89&59.13&56.78\\
            \multicolumn{1}{c|}{GraphSAGE}&53.85&51.78&44.36&74.91&70.02&65.32&63.89&69.99&59.24\\
            \hline
            \multicolumn{1}{c|}{\RioGNN$_{2l}$}&79.45&71.86&63.58&92.01&83.65&86.24&79.01&69.77&69.64\\
            \hline
            \multicolumn{1}{c|}{BIO-GNN}&79.49&71.86&63.58&\textbf{95.07}&88.19&86.51&81.21&\textbf{72.81}&\textbf{72.64}\\
            \multicolumn{1}{c|}{ROO-GNN}&82.15&74.23&\textbf{67.73}&94.79&87.43&\textbf{88.67}&81.01&72.39&72.23\\
            \hline
            \multicolumn{1}{c|}{RIO-Att}&78.72&71.78&62.38&93.79&\textbf{88.71}&83.72&79.84&71.31&71.28\\
            \multicolumn{1}{c|}{RIO-Weight}&81.06&72.79&65.59&94.67&88.58&85.12&\textbf{81.25}&72.72&72.28\\
            \multicolumn{1}{c|}{RIO-Mean}&78.17&71.41&62.12&93.53&87.32&85.75&80.29&71.92&71.74\\
            \hline
            \multicolumn{1}{c|}{\RioGNN}&\textbf{82.38}&\textbf{75.08}&65.26&94.03&88.58&86.46&81.23&72.63&72.53\\
            \hline
        \end{tabular}
    }
\end{table}

\begin{figure}[t]
\centering
\subfigure[Yelp (pos:neg).]{\label{fig:sample-yelp}
\begin{minipage}[t]{0.333\linewidth}
\centering
\includegraphics[width=5.3cm]{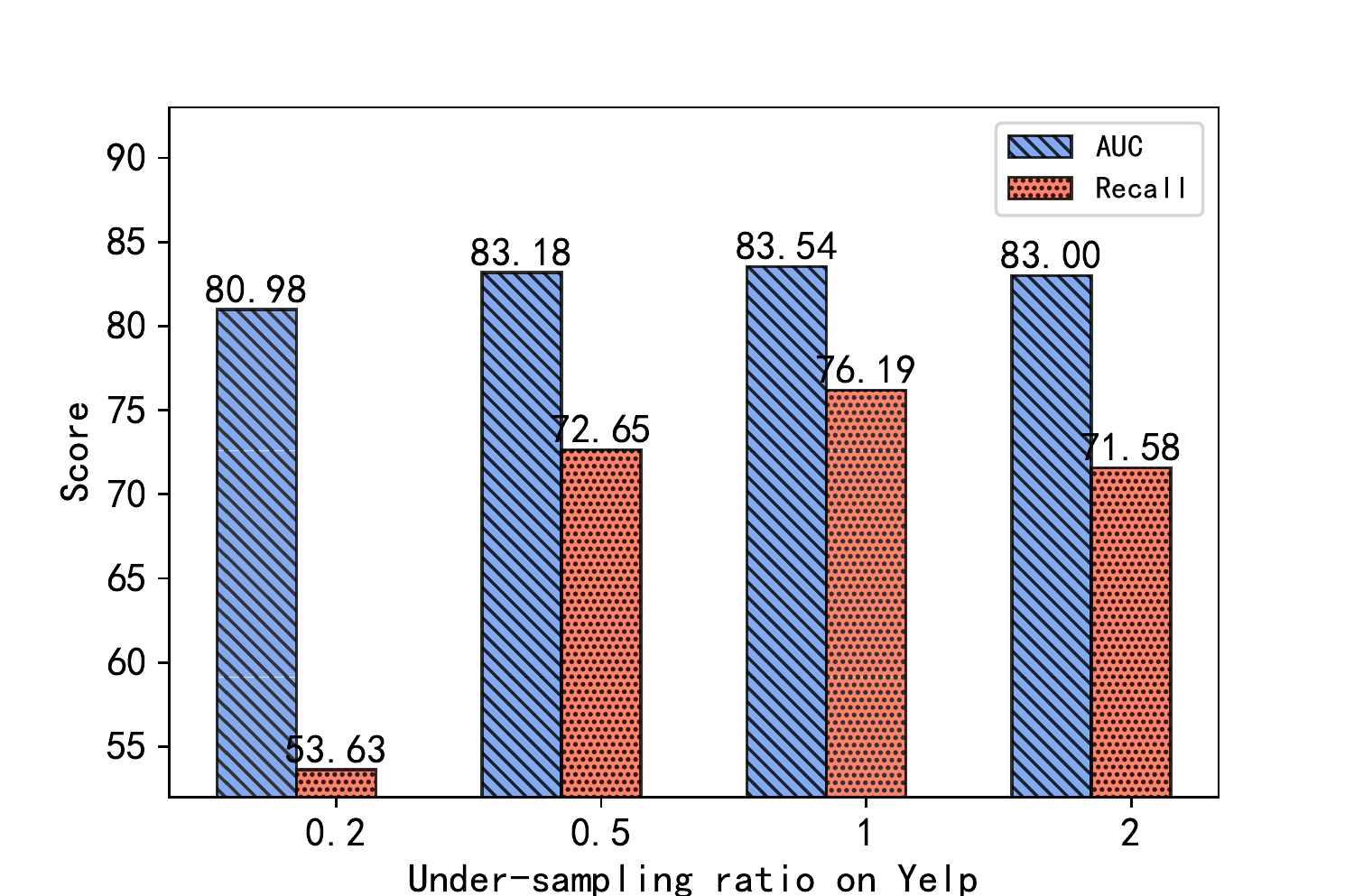}
\end{minipage}%
}%
\subfigure[Amazon (pos:neg).]{\label{fig:sample-amazon}
\begin{minipage}[t]{0.333\linewidth}
\centering
\includegraphics[width=5.3cm]{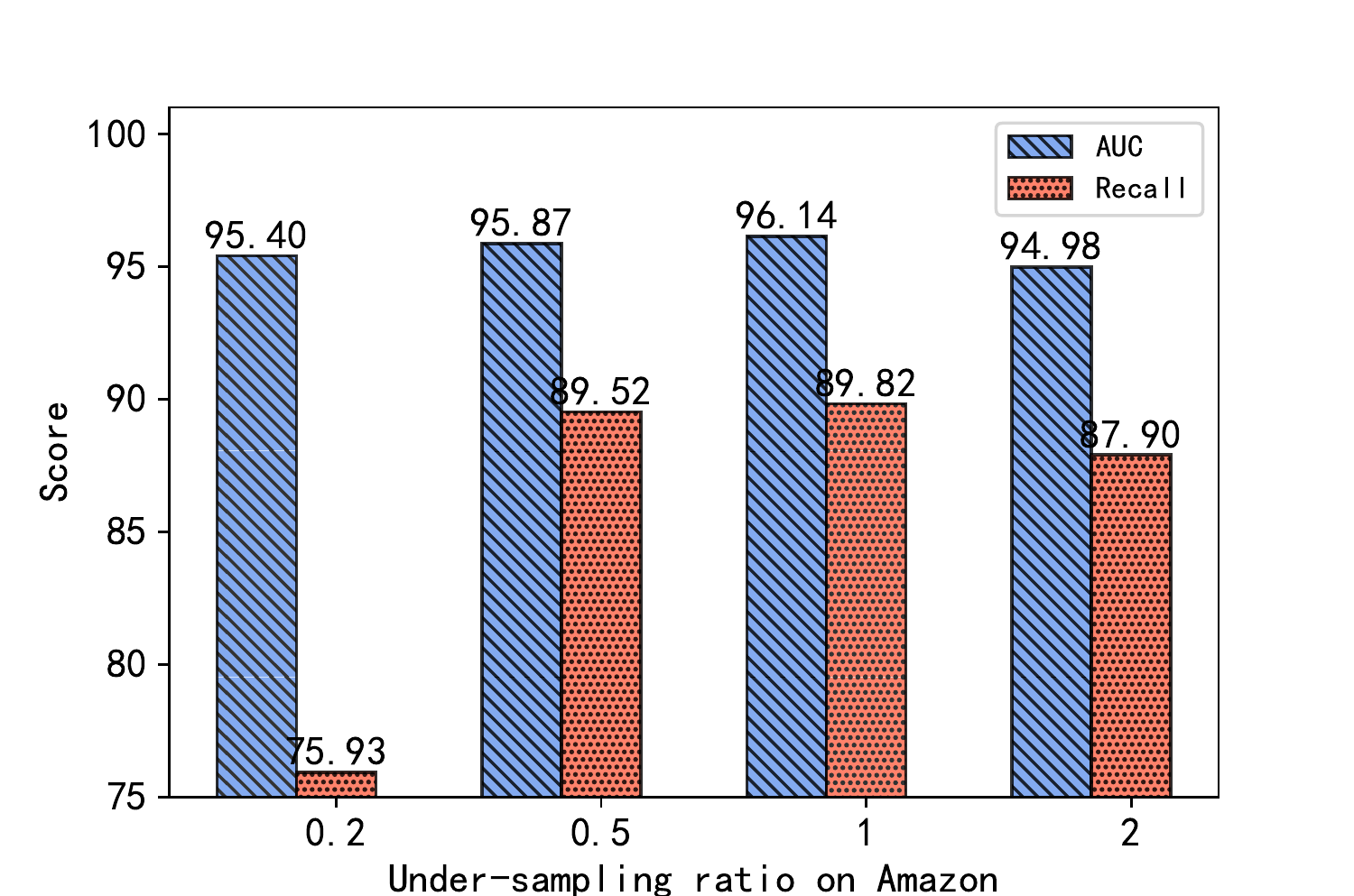}
\end{minipage}%
}%
\centering
\subfigure[MIMIC-III (pos:neg).]{\label{fig:sample-mimic}
\begin{minipage}[t]{0.333\linewidth}
\centering
\includegraphics[width=5.3cm]{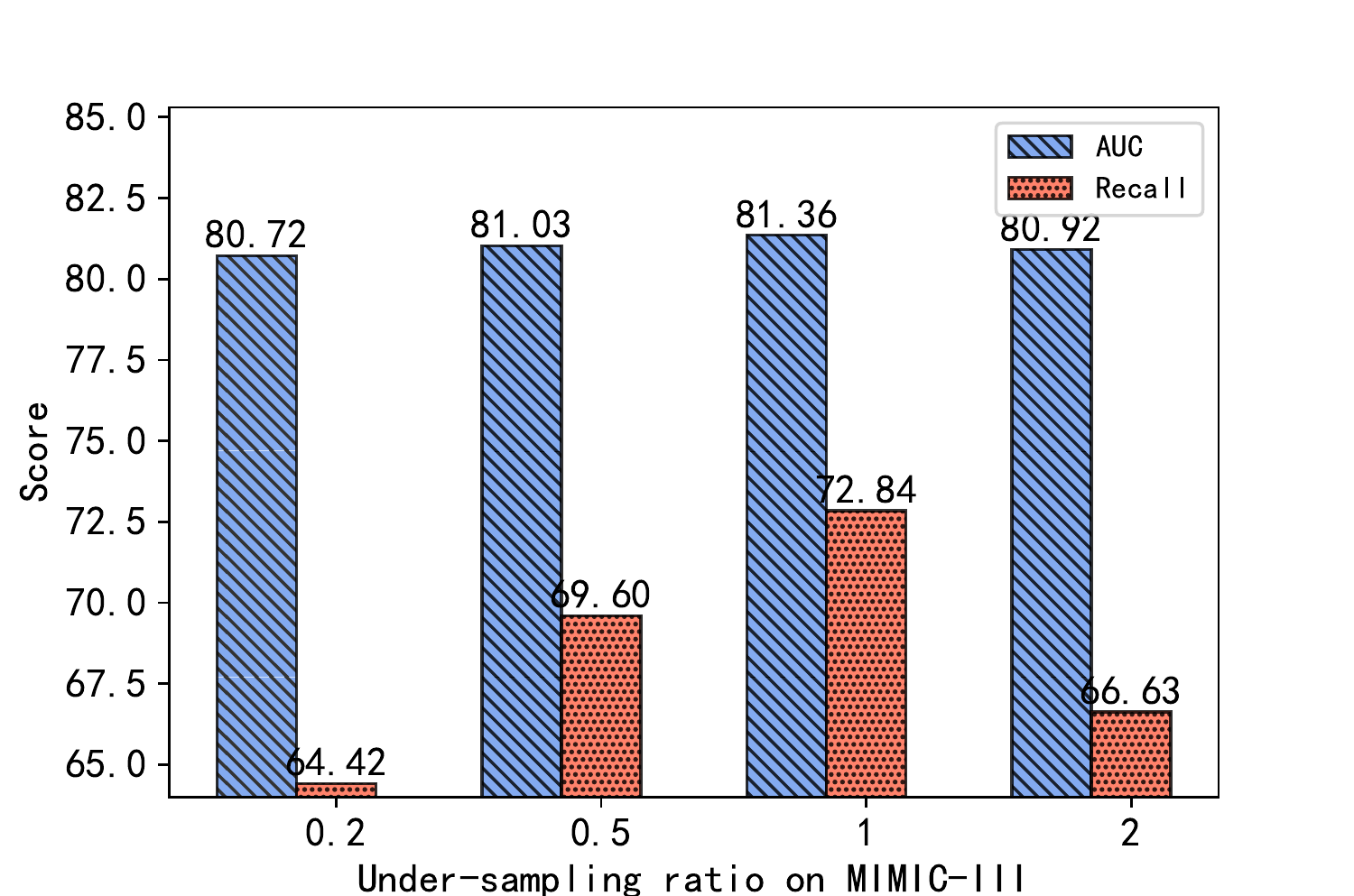}
\end{minipage}%
}%
\centering

\subfigure[Yelp.]{\label{fig:backtracking-yelp}
\begin{minipage}[t]{0.333\linewidth}
\centering
\includegraphics[width=5.3cm]{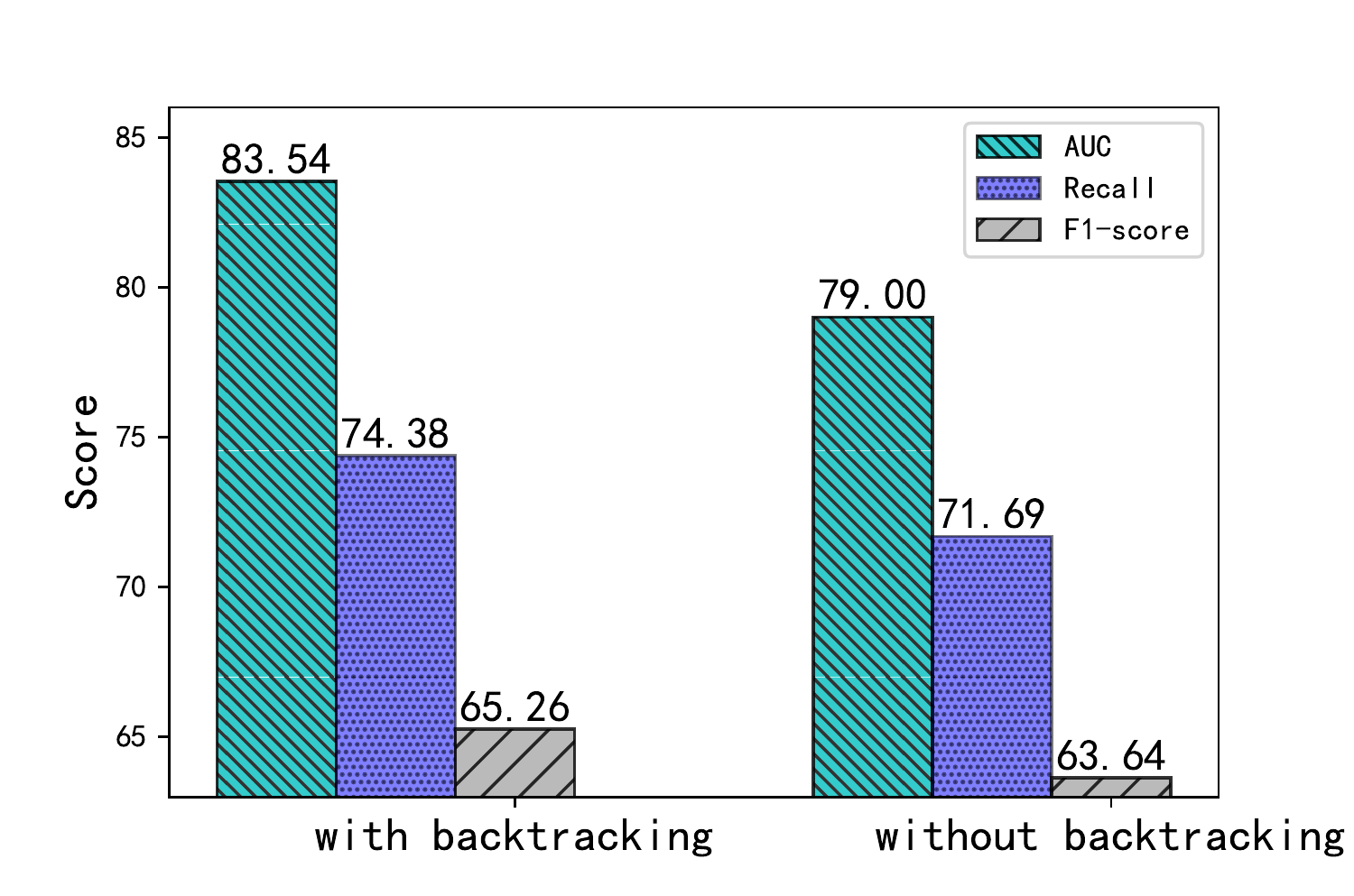}
\end{minipage}%
}%
\subfigure[Amazon.]{\label{fig:backtracking-amazon}
\begin{minipage}[t]{0.333\linewidth}
\centering
\includegraphics[width=5.3cm]{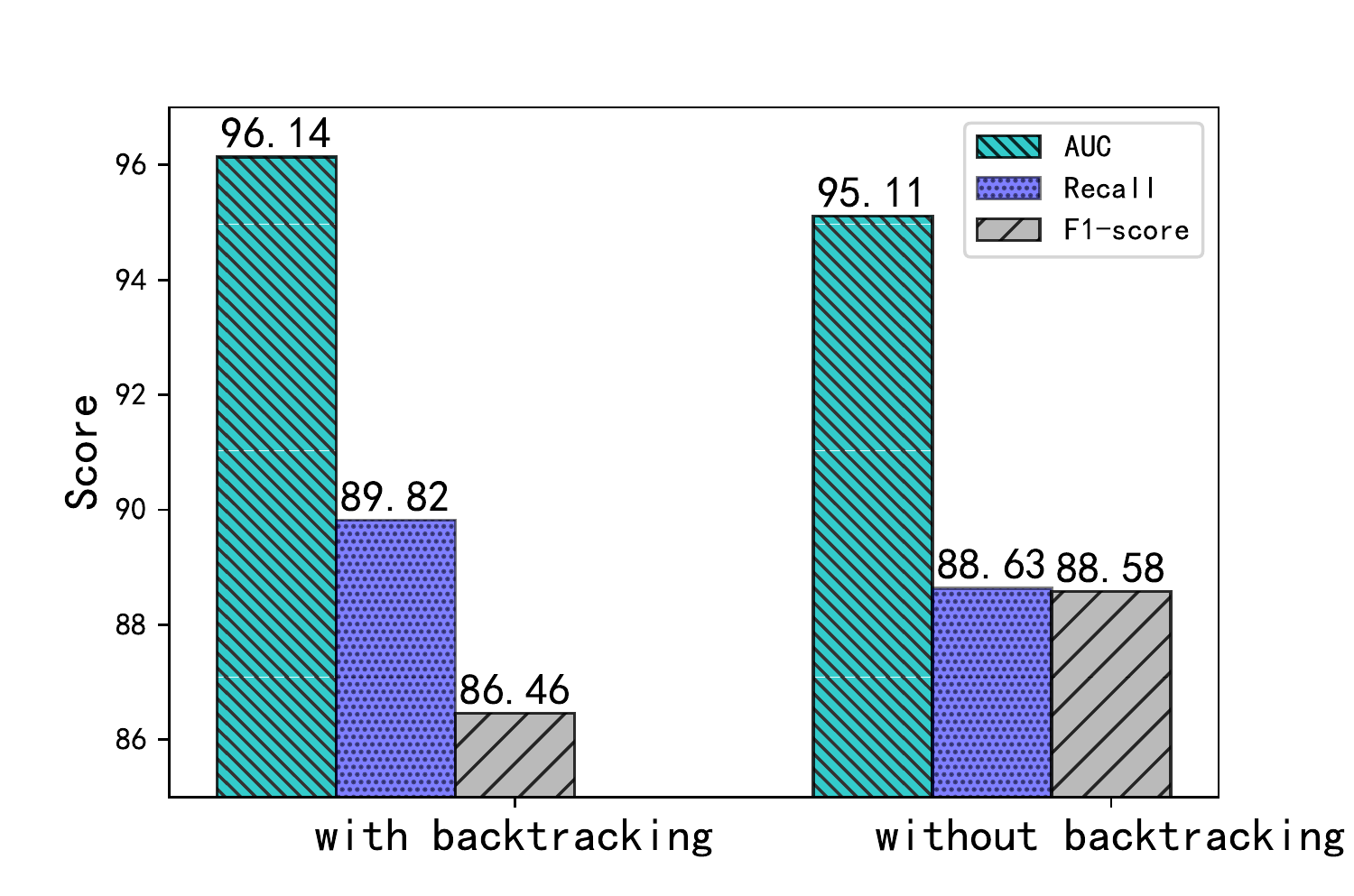}
\end{minipage}%
}%
\centering
\subfigure[MIMIC-III.]{\label{fig:backtracking-mimic}
\begin{minipage}[t]{0.333\linewidth}
\centering
\includegraphics[width=5.3cm]{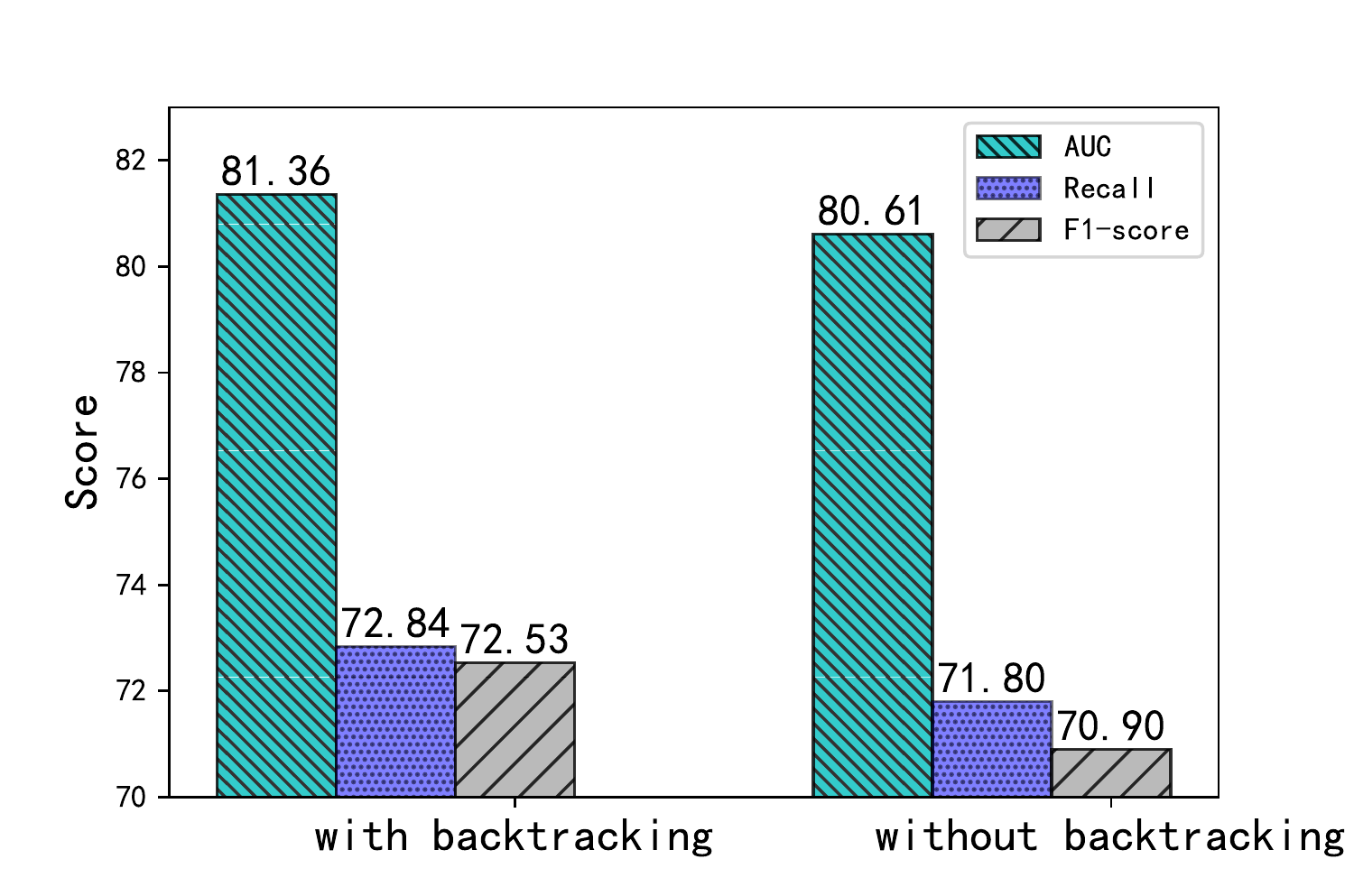}
\end{minipage}%
}%
\centering

\subfigure[Yelp.]{\label{fig:stopnum-yelp}
\begin{minipage}[t]{0.333\linewidth}
\centering
\includegraphics[width=5.3cm]{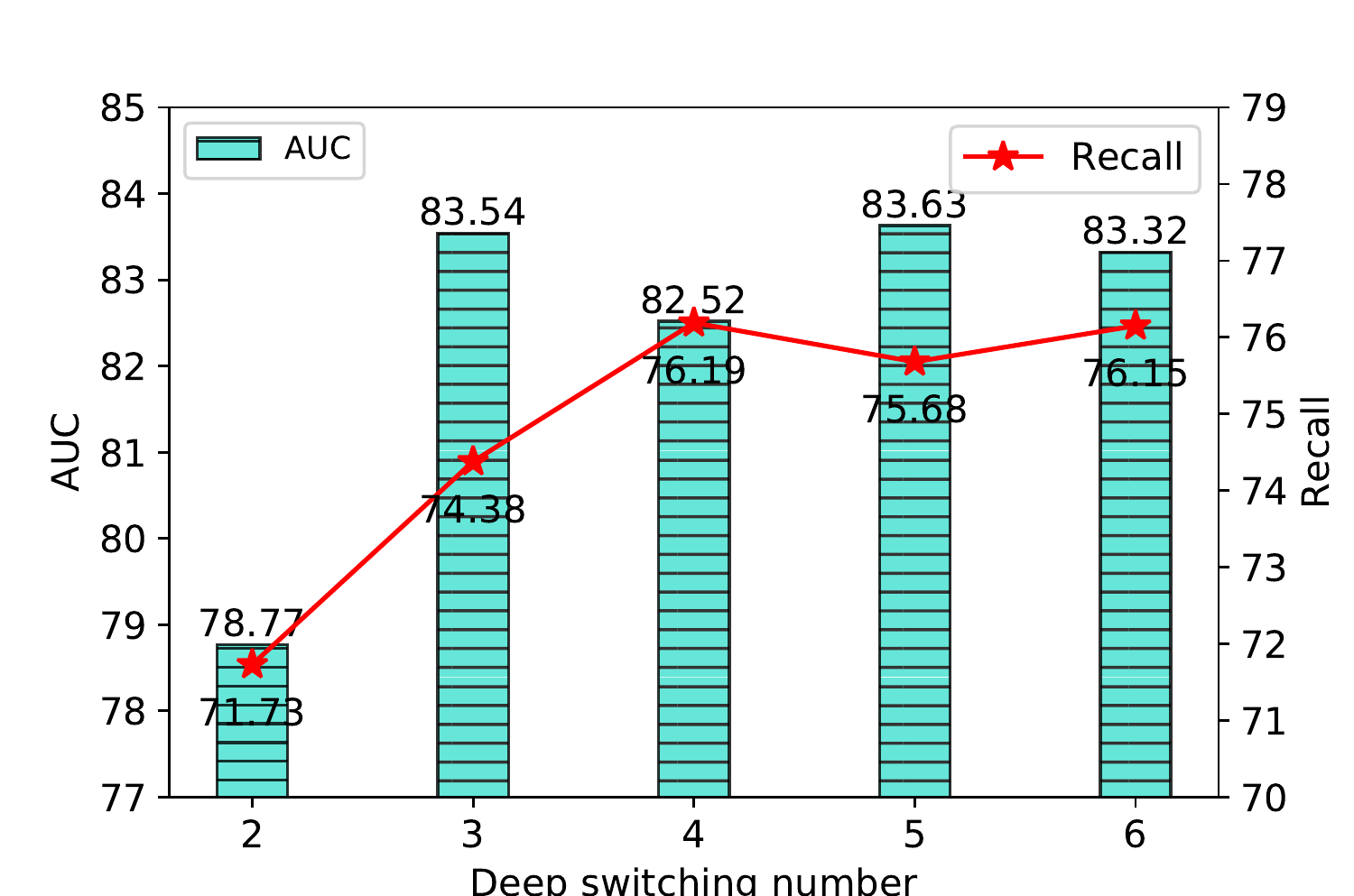}
\end{minipage}%
}%
\subfigure[Amazon.]{\label{fig:stopnum-amazon}
\begin{minipage}[t]{0.333\linewidth}
\centering
\includegraphics[width=5.3cm]{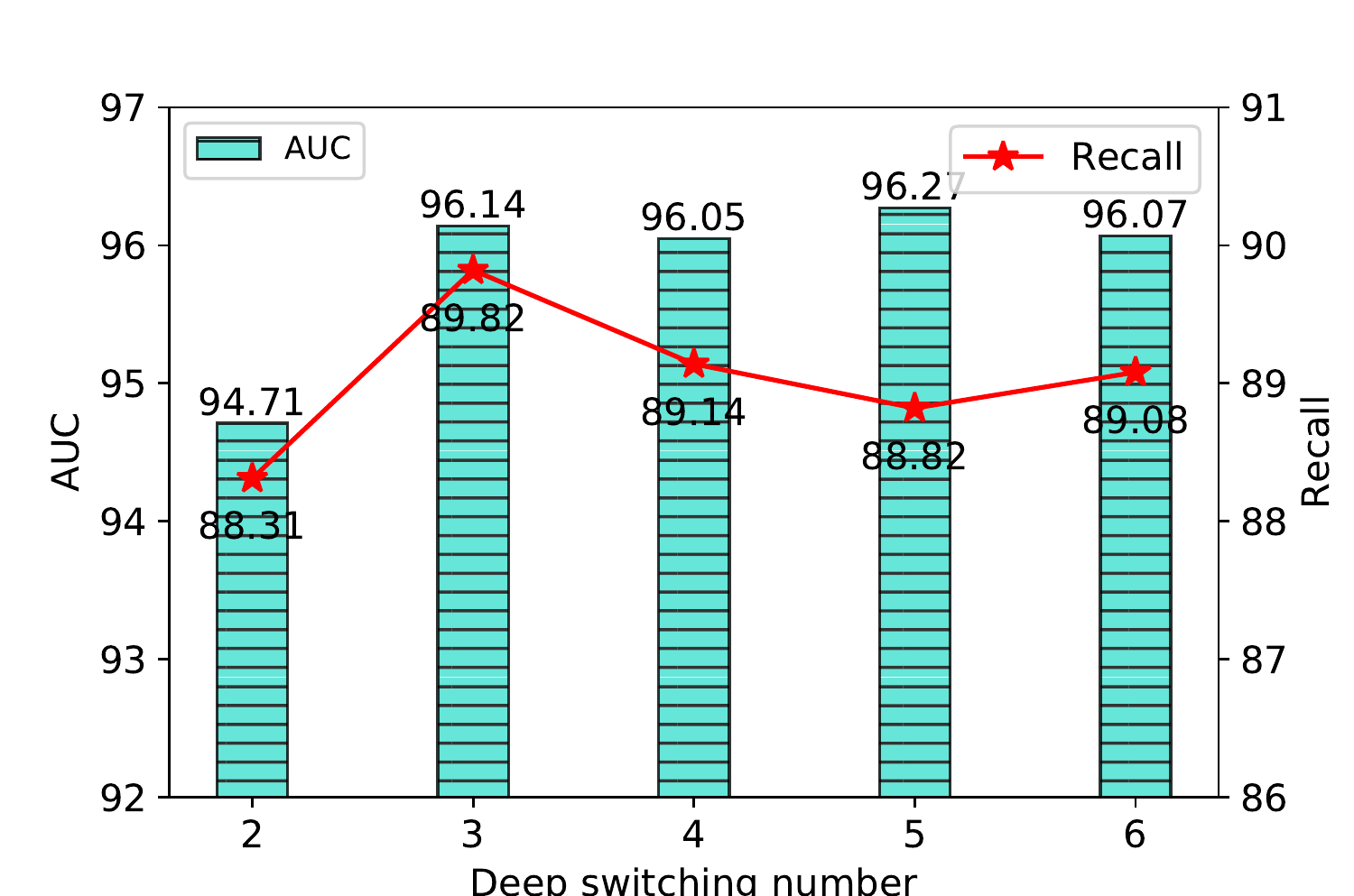}
\end{minipage}%
}%
\centering
\subfigure[MIMIC-III.]{\label{fig:stopnum-mimic}
\begin{minipage}[t]{0.333\linewidth}
\centering
\includegraphics[width=5.3cm]{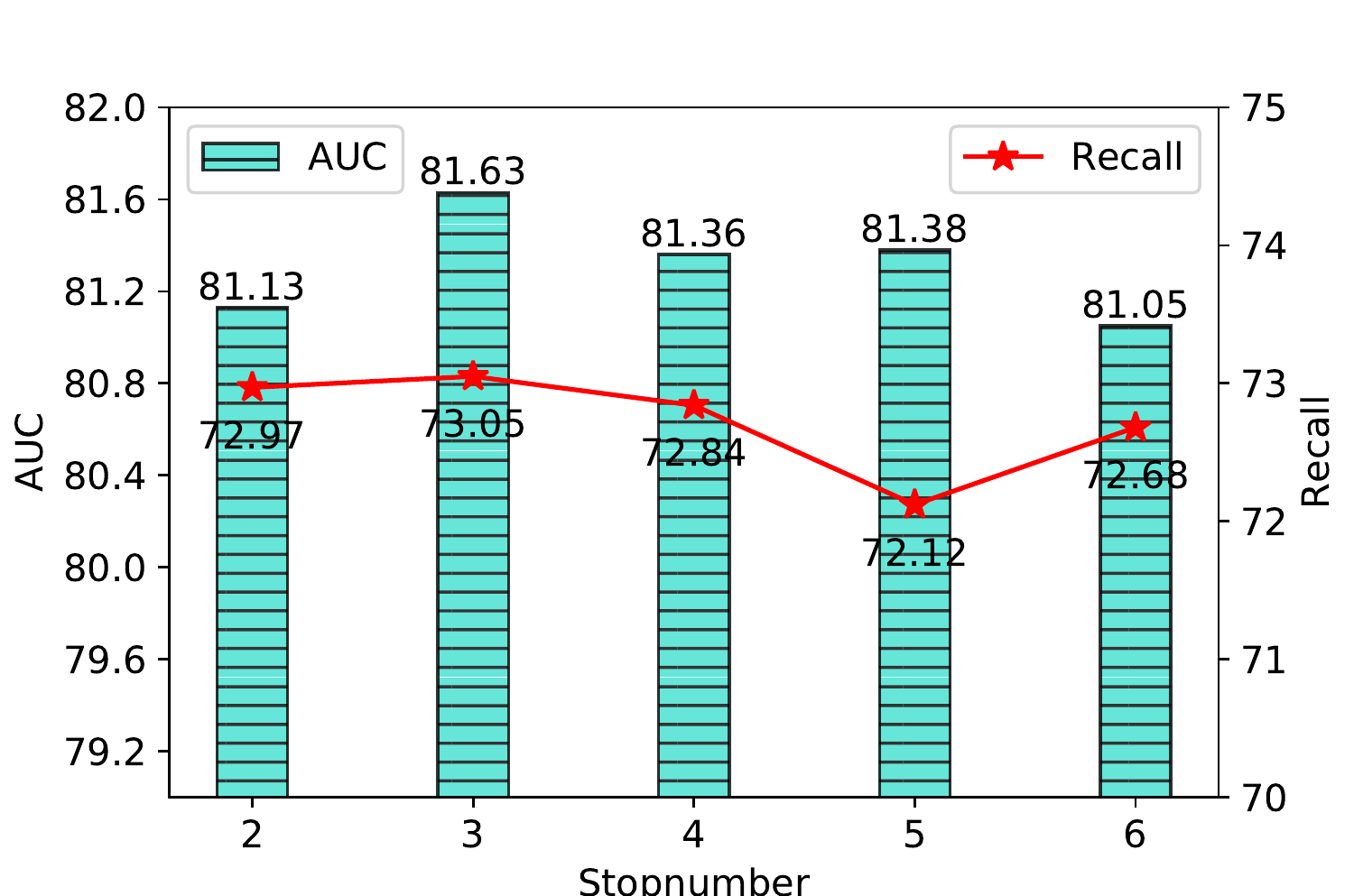}
\end{minipage}%
}%
\centering
\caption{Parameter sensitivity of under-sampling ratio, with \& without backtracking, deep switching number on Yelp, Amazon and MIMIC-III.}\label{fig:parameter}
\end{figure}

\subsection{Inductive Learning Analysis}\label{sec:inductive}
In this section, we perform inductive learning on \RioGNN, some representative baselines and variant models of \RioGNN.
In the previous experiment, we use transductive learning, that is, the graph passed into the model contains the test nodes. 
In inductive learning, we only pass the adjacency matrix of the nodes that need to be trained into the model, and record the best AUC, Recall and F1 indicators of Yelp and Amazon within $500$ epochs and MIMIC-III within $700$ epochs.
From the results in Table~\ref{tab:inductive}, it can be seen that \RioGNN still has obvious advantages compared with GAT and GraphSAGE, where AUC, Recall and F1 increase by $17.34\%$-$28.53\%$, $2.64\%$-$23.30\%$, $13.29\%$-$20.90\%$.
In addition, \RioGNN has a relatively stable evaluation index among many variants. 
Among them, the results on Yelp dataset are compared with those with transductive learning in Table~\ref{tab:fraud_variants}.
In the inductive learning, the AUC and Recall rate of \RioGNN constantly surpasses ROO-GNN variants although \RioGNN is slightly lower than ROO-GNN in the transductive learning shown in Table~\ref{tab:inductive}.
This represents the stability of the performance of the recursive framework in challenging tasks and illustrates its advantages in small-scale scenarios.
In the Amazon dataset, due to the expansion of the data scale, some variants are better evaluated in some aspects, but \RioGNN is stable at a relatively high level from the comprehensive situation of AUC, Recall and F1. 
This situation similarly appears in MIMIC-III. 
It is worth noting that the BIO-GNN variants in the Amazon and MIMIC-III datasets achieve a good performance improvement compared with their situation in the transductive learning task. 
We believe this is because BMAB has a relatively weak learning ability compared to Actor-Critic, which reduces the dependence of the learned model on the training set, so it is more compatible with newly added nodes.
Overall, \RioGNN has good applicability in both transductive learning and inductive learning.

\subsection{Hyper-parameter Sensitivity}\label{sec:hyper-parameter}
Figure~\ref{fig:parameter} shows the test performance of the three hyper-parameters we introduce in Section~\ref{sec:model-training} in three datasets.
The first row of Figure~\ref{fig:parameter} shows the AUC and Recall of the training set of \RioGNN at different sampling ratios (note that the test results come from an unbalanced test set). 
It can be seen that when the sampling ratio is $1:0.2$, that is, when the negative samples are much smaller than the positive samples, overfitting occurs in all three datasets. 
Compared with $1:0.5$ and $1:2$ sampling ratios, $1:1$ sampling show higher AUC and Recall indicators in all three datasets.
The second row of Figure~\ref{fig:parameter} studies the backtracking structure we set in Section~\ref{sec:rsrl}. 
From Figure~\ref{fig:backtracking-yelp}, Figure~\ref{fig:backtracking-amazon} and Figure~\ref{fig:backtracking-mimic}, models with backtracking settings bring stable performance in all datasets compared to models without backtracking.
In the third row of Figure~\ref{fig:parameter}, we test different depth switching conditions. 
When the deep switching number is set to $3$, AUC and Recall achieve good and balanced performance.

%% file: 6-Relatedwork.tex
\section{Related Work}\label{sec:relatedwork}

In the past years, Graph Neural Networks (GNNs) and Reinforcement Learning (RL) technologies have received increasing attention and many upgraded algorithms have been proposed.
Hence, the existing literature can be roughly classified into three categories: semi-supervised graph neural networks, RL, and RL-guided GNNs.

\subsection{Semi-supervised Graph Neural Networks}
According to the difference in data modeling for real-world graph data, we roughly divide the semi-supervised graph neural network methods into homogeneous graph neural networks, heterogeneous graph neural networks, and multiple graph learning models.

\textbf{Homogeneous Graph Neural Networks.} 
They are usually referred to those GNN methods that do not consider the data type of nodes or the attributes of the edges on the graph. 
Classical methods include GCN~\cite{kipf2017semi}, Graph-SAGE~\cite{hamilton2017inductive} and GAT~\cite{velivckovic2018graph}.
As discussed in the previous section, the GCN model defines the first successful graph convolutions in analogy to convolutional layers over Euclidean data, and is thus seen as a generalization of convolution neural networks (CNNs) for non-grid topologies.
The Graph-SAGE model exploits sampling for obtaining a fixed number of neighbors for each node to generate node embedding via aggregation functions, which can be invariant if the permutations of node orderings, such as a mean, sum, or max function, are applied. 
Moreover, the Graph-SAGE model presents the first general inductive learning framework that continuously samples and aggregates its local neighbors' features to generate embedding for the new node.  
By contrast, the GAT model firstly adopts attention mechanisms to learn the relative weights between two connected nodes. The multi-head self-attention is further enforced to increase the model's expressive capability. 
Despite the powerful graph representation learning capability of these models, the main limitation is the ignorance of the diversity of data types and relationships manifesting in the real-world data and applications.

\textbf{Heterogeneous Graph Neural Networks.} 
Such approaches generally consider the heterogeneity of node types or edge types when aggregating feature information from node's local neighbors via neural networks.
The classical meta-path and meta-graph based methods include GAS~\cite{li2019spam}, HAN~\cite{wang2019heterogeneous}, Player2Vec~\cite{zhang2019key}, HSGNN~\cite{liu2020health} and MAGNN~\cite{fu2020magnn}.
Considering the diversity of edges in real-world data, more relational graph neural network methods including R-GCN~\cite{schlichtkrull2018modeling}, SemiGNN~\cite{wang2019semi}, FdGars~\cite{wang2019fdgars} and GraphConsis~\cite{liu2020alleviating} are developed.
Other heterogeneous graph neural networks, including HGT~\cite{hu2020heterogeneous}, GEM~\cite{liu2018heterogeneous}, HetSANN~\cite{hong2020attention}, etc., implement complex neural aggregations among heterogeneous neighbors.
All these heterogeneous models are upgraded versions of the previous homogeneous models.
Nevertheless, there is no literature exploring how to select neighbor nodes to build the most expressive, explanatory and stable aggregation.

\textbf{Multiple Graph Learning Models.}
Apart from the above homogeneous and heterogeneous GNNs that solve single-graph representation learning, multi-graph neural network models~\cite{Yangsurvey2021,ma2018drug,zhang2018multi,ma2017multi,sun2020predicting} study fusing the multiple characterizes to comprehensively learn the embedding of graph data objects.
MGAT~\cite{xie2020mgat} explores both attention-based architecture for learning node representations from each single view and view-focused attention method to aggregate the view-wise node representations. 
A multi-view knowledge graph embedding~\cite{MultiKE2019} is presented by using cross-view entity identity inference to capture the alignment information between two knowledge graphs.
In order to filter out useless feature interactions, a Bayesian Personalized Feature Interaction Selection mechanism~\cite{Chenbayesian2019} is designed under the Bayesian Variable Selection (BVS) theory in recommendation tasks.
Moreover, a block-diagonal regularization~\cite{Chenblock2020} is proposed to guide the item similarities in the top-N recommendation task.

\subsection{Reinforcement Learning}
With the development of technology, reinforcement learning algorithms have derived many different development directions.
The more basic algorithms are value-based only Q-Learning~\cite{watkins1992q} and DQN~\cite{mnih2015dqn} algorithms, which use value functions to estimate and reduce the occurrence of local optimal situations. 
However, policy-based only algorithms such as PPO~\cite{schulman2017proximal} directly performs iterative calculation on the policy, which can achieve better convergence. 
The Actor-Critic type of reinforcement learning methods AC~\cite{konda2000actor}, DDPG~\cite{lillicrap2019ddpg}, TD3~\cite{scott2018td3}, and SAC~\cite{he2019hetespaceywalk} combine the advantages of value-based and policy-based to train Q functions and strategies at the same time.
In addition to the above division methods, from the perspective of action space types, DQN and Q-learning are suitable for discrete action spaces, DDPG, TD3, and SAC support continuous action spaces, while PPO and AC are suitable for both discrete and continuous action spaces.
Or from the perspective of learning methods, reinforcement learning algorithms such as DDPG, DQN, SAC, and TD3 combine deep learning and use the fitting ability of neural networks to obtain better optimization.
These different algorithms have different advantages, but also bring different limitations. For example, algorithms that only support discrete action spaces are incapable of continuous action space requirements, Actor-Critic type algorithms have inherited the desired shortcomings while absorbing the value-based method and the policy-based method. 
The framework that only supports the same reinforcement learning algorithm has limitations in adaptability to many types of tasks.

\subsection{Combination GNNs and RL}
There are a few attempts to marry GNNs and RL.
DGN+GNN~\cite{almasan2019deep} is a model used to generalize unseen network topologies, where GNNs that model the network environment allow the DRL agent to operate on different networks. 
G2S+BERT+RL~\cite{chen2019reinforcement} is a RL based graph-to-sequence model for natural question generation, where the answer information is utilized by an effective Deep Alignment Network and a novel bidirectional GNN is proposed to process the directed passage graph. 
Similarly, other work~\cite{janisch2020symbolic,hart2020graph,sun2020combining} investigates how to use GNNs to improve the generalization ability of RL. 
There are also numerous studies that leverage RL to optimize representation learning on graphs.
For example, DeepPath~\cite{xiong2017deeppath} a knowledge graph embedding and reasoning framework based on RL policy-based; the RL agent is trained to ascertain the reasoning paths in the knowledge base. 
RL-HGNN~\cite{zhong2020reinforcement} devises different meta-paths for any node in a HIN to learn its effective representations. It models the process of meta-path design as a Markov Decision Process by using a DRL-based policy network for adaptive meta-path selection.
As opposed to \RioGNN, the RL-HGNN model pays more attention to revealing meaningful meta-paths or relations in heterogeneous graph analysis.
GraphNAS~\cite{gao2019graphnas} employs a search space covering sampling functions, aggregation functions and gated functions and uses RL to search graph neural architectures.
Policy-GNN~\cite{Lai2020policy} formulates the GNN training problem as a Markov Decision Process, and can adaptively learn an aggregation policy to sample diverse iterations of aggregations for different nodes. 
However, neither GraphNAS nor Policy-GNN models considers heterogeneous neighborhoods in aggregation although they pay more attention to neural architecture searching.

%% file: 7-Conclusion.tex
\section{Conclusion and Future work}\label{sec:conclusion}

This paper studies \RioGNN, a reinforced, recursive and flexible neighborhood selection guided multi-relational Graph Neural Network architecture, to learn more discriminative node embedding and respond to the explanation of the importance of different relations in spam review detection and disease diagnosis tasks, respectively.
\RioGNN designs a label-aware neural similarity neighbor measure and reinforced relation-aware neighbor selectors using reinforcement learning technology, respectively.
To optimize the computational efficiency of the reinforcement neighbor selecting, we further design a recursive and scalable framework with estimable depth and width for different scales of multi-relational graphs.
The conducted experiments on three real-world benchmark datasets suggest that \RioGNN significantly, consistently and steadily outperforms the state-of-the-art alternatives across all the datasets.
Our work shows the promise in learning a reinforced neighborhood aggregation for GNNs, potentially opening new avenues for future research in boosting the performance of GNNs with adaptive neighborhood selection and analysing the importance of different relations in message passing.

In the future, we aim to adopt a multi-agent RL algorithm to further enable the \RioGNN to adaptively identify meaningful relations for each node, instead of the manual efforts in defining relations, for automated representation learning on heterogeneous data.
In addition, it is also interesting to study how to extend our models to other tasks on graph data analysis and application, such as the personalized recommendation system, social network analysis, and etc.